\documentclass[11pt,letterpaper]{article}
\usepackage[margin=1in]{geometry}
\usepackage[T1]{fontenc}
\usepackage{amsmath,amssymb,amsthm}
\newcommand{\E}{\mathbb E}
\newcommand{\KL}{\operatorname{KL}}
\newcommand{\Var}{\operatorname{Var}}

\usepackage[authoryear,round]{natbib}
\setcitestyle{citesep={,}}
\usepackage{graphicx,booktabs,array,float,needspace,placeins,pdflscape,wrapfig}
\usepackage{algorithm,algpseudocode}
\usepackage[font=small,skip=4pt]{caption}
\usepackage[table]{xcolor}

\usepackage{eso-pic}
\definecolor{paperlink}{HTML}{245A81}
\usepackage{hyperref}
\hypersetup{colorlinks=true,allcolors=paperlink,pdfborder={0 0 0},
 pdftitle={Regularized Score Centering},
 pdfauthor={Hao Bai, Chenlu Ye, Xuanchang Zhang, Aviral Kumar, Tong Zhang}}
\usepackage{url}
\newcommand{\MethodSC}{SC}
\newcommand{\MethodTIS}{SC+TIS}
\newcommand{\MethodNaiveOnly}{IS}
\newcommand{\MethodISPPOClip}{IS-PPOClip}
\newcommand{\MethodCurrentKL}{RSC (w/o IS)}
\newcommand{\MethodTISCurrentKL}{RSC (TIS)}

\newtheorem{theorem}{Theorem}[section]
\newtheorem{lemma}[theorem]{Lemma}
\newtheorem{proposition}[theorem]{Proposition}

\theoremstyle{definition}
\newtheorem{assumption}[theorem]{Assumption}
\theoremstyle{remark}

\title{Regularized Score Centering}
\author{Hao Bai$^{1,*}$\quad Chenlu Ye$^{1,*}$\quad Xuanchang Zhang$^{1}$\\
Aviral Kumar$^{2}$\quad Tong Zhang$^{1}$\\[0.5em]
{\small $^{1}$UIUC\qquad $^{2}$CMU}}
\date{}
\makeatletter
\renewcommand{\@maketitle}{\newpage\null\vspace*{-12pt}
  {\centering
    {\LARGE\@title\par}\vskip .6em
    {\large\lineskip .5em\begin{tabular}[t]{c}\@author\end{tabular}\par}}\vskip .25em}
\makeatother
\begin{document}
\maketitle
\begingroup
\renewcommand{\thefootnote}{\ensuremath{*}}
\footnotetext[1]{Equal contribution. Correspondence: Hao Bai (\href{mailto:haob2@illinois.edu}{haob2@illinois.edu}).}
\endgroup
\begingroup
\centering\small
Project page: \url{https://biechi.github.io/blog/rsc/}\par
\endgroup
\begin{abstract}
Asynchronous reinforcement learning increases training throughput, but data from stale policies and numerical mismatch between training and inference can destabilize learning. To stabilize off-policy learning, score centering (SC) removes the effect of reward baselines on the expected update, even without full importance sampling (IS).
Our empirical study shows that full IS produces extreme gradients and that SC significantly reduces their magnitude. However, SC can still fail when the learner moves far from the behavior policy that generated the data. To limit this policy mismatch, we introduce Regularized Score Centering (RSC), which adds a reverse Kullback--Leibler (KL) penalty to encourage the learner's next-token distribution to stay close to the behavior policy's next-token distribution.
RSC works with or without truncated importance sampling (TIS), which caps token importance weights.
Our theoretical analysis supports this approach, showing that KL regularization improves convergence guarantees.
We evaluate dense and mixture-of-experts models on mathematical reasoning and agentic tasks, including online shopping, tool use, games, and fully asynchronous software engineering.
RSC with TIS substantially improves sample efficiency across all agentic benchmarks, attaining higher training rewards with the same number of sampled responses.
In final evaluations, it achieves top performance across all six agentic tasks, improving by $14.3\%$ on AppWorld and reaching $9.9\times$ the strongest baseline's score on BabyAI.
\end{abstract}
\begin{figure}[H]
\centering
\includegraphics[width=\linewidth]{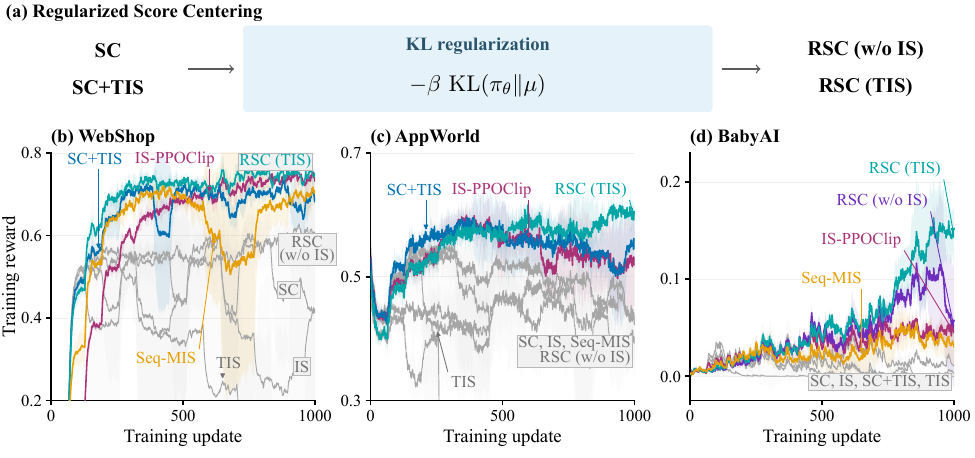}
\caption{\textbf{Regularization strengthens score centering.}
(a) RSC adds reverse KL to SC or SC+TIS, comparing distributions over next actions under the learner $\pi_\theta$ and behavior policy $\mu$ at the same history.
(b--d) Training curves show mean $\pm$ sample standard deviation across three seeds (settings in Appendix~\ref{sec:settings-agentic}). Stronger methods are colored, and others are gray.}
\label{fig:kl-role-agentic}
\end{figure}
\section{Introduction}\label{sec:introduction}
Efficient reinforcement learning often uses data generated by a policy that differs from the current learner.
Asynchronous reinforcement learning (async RL) creates this difference by running generation and training concurrently: while a response is being generated, the learner can move on to a newer policy. Reusing responses and refreshing the sampler less frequently further separate the learner $\pi_\theta$ from the behavior policy $\mu$ that generated the data~\citep{espeholt2018impala,hilton2022batch}.
A difference can also arise without any change in model weights. When training and inference use different numerical precision or execution kernels, they can assign different probabilities to the same action, creating a mismatch between training and inference (TIM)~\citep{yao2025rolloutmismatch}. In both cases, training is off-policy: the data-generating distribution differs from the current learner. This matters because an update computed from behavior data may differ from the update the learner would obtain from its own responses. In agentic tasks, the difference can be larger when actions lead to different future states and rewards. Section~\ref{sec:math-results} analyzes how task structure and policy staleness contribute to this effect.

To account for this difference, importance sampling (IS) gives behavior samples more or less weight according to their probability under the learner. Full sequence-level IS uses the likelihood ratio of the complete response. When $\mu$ assigns positive probability to every response that $\pi_\theta$ can generate, weighting behavior samples by this ratio gives an unbiased estimate of the learner's reward gradient, but multiplying token ratios can produce extremely high variance~\citep{owen2013montecarlo}. Practical sequence-level methods such as GSPO stabilize the weights through length normalization and clipping~\citep{zheng2025gspo}. Another way to avoid multiplying ratios across the response is to use token-level IS, which corrects only the distribution over the next action at a history sampled under the behavior policy. Large token ratios can still cause high gradient variance~\citep{bai2026proximal}. To limit these extreme gradient contributions, truncated importance sampling (TIS) caps token weights at the cost of additional clipping bias~\citep{ionides2008truncated,minimax2025m1}. Yet omitting or truncating the weights also leaves a specific problem: parts of the reward that do not distinguish between actions can still drive the update.

To see why, consider the learner's score (gradient of its log-probability). Its mean is zero under the learner's own action distribution, so a reward baseline independent of the action makes no contribution to the gradient when sampling from the learner. By contrast, under the behavior policy, that mean need not be zero, allowing the same baseline to contribute to an update without full importance weighting. Score centering (SC) removes this contribution by subtracting the mean of the score under the behavior policy, or the mean of the weighted score under the behavior policy when combined with TIS~\citep{marek2026scorecentering}.
This mean is taken over actions at the same history, and exact centering therefore cancels the unwanted contribution in expectation. In practice, Monte Carlo estimates approximate this conditional mean using auxiliary actions drawn at that history.
However, removing the baseline contribution does not remove every difference between behavior data and learner responses, because the remaining token advantages still depend on continuations generated by $\mu$. Even with a single terminal reward, a token can have different value when the learner completes the response differently. Although TIS limits the weight of the current token, it does not correct the distribution of the preceding history or the subsequent continuation.
As the learner changes, improving the surrogate on behavior trajectories can therefore fail to improve the reward obtained from the learner's own trajectories.

This dependence on behavior continuations motivates limiting how far the learner moves from the behavior policy. We introduce Regularized Score Centering (RSC), a family of methods that add reverse KL regularization to SC: \textbf{RSC = SC + KL (+TIS)}. The family regularizes both SC without IS and SC+TIS~\citep{marek2026scorecentering}. At each history, the KL penalty compares the learner's next-token distribution with the behavior policy's next-token distribution and encourages them to stay close. Because the behavior policy serves as the KL reference, refreshing it also updates the reference used by the penalty.
RSC thus preserves SC's baseline correction while limiting movement away from the policy used to estimate token advantages. The same token-level regularizer applies to both choices of reward weights and requires no additional environment interactions.

To explain how regularization can help, our analysis connects accurate fitting of the regularized surrogate to improvement in expected reward. The two RSC variants use different proof arguments, but both turn accurate fitting into reward progress. Repeated fitting gives a final policy whose reward from a single response nearly matches the reward obtained by best-of-$N$ sampling from the initial policy. Section~\ref{sec:theory-summary} explains these results, with proofs and sample bounds in Appendix~\ref{sec:appendix-theory}.

We then test whether this additional control helps when behavior data become less representative of the learner. Both agentic interaction and sampler staleness can amplify the consequences of policy mismatch. In agentic tasks, choosing a different action can change later observations, the actions available afterward, and the eventual reward, so a token's advantage under behavior continuations may poorly represent its advantage under learner continuations. Staleness introduces the same difficulty by allowing more learner updates while the behavior policy remains fixed, changing both action probabilities and the continuations they induce. Section~\ref{sec:math-results} examines these two effects through task structure and sampler refresh intervals. In these comparisons, RSC achieves substantially higher rewards than unregularized SC in agentic environments, while increasing the refresh interval weakens SC on Countdown and math tasks with mixture-of-experts (MoE) models. On Countdown, these failures persist even with Monte Carlo centering.

Beyond these controlled comparisons, we evaluate RSC on mathematical reasoning with dense and MoE models, and on agentic tasks spanning online shopping, tool use, games, and fully asynchronous software engineering. RSC with TIS attains substantially higher training rewards on Crafter and Craftax for the same number of sampled responses (Figure~\ref{fig:agentic-training}). In final evaluations, RSC with TIS achieves top performance across all six agentic tasks against baselines including IS-PPOClip, which combines importance sampling with PPO-style clipping. It raises AppWorld's mean evaluation score by $14.3\%$ over SC and SC+TIS, and reaches $9.9\times$ the score of the strongest BabyAI baseline, IS-PPOClip.

\section{Related Work}\label{sec:related}
\textbf{Score centering and off-policy learning.} RL for language models creates a mismatch between the learner and behavior policy through asynchronous generation, response reuse and different numerical implementations for training and inference~\citep{fu2025areal,yao2025rolloutmismatch}. Engineering solutions align training and inference: kernels with batch invariance remove dependence on batch shape from normalization, matrix multiplication and attention~\citep{he2025nondeterminism}, while consistent FP16 computation reduces rounding discrepancies~\citep{qi2025fp16}. Alongside these engineering solutions, algorithmic methods correct updates computed from stale data. For example, IMPALA uses importance correction in asynchronous learning~\citep{espeholt2018impala}, and \citet{hilton2022batch} separate the sampling policy from the proximal policy used to control updates. AReaL and AsyncWebRL adopt this separation, combining importance weights for data from earlier policies with PPO-style clipping around a more recent policy~\citep{schulman2017proximal,fu2025areal,bai2026asyncwebrl}. We include this combination as the IS-PPOClip baseline. TIS caps token importance weights~\citep{ionides2008truncated,yao2025rolloutmismatch,minimax2025m1}, IcePop masks token corrections outside an interval~\citep{ling2025everystep}, Sequence MIS~\citep{liu2025speed} combines sequence-level importance weights with a sequence-level mask, and GSPO clips a length-normalized sequence ratio~\citep{zheng2025gspo}. Adaptive Layerwise Perturbation (ALP) introduces learnable perturbations into each layer's input hidden states and uses the perturbed learner in the numerator of the importance ratio to address policy staleness and mismatch between training and inference~\citep{ye2026adaptive}. SC adds a different operation: subtracting the expectation of the learner score under the behavior policy, including the expectation of the weighted score when composed with TIS or MIS~\citep{marek2026scorecentering}. Exact centering removes the expected contribution of a reward baseline independent of the action. RSC preserves this correction and penalizes movement away from the behavior policy.

\textbf{Policy optimization with KL regularization.} Alongside corrections to the sampled update, KL regularization limits policy movement relative to a reference distribution. Relative Entropy Policy Search and TRPO constrain distributional change~\citep{peters2010relative,schulman2015trust}, with regularized MDP theory linking such updates to Bellman operators and mirror descent~\citep{geist2019regularized}. In language model alignment, RLHF penalizes deviation from a supervised reference~\citep{stiennon2020summarize,ouyang2022training}, and DPO uses the closed-form optimizer of a KL-regularized objective to learn from preference pairs~\citep{rafailov2023direct}. Theory for preference learning provides guarantees for finite samples in offline, online and hybrid settings~\citep{xiong2024iterative}, including general preference oracles~\citep{ye2024online}. Divergence choice and data coverage also shape learning: $\chi^2$ preference optimization addresses overoptimization of reward models~\citep{huang2025correcting}, and the coverage principle relates pretraining coverage to posttraining success~\citep{chen2025coverage}. \citet{huang2026theory} study coverage collapse and establish convergence through KL regularization and resets of the reference policy, treating complete responses as actions and using a sequence-level importance ratio. At the token level, Future-KL propagates subsequent tokens' regularization costs into the update for the current token~\citep{yao2026futurekl}, while Proximal KL combines credit assignment for later tokens with token-level importance sampling and a refreshed proximal reference policy~\citep{bai2026proximal}. RSC applies token-level reverse KL directly to the score centering objective, using the behavior policy as its reference. The same penalty applies to SC without IS and SC+TIS, preserving their correction for the reward baseline's contribution.

\section{Methodology}\label{sec:method}

\subsection{Background: Score Centering}\label{sec:sc-background}
\label{sec:setup}
We optimize the learner $\pi_\theta$ using responses generated by the behavior policy $\mu$. The gradient $\nabla_\theta$ differentiates through $\pi_\theta$ while holding $\mu$ fixed. We draw a prompt $x\sim\mathcal D$, where $\mathcal D$ is the prompt distribution, and represent its response as $Y=(a_1,\ldots,a_H)$, with corresponding history $h_t=(x,a_{<t})$. The learner maximizes its expected reward, $J(\pi_\theta)=\E_{x\sim\mathcal D,\,Y\sim \pi_\theta(\cdot\mid x)}R(x,Y)$.
Here $\mu$ denotes the policy that generated the training data. With periodic sampler refreshes, it remains fixed between refreshes. To express token-level updates, we define the learner score $\psi_\theta$ and token importance ratio $\rho$ by
\begin{equation}\label{eq:method-score-ratio}
 \psi_\theta(h,a):=\nabla_\theta\log \pi_\theta(a\mid h),
 \qquad
 \rho(h,a):=\frac{\pi_\theta(a\mid h)}{\mu(a\mid h)}.
\end{equation}
The index $i$ labels responses and $t$ labels token positions, so $\psi_{i,t}$ and $\rho_{i,t}$ are the score and ratio at history $h_{i,t}$. Let $A_i$ be the detached group-centered reward advantage of response $i$, computed as specified in Appendix~\ref{sec:experimental-settings}, and write $\widehat g_{i,t}$ for the token's stochastic contribution to the ascent gradient.

\textbf{Score centering removes the baseline contribution.} The learner score has mean zero under the learner's own action distribution, but need not have mean zero under behavior $\mu$. A reward baseline can therefore contribute to the expected update computed from behavior data. At a fixed history, let $Q_t^{\mu}$ be the expected reward after choosing an action and following $\mu$, let $V_t^{\mu}$ average this value over actions from $\mu$, and let $A_t^{\mu}=Q_t^{\mu}-V_t^{\mu}$ be the token advantage~\citep{schulman2015trust}. SC~\citep{marek2026scorecentering} subtracts the behavior mean of the score to remove the $V_t^{\mu}$ contribution without learning a value function. For SC without IS and exact centering, conditioning all expectations on that history gives
\begin{equation}\label{eq:sc-drift-intuition}
\begin{aligned}
 \E_{\mu}\!\left[R(\psi_\theta-\E_{\mu}[\psi_\theta])\right]
 &=\E_{\mu}[A_t^{\mu}\psi_\theta]\\
 &=\E_{\mu}[Q_t^{\mu}\psi_\theta]
 -\underbrace{V_t^{\mu}\E_{\mu}[\psi_\theta]}_{\text{baseline-induced drift}}.
\end{aligned}
\end{equation}
Thus centering removes the common-value contribution from the expected update. The removed term is a bias in the expected update, so collecting more data alone does not remove it.

\textbf{SC and SC+TIS fit a common TIS framework.} TIS uses the effective importance ratio $\rho^{\mathrm{TIS}}:=\operatorname{clip}(\rho,c_{\mathrm{low}},c_{\mathrm{high}})$, with $0\le c_{\mathrm{low}}\le1\le c_{\mathrm{high}}$, using the CISPO clipping form with both lower and upper bounds~\citep{minimax2025m1}. Upper truncation alone sets $c_{\mathrm{low}}=0$~\citep{ionides2008truncated,yao2025rolloutmismatch}. SC without IS is the special case $c_{\mathrm{low}}=c_{\mathrm{high}}=1$, so every token receives $\rho^{\mathrm{TIS}}=1$, whereas SC+TIS uses a wider clipping interval and retains token-dependent weights. Both methods then subtract the conditional mean of the weighted score. They use the same clipping rule for the sampled reward contribution and for the conditional mean used for centering.

Both variants therefore use the following sampled token update:
\begin{equation}\label{eq:method-comparison}
 \widehat g_{i,t}^{\rm SC}
 =A_i\!\left(\rho^{\mathrm{TIS}}_{i,t}\psi_{i,t}
 -\widehat{\E}_{\mu}[\rho^{\mathrm{TIS}}\psi_\theta\mid h_{i,t}]\right).
\end{equation}
The conditional mean $\widehat{\E}_{\mu}$ is estimated using auxiliary actions sampled at the same history, each with its own weight. Clipping an auxiliary action can therefore change the center even if the sampled action is unclipped. When all action ratios are unclipped, however, the score weighted by full IS has zero conditional mean, so subtracting the exact center leaves the expected update unchanged. We use Monte Carlo centers as described in Section~\ref{sec:countdown-mc}. Appendix~\ref{app:group-centering} gives the conditions on the group reward baseline.

\subsection{Introducing KL Regularization to Score Centering}\label{sec:rsc-method}

\textbf{SC removes the baseline contribution but retains the behavior policy's continuation values.} For updates weighted by response rewards, token-level TIS uses $Q_t^{\mu}$ in its conditional expected update. With exact centering, SC replaces this value by $A_t^{\mu}=Q_t^{\mu}-V_t^{\mu}$. The superscript $\mu$ identifies the continuation policy: this token advantage compares the expected reward after choosing the token and continuing under $\mu$ with the average reward at that history. The learner advantage $A_t^{\pi_\theta}$ makes the same comparison under learner continuations. The same token at the same history can therefore have different advantages under the two policies, even with a single terminal reward. These are conditional expected values, while $A_i$ is the sampled response advantage reused across tokens in the estimator. Centering also leaves the distribution of sampled histories unchanged. SC's update uses $A_t^{\mu}$ on behavior histories, while the learner's reward gradient uses $A_t^{\pi_\theta}$ on learner histories, motivating reverse KL to control policy mismatch during fitting.

\textbf{RSC uses reverse KL $\KL(\pi_\theta\|\mu)$ to control this mismatch during fitting.} At each sampled history, the penalty discourages separation between the learner and behavior distributions over next tokens, with $\mu$ held fixed. The regularized objective balances the improvement predicted by $A_t^{\mu}$ against the cost of policy movement. It therefore penalizes the changes that can make these fixed continuation values less representative of learner outcomes. In the outer loop of our analysis, refreshing $\mu$ from the fitted learner aligns the policies and updates the histories and token advantages that define the next surrogate. KL then controls the separation that develops during the next inner fit. Section~\ref{sec:theory-summary} shows that sufficiently accurate fitting of the regularized objective yields improvement in true reward.

In experiments with periodic sampler refreshes, we replace $\mu$ with a newer learner checkpoint at each refresh, so the KL reference changes along with the sampling policy. In fully asynchronous training, a batch can contain data from different checkpoints, and $\mu$ denotes the corresponding behavior policy for each sample. In both settings, we hold $\mu$ fixed when differentiating the objective. With regularization strength $\beta\ge0$, subtracting $\beta\KL(\pi_\theta\|\mu)$ encourages the two distributions over the next token to remain close. Applying this penalty to the two choices of reward weights in Section~\ref{sec:sc-background} gives \MethodCurrentKL{} and \MethodTISCurrentKL{}. At a fixed history $h$, the penalty is
\begin{equation}\label{eq:method-kl-objective}
 \begin{aligned}
 \KL({\pi_\theta}(\cdot\mid h)\|{\mu}(\cdot\mid h))
 &:=\sum_a{\pi_\theta}(a\mid h)
             \log\frac{{\pi_\theta}(a\mid h)}{{\mu}(a\mid h)}\\
 &=\E_{a\sim{\pi_\theta}(\cdot\mid h)}[\log \rho(h,a)]
  =\E_{a\sim {\mu}(\cdot\mid h)}[\rho(h,a)\log \rho(h,a)].
 \end{aligned}
\end{equation}
The sum runs over next tokens, with $\mu(a\mid h)>0$ wherever $\pi_\theta(a\mid h)>0$. The expectation under the behavior policy lets us estimate the penalty from the same data as the reward term.

To combine this penalty with centered reward updates, let $a(\rho)=\rho^{\mathrm{TIS}}$ and $F_a(\rho)=\int_1^\rho a(u)/u\,du$. This choice gives $\nabla_\theta F_a(\rho)=a(\rho)\psi_\theta$, with $F_a(\rho)=\log\rho$ for SC without IS. Holding the behavior policy $\mu$ and the token advantages $A_t^{\mu}$ fixed, we write $\rho_t=\pi_\theta(a_t\mid h_t)/\mu(a_t\mid h_t)$ and define the objective as
\begin{equation}\label{eq:method-regularized-target}
 \mathcal W(\pi_\theta)=\E_{x\sim\mathcal D,\,Y\sim \mu(\cdot\mid x)}
 \left[\sum_{t=1}^{H}
 \underbrace{A_t^{\mu}(h_t,a_t)F_a(\rho_t)
 -\beta(\rho_t\log\rho_t-\rho_t+1)}_{w_t}
 \right].
\end{equation}
Here $w_t$ is the token contribution to the objective. Its first term rewards actions according to $A_t^{\mu}$ on behavior histories. Its second term penalizes policy separation: since $\E_{\mu}[\rho_t\mid h_t]=1$, it has conditional expectation $\beta\KL(\pi_\theta(\cdot\mid h_t)\|\mu(\cdot\mid h_t))$. Combining these terms defines the regularized surrogate whose population gradient is
\begin{equation}\label{eq:method-regularized-surrogate-gradient}
 \nabla_\theta\mathcal W(\pi_\theta)
 =\E_{\mu}\!\left[\sum_{t=1}^H
 \left(A_t^{\mu}(h_t,a_t)\rho_t^{\mathrm{TIS}}
 -\beta\rho_t\log\rho_t\right)\psi_\theta(h_t,a_t)\right].
\end{equation}
The expectation averages over prompts and responses generated by the behavior policy. Appendix~\ref{sec:surrogate-details} gives the derivative identities.

More generally, the KL reference can be chosen separately from the policy that generates the data. Writing $\pi_{\mathrm{ref}}$ for this reference, the penalty and sampling ratio are
\[
 \KL\!\left(\pi_\theta(\cdot\mid h)\|\pi_{\mathrm{ref}}(\cdot\mid h)\right),
 \qquad \rho(h,a)=\frac{\pi_\theta(a\mid h)}{\mu(a\mid h)}.
\]
The reference is held fixed within each stage and refreshed from a learner checkpoint between stages. It can be refreshed less often than the behavior policy $\mu$, while importance weighting continues to use the ratio of learner to behavior probabilities. We take $\pi_{\mathrm{ref}}=\mu$ throughout this paper to keep the method and analysis simple.

\textbf{Estimating the combined gradient.} The population gradient in Equation~\eqref{eq:method-regularized-surrogate-gradient} uses expected continuation rewards through $A_t^{\mu}$, which are not observed directly. RSC instead uses the rewards of sampled responses. Score centering removes their common-value contribution with exact centers or conditionally unbiased independent estimates, so the expected reward update is the same as if the token advantages were available. The estimator below in Equation~\eqref{eq:rsc-total-gradient} combines the group-centered rewards $A_i$ with auxiliary actions sampled at the same history. We also allow the penalty weight to be capped independently of the reward weight: $\rho^{\mathrm{KL}}=\min(\rho,c_b)$, with $c_b\in[1,\infty]$. Both effective ratios are computed from the original $\rho$. An infinite cap gives standard reverse KL, whereas a finite cap limits contributions from large ratios. The sampled token update is
\begin{equation}\label{eq:rsc-total-gradient}
 \widehat g_{i,t}^{\rm RSC}
 =\underbrace{A_i\!\left(\rho^{\mathrm{TIS}}_{i,t}\psi_{i,t}
       -\widehat{\E}_{\mu}[\rho^{\mathrm{TIS}}\psi_\theta\mid h_{i,t}]\right)}_{\text{centered reward gradient}}
 -\beta\underbrace{\widehat{\E}_{\mu}\!\left[
       \rho^{\mathrm{KL}}\log\rho^{\mathrm{KL}}\,\psi_\theta\mid h_{i,t}\right]}_{\text{KL penalty gradient}}.
\end{equation}
For an uncapped penalty ($c_b=\infty$), draw a prompt and $G\ge2$ independent behavior responses. With the corrected group rewards and unbiased conditional centers of Proposition~\ref{prop:sc-kl-sgd},
\begin{equation}\label{eq:rsc-sgd-identity}
 \E\!\left[\frac1G\sum_{i=1}^G\sum_{t=1}^H
       \widehat g_{i,t}^{\rm RSC}\right]
 =\nabla_\theta\mathcal W(\pi_\theta).
\end{equation}
The expectation includes the prompt, responses, and auxiliary action samples. This identity shows that RSC performs stochastic gradient ascent on $\mathcal W$ in Equation~\eqref{eq:method-regularized-target}. With exact conditional centers and the penalty estimated from generated actions, the proposition also gives a gradient identity for the fixed empirical objective when data are reused. A finite penalty cap uses the modified objective in Equation~\eqref{eq:sc-kl-objective}. Appendix~\ref{sec:experimental-settings} gives the sampling procedure and numerical settings.

\subsection{Other Compared Algorithms}\label{sec:algorithms}
\textbf{Token weighting: IS and TIS.} The uncentered baselines use the sampled advantage $A_i$ from Section~\ref{sec:sc-background}, with loss averaging specified for each task in Appendix~\ref{sec:experimental-settings}. \MethodNaiveOnly{} applies the uncapped token ratio to correct the distribution over the next action at a sampled history~\citep{owen2013montecarlo}:
\begin{equation}\label{eq:method-naive-is}
 \widehat g_{i,t}^{\rm IS}=A_i\rho_{i,t}\psi_{i,t}.
\end{equation}
To bound these importance weights, our TIS baseline uses the CISPO update~\citep{minimax2025m1}, which clips token importance weights~\citep{ionides2008truncated,yao2025rolloutmismatch}. This clipping introduces a bias relative to uncapped IS:
\begin{equation}\label{eq:method-tis-only}
 \widehat g_{i,t}^{\rm TIS}
 =A_i\rho_{i,t}^{\mathrm{TIS}}\psi_{i,t}.
\end{equation}
SC+TIS adds conditional score centering to these weighted updates. Treating the clipped weights as constants during differentiation preserves the score gradient. Appendix~\ref{sec:hyperparameters} specifies the bounds.

\textbf{IS-PPOClip: importance sampling with PPO-style clipping.} This baseline combines token-level importance correction with PPO-style loss clipping around a separately refreshed learner snapshot $\pi_{\mathrm{prox}}$~\citep{schulman2017proximal,hilton2022batch,fu2025areal,bai2026asyncwebrl}. Suppressing the shared token and history arguments, its token loss is
\[
 -\frac{\pi_{\mathrm{prox}}}{\mu}
 \min\!\left\{
 A_i\frac{\pi_\theta}{\pi_{\mathrm{prox}}},\,
 A_i\operatorname{clip}\!\left(
 \frac{\pi_\theta}{\pi_{\mathrm{prox}}},c_{\mathrm{low}},c_{\mathrm{high}}\right)
 \right\}.
\]
Both $\mu$ and $\pi_{\mathrm{prox}}$ are held fixed during differentiation. Without active clipping, the two ratios combine to give the IS update in Equation~\eqref{eq:method-naive-is}. PPO clipping acts on $\pi_\theta/\pi_{\mathrm{prox}}$: once its clipped branch is active, the token's reward gradient is zero. TIS instead retains a score gradient weighted by the capped ratio. Appendix~\ref{sec:settings-agentic} specifies the clipping bounds and snapshot refresh interval.

\textbf{Sequence MIS: masked sequence-level importance sampling.} Sequence MIS~\citep{liu2025speed} assigns one importance weight and one acceptance decision to the entire response. The original rule accepts responses whose ratio is below an upper threshold, whereas our controlled experiments use an acceptance interval with lower and upper bounds. For response $i$, let $\mathcal G_i$ contain the positions of its generated tokens and write $\rho_i^{\mathrm{seq}}$ for the ratio of the complete response:
\begin{equation}\label{eq:algorithm-sequence-is}
 \rho_i^{\mathrm{seq}}=\exp\!\left(\sum_{t\in\mathcal G_i}\log \rho_{i,t}\right),
 \qquad
 \widehat g_{i,t}^{\text{Seq-MIS}}
 =A_i\rho_i^{\mathrm{seq}}\mathbf1\{c_{\mathrm{low}}\le\rho_i^{\mathrm{seq}}\le c_{\mathrm{high}}\}\psi_{i,t}.
\end{equation}
Every generated token shares the same detached response weight, and accepted responses use the unmodified response ratio. A ratio outside the acceptance interval removes the whole response's contribution to the reward gradient, while its tokens remain in the loss normalization. Appendix~\ref{sec:experimental-settings} gives the bounds.

\suppressfloats[t]
\section{Analysis of Regularized Score Centering}\label{sec:sc-analysis}

\subsection{Surrogate Objectives and Empirical Updates}\label{sec:surrogate-analysis}\label{sec:logit-diagnostic-results}
\begin{figure}[ht]
\centering
\includegraphics[width=\linewidth]{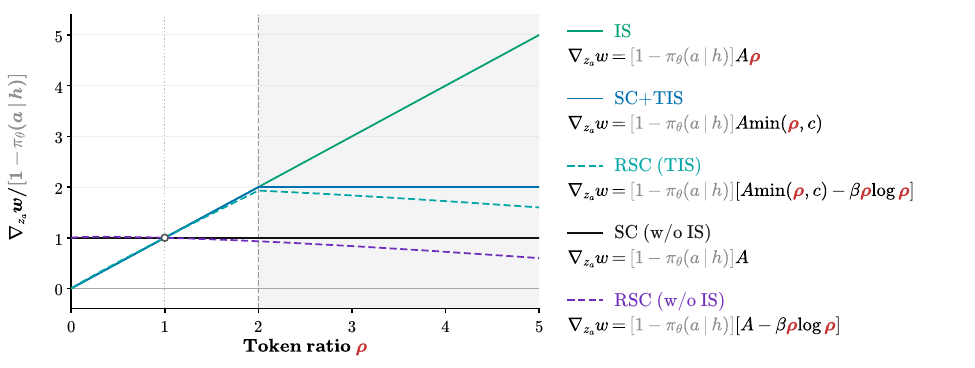}
\caption{\textbf{Weighting and KL shape logit gradients.} Curves show the derivative coefficient after factoring out $1-\pi_\theta(a\mid h)$, with $A=1$, $c=2$ and $\beta=0.05$ for RSC. The expressions on the right give $\nabla_{z_a}w$ for fixed $\mu$, $A=A^{\mu}(h,a)$ and standard, uncapped reverse KL.}
\label{fig:regularized-potentials}
\label{fig:weighting-regularization}
\end{figure}

\textbf{Token contributions to the regularized surrogate.} Equation~\eqref{eq:method-regularized-target} defines the regularized objective $\mathcal W(\pi_\theta)$. We first examine its token contributions to understand how weighting and KL affect the update.
At a fixed history, write $w$ for the token contribution $w_t$ in that objective. Table~\ref{tab:surrogate-methods} lists these contributions for each method, and Figure~\ref{fig:regularized-potentials} shows the resulting logit derivatives.

Equation~\eqref{eq:rsc-sgd-identity} shows how the sampled update optimizes $\mathcal W$. This objective uses histories and continuation values from $\mu$, whereas the reward $J(\pi_\theta)$ is evaluated on responses generated by $\pi_\theta$. Section~\ref{sec:theory-summary} explains how accurate fitting of $\mathcal W$ can improve $J(\pi_\theta)$.

\begin{figure}[!t]
\centering
\includegraphics[width=\linewidth]{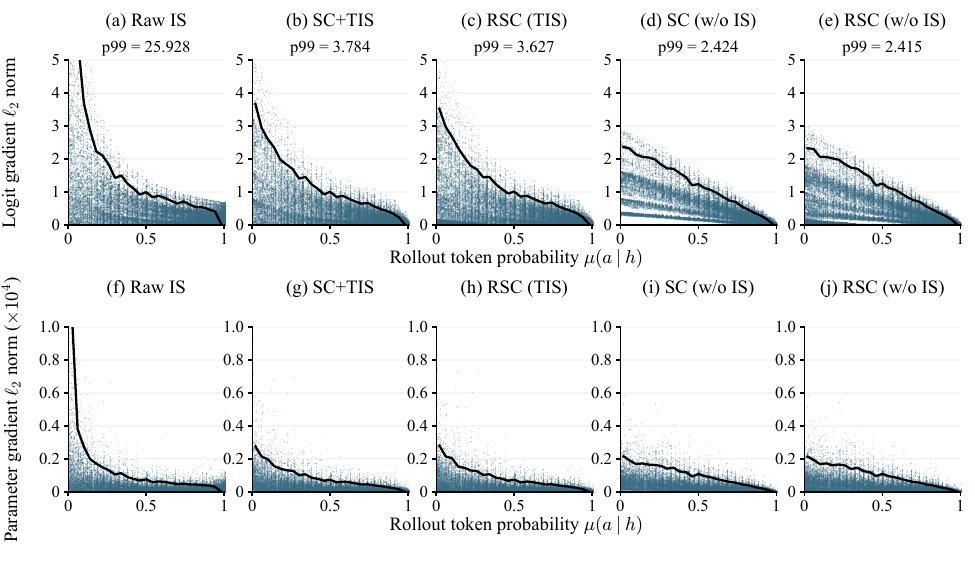}
\caption{\textbf{Weighting, centering and KL shape the update.} We hold the learner and behavior checkpoints fixed and reuse the sampled actions, response advantages, and 128 auxiliary Monte Carlo draws across methods. The panels therefore compare update rules on the same data, before any optimizer step. Both rows use behavior probability $\mu(a\mid h)$: panels (a--e) show logit gradient norms, and panels (f--j) show parameter gradient norms. Black lines show the 99th percentile within each probability bin, and the top row reports this percentile for $\mu<0.01$. Appendix~\ref{sec:gradient-diagnostic-settings} gives the diagnostic protocol.}
\label{fig:logit-diagnostic}
\end{figure}

\textbf{SC+TIS limits gradient amplification by importance ratios, and KL damps large positive updates.} The integrands in Equation~\eqref{eq:method-regularized-target} let us examine how each method weights an individual token. IS has $\nabla_{z_a}w=A\rho[1-\pi_\theta(a\mid h)]$, so a large ratio between learner and behavior probabilities can amplify the gradient. SC+TIS replaces $\rho$ with $\min(\rho,c)$, bounding the reward multiplier, while SC without IS fixes it at one. For $\rho>1$, RSC subtracts $\beta\rho\log\rho[1-\pi_\theta(a\mid h)]$, opposing further probability increases (Figure~\ref{fig:regularized-potentials}). Since $\rho=\pi_\theta/\mu\le1/\mu$, very large ratios require small behavior probabilities, motivating $\mu(a\mid h)$ as the horizontal axis in Figure~\ref{fig:logit-diagnostic}. The matched comparison in Figure~\ref{fig:logit-diagnostic} shows smaller extreme gradient norms under SC and SC+TIS in both logit and parameter space. Adding KL further reduces the 99th percentile of logit gradient norms for tokens with low behavior probability under both variants.

\subsection{Why Regularization Makes Fitting More Informative}\label{sec:theory-summary}
\begingroup
\setlength{\abovedisplayskip}{3pt}
\setlength{\belowdisplayskip}{3pt}
\setlength{\abovedisplayshortskip}{0pt}
\setlength{\belowdisplayshortskip}{3pt}
Beyond its effect on individual updates, KL helps in two ways: it makes accurate fitting easier to certify from data, and it lets fitting accuracy translate into reward progress without shrinking the token radius as responses get longer. Appendix~\ref{sec:theory-setup} gives the proofs and sample bounds.

\textbf{The improving target.} Consider \MethodTISCurrentKL{} with deterministic transitions and binary terminal rewards. Fix a prompt and behavior policy $\mu$, and write $\mathcal W_{\mu}$ for the objective in Equation~\eqref{eq:method-regularized-target} conditional on this prompt. In this target comparison, $J$ is also conditional on the prompt. For $\beta\ge2$, take the full policy family satisfying $|\log\rho_t|\le1/\beta$ at every history and action. Reward and penalty caps of two are inactive, and the maximizing policy is
\begin{equation}\label{eq:summary-improving-target}
 \pi^\star(a\mid h_t)=
 \frac{\mu(a\mid h_t)\exp\!\left(A_t^{\mu}(h_t,a)/\beta\right)}
 {\E_{u\sim \mu(\cdot\mid h_t)}\exp\!\left(A_t^{\mu}(h_t,u)/\beta\right)}.
\end{equation}
The target assigns more weight to actions with higher advantage and belongs to the allowed policy family. It multiplies success odds by at least $1+1/\beta$ for $0<J(\mu)<1$ (Lemmas~\ref{lem:rsc-target-gap} and~\ref{lem:rsc-target-progress}).

\textbf{Inner fitting.} Without KL, changes in token probabilities can leave the surrogate unchanged at histories where all token advantages are zero. A small objective gap then need not identify a particular target policy. For the regularized TIS objective in Equation~\eqref{eq:method-regularized-target}, KL adds strong concavity in the token probabilities, so moving away from the target incurs an objective loss. Every~candidate~$\pi_\theta$ in the family satisfies
\begin{equation}\label{eq:summary-target-gap}
 \mathcal W_{\mu}(\pi^\star)-\mathcal W_{\mu}(\pi_\theta)
 =\beta\sum_t\E_{h_t\sim d_t^{\mu}}
 \KL\!\left(\pi_\theta(\cdot\mid h_t)\|\pi^\star(\cdot\mid h_t)\right).
\end{equation}
Here $d_t^{\mu}$ is the distribution of histories at position $t$ under $\mu$. The identity shows that a small objective gap makes the fitted policy close to the improving target on behavior histories. This closeness also controls sampling noise: the variance of the sampled objective difference between a candidate and the target is bounded by a constant times $H$ times the population gap (Lemma~\ref{lem:rsc-response-moments}). As fitting improves, this comparison becomes less noisy, giving a sharper sample bound than treating the noise as constant. This statistical use of KL curvature is analogous to the fast-rate analysis of \citet{zhao2025sharp} for KL-regularized contextual bandits and RLHF.

\textbf{Outer progress.} To relate this fitting guarantee to reward, we next examine changes in the history distribution. Without KL, our analysis controls these changes by bounding each token log-ratio. For a response of $H$ tokens, a bound $|\log\rho_t|\le r$ allows the response ratio to be as large as $e^{Hr}$. At a fixed radius, this worst-case bound grows exponentially with $H$. Keeping it bounded using the token constraint alone therefore requires $r=O(1/H)$ (Lemma~\ref{lem:token-radius-sequence-ratio}). For fixed $\beta$, the KL-regularized target $\pi^\star$ instead has a worst-case bound on every prefix and full-response ratio that is independent of $H$, even at the constant token radius $r=O(1)$ (Lemma~\ref{lem:rsc-prefix-coverage}). With this coverage bound, a policy that fits the surrogate accurately achieves reward close to the improving target (Lemma~\ref{lem:rsc-gap-transfer}). RSC without IS also turns accurate fitting into reward progress, as shown in Appendix~\ref{app:chenlu-pure-sc-kl}.

Together, the inner fitting bounds and outer progress bounds give a comparison with best-of-$N$ sampling: one response from the final policy nearly matches the expected reward of the best of $N$ independent responses from the initial policy. On the finite deterministic token tree with binary rewards, both RSC variants achieve reward error at most $\epsilon$ with high probability after $O(\log(N/\epsilon))$ outer rounds under the stated fitting and sampling conditions. These bounds use $\beta=2$, a token log-ratio radius of $1/2$, and fresh data at each round. Without KL, the sufficient radius is $1/(8H)$ and the round bound is $O(H\log(N/\epsilon))$, where $H$ is the response horizon. Theorems~\ref{thm:sc-kl-progress} and~\ref{thm:chenlu-sc-kl-convergence} give the regularized guarantees, and Theorem~\ref{thm:sc-sample-complexity} gives the unregularized comparison.

In these guarantees, each outer round fits a policy on a fixed dataset before refreshing the behavior policy. These counts therefore measure policy refreshes, while optimizer iterations and sampling costs remain separate. The TIS theorem uses exact centers, while the theorem for unit weights also allows Monte Carlo centers. Appendix~\ref{sec:theory-setup} gives the sample bounds that depend on the horizon and discusses the experimental KL coefficients.
 
These guarantees concern score-centered updates. KL regularization encourages the learner to remain close to
  the behavior policy, but does not ensure that every token
  ratio stays within the TIS bounds. When clipping is active,
  the weighted score can have a nonzero conditional mean under
  the behavior policy, so a reward baseline can contribute to
  the expected update. Score centering subtracts this mean and
  removes the baseline contribution even with clipped reward
  weights, as shown in Section~\ref{sec:sc-background}. This
  cancellation is essential to our analysis. In other words, our analysis does not
  establish reward improvement for TIS+KL without score
  centering when clipping is active.

\endgroup

\FloatBarrier
\section{Identifying Failure Modes of Score Centering}\label{sec:countdown}\label{sec:countdown-reproduction}

The reward guarantees in Section~\ref{sec:theory-summary} rely on accurate fitting within a specified policy neighborhood. In training, both the score center estimate and movement away from the behavior policy can affect learning. We examine these two potential sources of SC failure in turn. Section~\ref{sec:countdown-mc} shows that Monte Carlo centering improves learning over approximate centers. These improved estimates leave the dependence on behavior histories and continuations unchanged, motivating the task and staleness comparisons in Section~\ref{sec:math-results}. We first contrast SC's successful learning on math tasks using a dense model with its behavior on Crafter, then examine how SC can degrade as staleness increases on Countdown and math tasks with an MoE model.

\subsection{Experimental Setup}\label{sec:reasoning-experimental-setup}

\needspace{4\baselineskip}
In the controlled experiments in Figures~\ref{fig:score-center-estimators} through \ref{fig:agentic-training}, the sampler refresh interval controls how long the learner updates against a fixed behavior policy. The learner $\pi_\theta$ updates while $\mu$ remains fixed for $R$ updates. The sampler then copies the learner weights to refresh $\mu$. In agentic experiments, rollout generation and learner training run concurrently: the learner consumes completed response groups from a queue, taking one update per new batch while generation continues under fixed $\mu$. Countdown alternates sampling and training within the same schedule with $\mu$ fixed. Larger $R$ therefore permits greater policy separation. Training curves measure reward under $\mu$, while heldout scores evaluate $\pi_\theta$ on separate tasks. Appendix~\ref{sec:experimental-settings} specifies the models, datasets, optimization and evaluation settings.

\subsection{Score Centering Fails Under Top-k Estimation}\label{sec:countdown-mc}

\textbf{Monte Carlo gives an unbiased estimate of the conditional center.} The original SC implementation uses the 128 highest behavior probabilities and a modeled tail~\citep{marek2026scorecentering}, introducing approximation bias. We replace this biased approximation with auxiliary draws from $\mu$. At the history $h$ preceding each generated token, we draw auxiliary tokens independently with replacement from $\mu(\cdot\mid h)$. These draws are independent of the generated action and the continuation of the response, and require no additional response generation or reward evaluation. Backpropagation through the average learner log-probability of the auxiliary tokens gives an unbiased estimate of $\E_{\mu}[\psi_\theta\mid h]$. For SC+TIS, each auxiliary log-probability receives its own detached $\rho^{\mathrm{TIS}}$ weight, giving an unbiased estimate of $\E_{\mu}[\rho^{\mathrm{TIS}}\psi_\theta\mid h]$. If a token is sampled more than once, each draw contributes to the estimate. Appendix~\ref{sec:experimental-settings} lists the sample counts.

\begin{figure}[!ht]
\centering
\includegraphics[width=\linewidth]{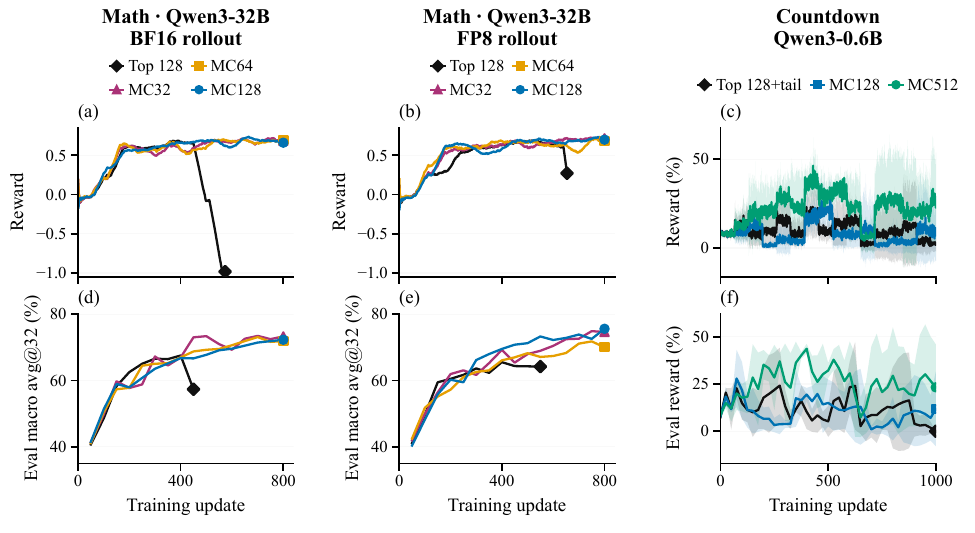}
\caption{\textbf{Monte Carlo centering improves on Top-128.} Rows show training (top) and evaluation (bottom). Countdown uses three seeds, with bands indicating sample standard deviation. The math experiments use one seed per method and rollout precision.}
\label{fig:countdown-early-policy}
\label{fig:score-center-estimators}
\end{figure}

\textbf{Monte Carlo centering improves learning over top-$k$ estimation.} Every tested Monte Carlo variant in Figure~\ref{fig:score-center-estimators} achieves a higher final evaluation score than the corresponding Top-128 baseline. The math comparisons show substantial evaluation gains with both BF16 and FP8 rollouts. Countdown shows the same advantage across the tested Monte Carlo sample counts, supporting direct sampling from the behavior policy to estimate the center.

\subsection{Score Centering Fails under Agentic and Highly Stale Settings}\label{sec:math-results}\label{sec:countdown-failure}\label{sec:moe-results}

SC's expected update weights actions by $A_t^{\mu}$, which evaluates them using continuations under the behavior policy. The learner's advantage $A_t^{\pi_\theta}$ instead uses continuations under $\pi_\theta$. At $\pi_\theta=\mu$, the advantages and history distributions agree. When the relevant action and continuation distributions remain close, advantages under the behavior policy provide a local approximation to those under the learner policy. As the policies separate, the value of the same action can change and learner trajectories can visit histories poorly represented in the behavior data. The resulting error depends on how far the policies move apart and how much their different choices change the eventual reward.

\begin{figure}[!ht]
  \centering
  \includegraphics[width=0.98\linewidth]{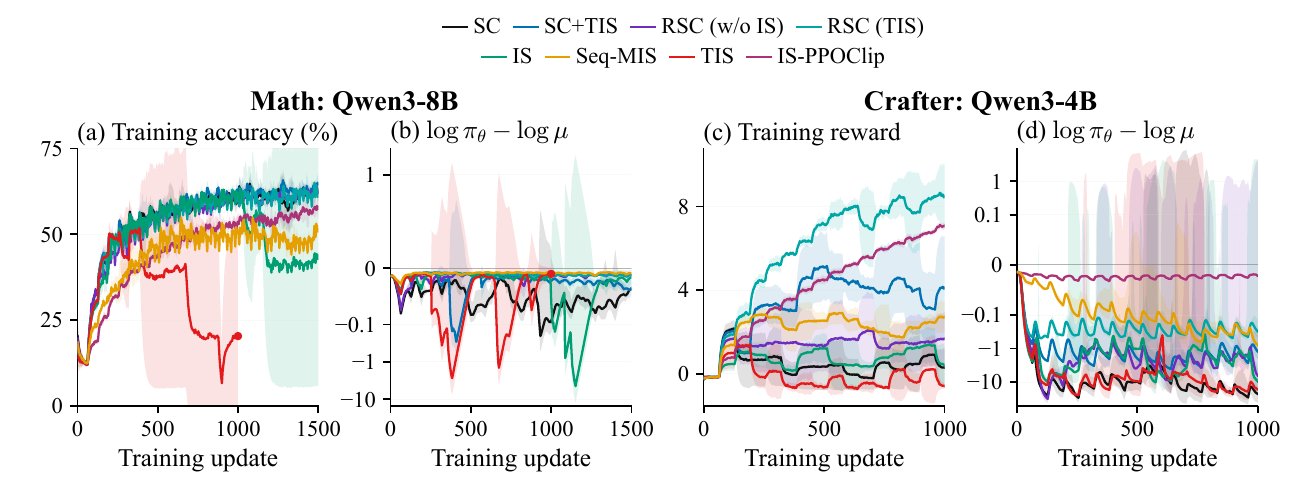}
  \caption{\textbf{Learning on math tasks and Crafter at the same refresh interval.} The left pair shows math tasks with a dense model and the right pair shows Crafter, each with training performance and signed mean $\log \pi_\theta-\log \mu$. Curves and bands show mean $\pm$ sample standard deviation across three seeds. The policy-gap metric for IS-PPOClip is available only for Crafter.}
  \label{fig:sc8b-training}
\end{figure}

\textbf{Task structure can change the consequences of policy mismatch.} At the same sampler refresh interval, Figure~\ref{fig:sc8b-training} compares math tasks using a dense model with Crafter, an agentic environment where the agent's actions produce feedback that guides subsequent decisions. SC and SC+TIS sustain learning on the math tasks, and adding KL changes each final mean accuracy by less than $1\%$ in absolute accuracy. On Crafter, however, \MethodTISCurrentKL{} achieves substantially higher training and evaluation scores than SC+TIS. Although the refresh interval is matched, the degree of separation between learner and behavior policies can still differ. Figure~\ref{fig:agentic-training} compares all six agentic tasks, with final evaluations in Table~\ref{tab:agentic-final-results}. Appendix~\ref{sec:settings-math} gives the evaluation protocol for the math tasks.

To understand how task structure can amplify this mismatch, consider agentic actions that change the environment state and the choices available afterward. In Crafter, gathering a resource or taking a route can open a different branch of future interactions. Because the value of that action depends on how the policy uses the resulting opportunities, it may have little advantage under $\mu$ if behavior continuations rarely exploit them, but a much larger advantage under $\pi_\theta$ if learner continuations do. Such branching can make $A_t^{\pi_\theta}$ and $A_t^{\mu}$ differ substantially even at the same sampler refresh interval. The math and Crafter runs differ in both task and model, but the effect of actions on later choices may help explain the contrast in Figure~\ref{fig:sc8b-training}.

\begin{figure}[!ht]
\centering
\includegraphics[width=\linewidth]{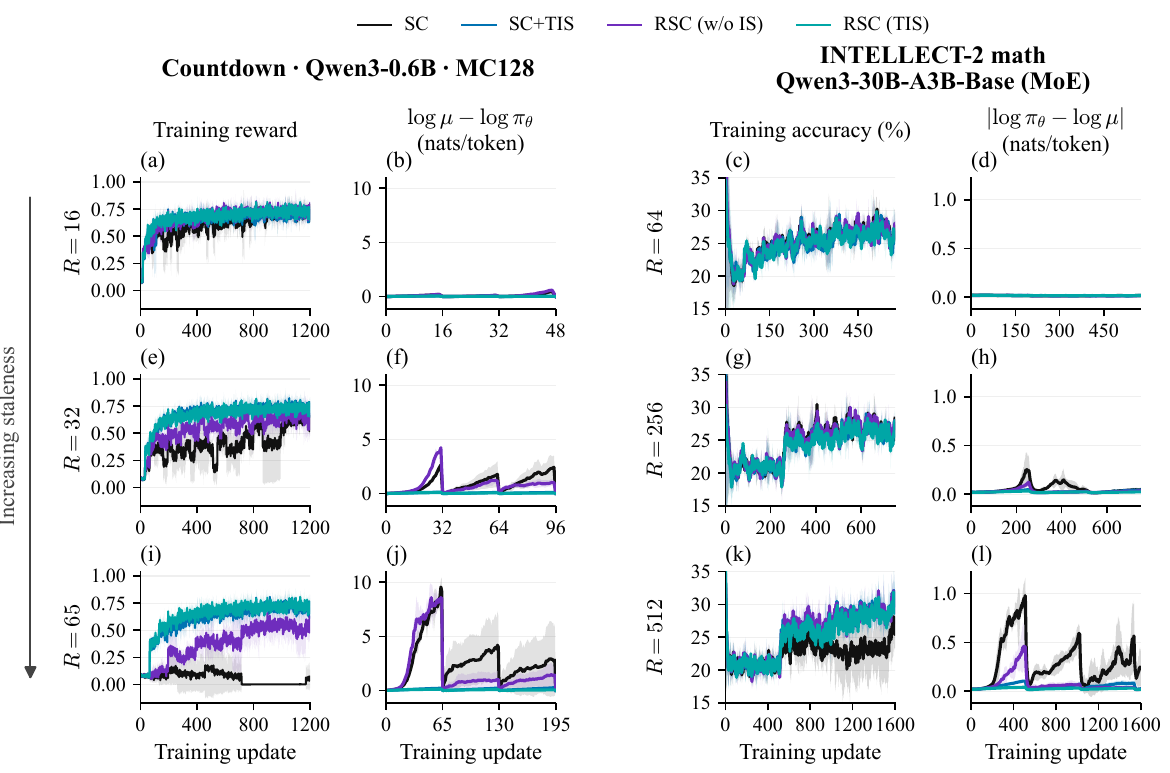}
\caption{\textbf{Score centering under increasing staleness.}
Columns show Countdown reward and signed $\log {\mu}-\log{\pi_\theta}$, then
math training accuracy with the MoE model and $|\log{\pi_\theta}-\log {\mu}|$. Sampler refresh intervals increase down the rows.
Curves and bands show mean $\pm$ sample standard deviation. Countdown uses three seeds. MoE uses two seeds at the two shorter intervals and three at the longest. Countdown log-probability panels cover the first three refresh cycles.}
\label{fig:staleness-comparison}
\label{fig:countdown-formal}
\label{fig:countdown-policy-comparison}
\label{fig:moe-formal}
\end{figure}

\textbf{Longer refresh intervals allow policy mismatch to accumulate.} Task structure can affect the consequences of different continuations, while the refresh interval controls how long the policies can move apart. With more learner updates against a fixed $\mu$, $\pi_\theta$ can change both action probabilities and continuation values further before the sampler is refreshed. Figure~\ref{fig:staleness-comparison} tests this source of error on Countdown and math tasks with an MoE model. SC remains competitive at shorter intervals, while SC+TIS and RSC retain stronger performance at the longest intervals. At the largest tested Countdown interval, \MethodTISCurrentKL{} substantially outperforms SC in final heldout reward, with an absolute reward gain of approximately $65\%$ even with Monte Carlo centers. In the experiments with the MoE model, SC exceeds SC+TIS at shorter refresh intervals but falls behind at the longest interval, alongside greater policy separation. Learner evaluations reveal degradation that rewards from the stale behavior policy can hide. Settings for Countdown and the math experiments with the MoE model appear in Appendices~\ref{sec:settings-countdown} and~\ref{sec:settings-math}, respectively.

\section{Scaling Experiments to Agentic Settings}\label{sec:experiments}
The controlled comparisons in Section~\ref{sec:countdown} show why accurate centering can still benefit from KL regularization. We now evaluate the resulting methods across a wider set of interactive tasks, using both periodic sampler refreshes and fully asynchronous data collection.

\subsection{Experimental Setup}\label{sec:empirical-protocols}
\textbf{Controlled asynchronous environments.} Figure~\ref{fig:agentic-training} uses the same environment interaction and controlled asynchronous training setup described in Section~\ref{sec:reasoning-experimental-setup} for WebShop shopping~\citep{yao2022webshop}, AppWorld application tool use~\citep{trivedi2024appworld}, BabyAI navigation and object manipulation guided by language~\citep{chevalier2019babyai}, Cooking games generated with TextWorld~\citep{cote2018textworld}, and survival in Crafter~\citep{hafner2021crafter} and Craftax~\citep{matthews2024craftax}. Appendix~\ref{sec:settings-agentic} gives task rewards, training settings and episode limits (Table~\ref{tab:settings-agentic-limits}).

\textbf{Fully asynchronous software engineering environment.} Figure~\ref{fig:swe-preliminary} studies software engineering through a fully asynchronous pipeline. Completed response groups enter a training queue as generation and tool interaction finish. Whenever a batch is ready, the learner performs an optimizer update and publishes its new weights, while generation continues for other trajectories. Policy lag therefore emerges from generation latency, environment interaction and queueing. This resembles production RL workloads, where trajectories finish at different times and training consumes data as they arrive. Appendix~\ref{sec:swe-protocol} specifies the model, datasets, sampling and optimization settings.

\needspace{5\baselineskip}
\textbf{Compared methods.} In Figures~\ref{fig:agentic-training} and~\ref{fig:swe-preliminary}, our main baselines, SC and SC+TIS, are among the strongest methods in the original score centering study~\citep{marek2026scorecentering}. IS and TIS provide baselines without score centering. Figure~\ref{fig:agentic-training} also includes \MethodISPPOClip{}, which combines token-level importance correction with PPO-style loss clipping, and Sequence MIS, which combines sequence-level importance weights with a sequence-level mask~\citep{liu2025speed}. Section~\ref{sec:algorithms} defines these methods, and Appendix~\ref{sec:experimental-settings} gives their settings and evaluation protocols.

\subsection{Controlled Asynchronous Agentic Tasks}\label{sec:agentic-results}
\label{sec:webshop-results}\label{sec:agentic-extension}

\textbf{RSC with TIS substantially improves final performance.} \MethodTISCurrentKL{} achieves top final performance across all six tasks (Table~\ref{tab:agentic-final-results}). Compared with \MethodISPPOClip{}, the strongest baseline outside the RSC family on Crafter, Craftax and BabyAI, it improves the mean score by $52\%$ on Crafter and $16\%$ on Craftax, and reaches $9.9\times$ its score on BabyAI. It also improves on SC+TIS by $43\%$ on WebShop and on SC and SC+TIS by $14.3\%$ on AppWorld, while scoring close to the maximum on TextWorld.

\begin{figure}[!ht]
\centering
\includegraphics[width=.96\linewidth]{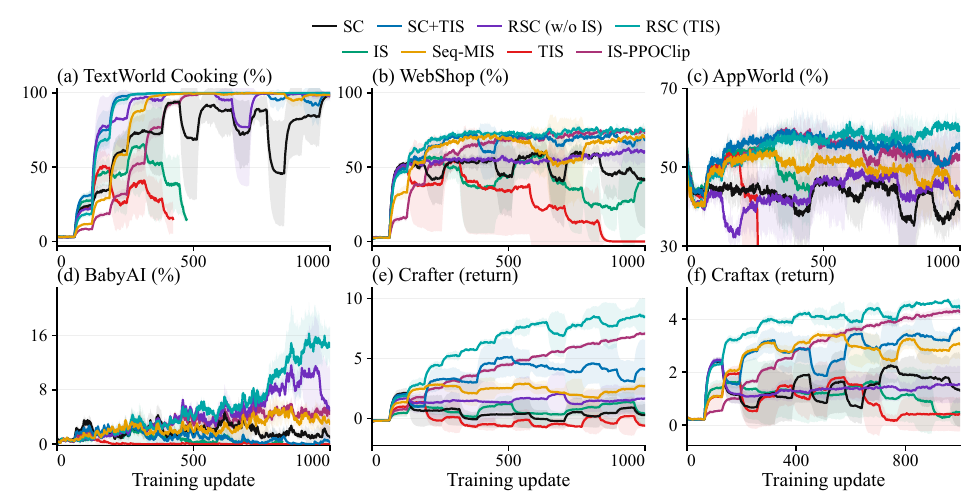}
\caption{\textbf{Training reward across six interaction tasks.}
Curves show mean $\pm$ sample standard deviation across three seeds. Crafter and Craftax use total episode rewards. Appendix~\ref{sec:settings-agentic} specifies training settings.}
\label{fig:agentic-training}
\label{fig:webshop-training}\label{fig:agentic-extension}
\end{figure}

\textbf{RSC with TIS substantially improves sample efficiency on Crafter and Craftax.} For the same number of sampled responses, RSC with TIS attains higher training rewards on both tasks, with the advantage persisting as training proceeds (Figure~\ref{fig:agentic-training}). Table~\ref{tab:agentic-final-results} complements these learning curves with final learner evaluations.

\begin{table}[!ht]
\centering
\caption{Checkpoint evaluation scores (mean $\pm$ sample standard deviation across three seeds). Crafter uses its geometric achievement score, Craftax uses total episode reward, and other scores are reported in \%. Bold marks the highest reported mean and the second highest mean if it is within 5\% of the highest.}
\label{tab:agentic-final-results}
\begingroup
\fontfamily{ptm}\selectfont\fontsize{10}{12}\selectfont
\setlength{\tabcolsep}{2pt}
\newcommand{\AgenticScore}[2]{#1{\fontsize{8.5}{10}\selectfont\(\pm\)#2}}
\renewcommand{\arraystretch}{1.18}
\newsavebox{\AgenticTableNatural}
\newsavebox{\AgenticTableFitted}
\newlength{\AgenticTableLimit}
\setlength{\AgenticTableLimit}{\linewidth}
\sbox{\AgenticTableNatural}{\begin{tabular}{@{}lrrrrrr>{\columncolor{black!7}}r>{\columncolor{black!7}[\tabcolsep][0pt]}r@{}}
\toprule
 & \multicolumn{4}{c}{Importance Sampling} & \multicolumn{4}{c}{Score Centering} \\
\cmidrule(lr){2-5}\cmidrule(lr){6-9}
\multicolumn{1}{@{}c}{Benchmark} & \multicolumn{1}{c}{\MethodNaiveOnly{}} & \multicolumn{1}{c}{\shortstack{Sequence\\MIS}} & \multicolumn{1}{c}{TIS} & \multicolumn{1}{c}{\MethodISPPOClip{}} & \multicolumn{1}{c}{\MethodSC{}} & \multicolumn{1}{c}{\MethodTIS{}} & \multicolumn{1}{>{\columncolor{black!7}}c}{\MethodCurrentKL{}} & \multicolumn{1}{>{\columncolor{black!7}[\tabcolsep][0pt]}c@{}}{\MethodTISCurrentKL{}} \\
\midrule
\multicolumn{9}{l}{\textbf{Final evaluation score} ($\uparrow$)} \\
\addlinespace[2pt]
WebShop & \AgenticScore{40.94}{36.13} & \AgenticScore{47.25}{40.95} & \AgenticScore{0.00}{0.00} & \textbf{\AgenticScore{74.81}{0.28}} & \AgenticScore{41.42}{35.57} & \AgenticScore{52.21}{30.06} & \AgenticScore{60.46}{9.02} & \textbf{\AgenticScore{74.80}{1.15}} \\
TextWorld Cooking & \AgenticScore{21.77}{16.42} & \AgenticScore{96.95}{3.22} & \AgenticScore{19.77}{13.45} & \textbf{\AgenticScore{99.73}{0.25}} & \AgenticScore{99.06}{0.80} & \AgenticScore{92.14}{13.62} & \AgenticScore{95.29}{4.32} & \textbf{\AgenticScore{99.95}{0.08}} \\
AppWorld dev & \AgenticScore{6.43}{11.14} & \AgenticScore{7.02}{5.26} & \AgenticScore{0.00}{0.00} & \AgenticScore{9.36}{3.65} & \AgenticScore{12.28}{1.75} & \AgenticScore{12.28}{4.64} & \AgenticScore{11.11}{3.65} & \textbf{\AgenticScore{14.04}{1.75}} \\
BabyAI & \AgenticScore{0.13}{0.13} & \AgenticScore{1.71}{1.75} & \AgenticScore{0.00}{0.00} & \AgenticScore{1.83}{1.36} & \AgenticScore{0.80}{0.95} & \AgenticScore{0.30}{0.32} & \AgenticScore{7.32}{5.80} & \textbf{\AgenticScore{18.06}{1.77}} \\
Crafter & \AgenticScore{1.02}{0.70} & \AgenticScore{1.93}{0.82} & \AgenticScore{0.57}{0.44} & \AgenticScore{9.34}{0.46} & \AgenticScore{1.12}{1.12} & \AgenticScore{3.27}{1.89} & \AgenticScore{0.94}{0.29} & \textbf{\AgenticScore{14.21}{3.40}} \\
Craftax & \AgenticScore{0.83}{0.96} & \AgenticScore{2.39}{1.43} & \AgenticScore{0.39}{0.63} & \AgenticScore{4.04}{0.27} & \AgenticScore{1.58}{1.10} & \AgenticScore{3.20}{0.58} & \AgenticScore{1.46}{0.64} & \textbf{\AgenticScore{4.71}{0.03}} \\
\bottomrule
\end{tabular}}
\ifdim\wd\AgenticTableNatural>\AgenticTableLimit
  \sbox{\AgenticTableFitted}{\resizebox{\dimexpr\AgenticTableLimit-0.1pt\relax}{!}{\usebox{\AgenticTableNatural}}}\else
  \sbox{\AgenticTableFitted}{\usebox{\AgenticTableNatural}}\fi
\ifdim\wd\AgenticTableFitted>\AgenticTableLimit
  \PackageError{sc-paper}{Agentic table exceeds its width limit}{Check the saved table box and graphicx scaling.}
\else
\fi
\usebox{\AgenticTableFitted}
\endgroup
\end{table}

\subsection{Fully Asynchronous RL on Software Engineering Tasks}\label{sec:swe-results}

\begin{figure}[!ht]
\centering
\includegraphics[width=\linewidth]{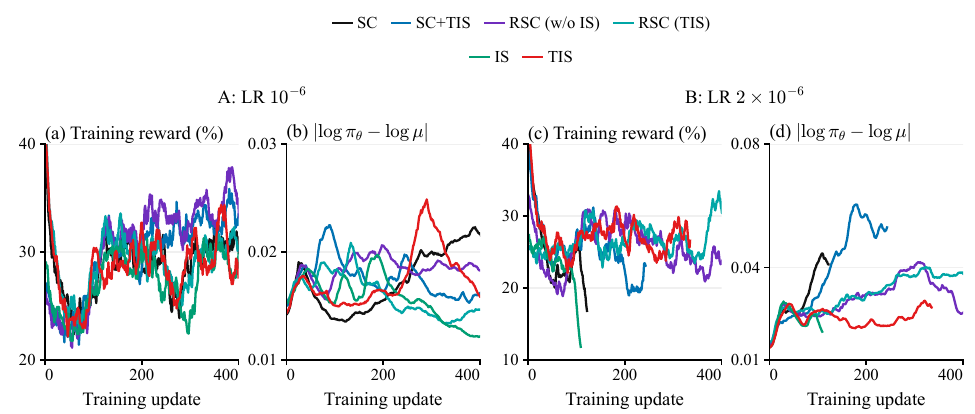}
\caption{\textbf{Fully asynchronous SWE training.}
Each method and configuration uses one seed. The left pair shows configuration A and the right pair configuration B, each with training reward followed by the absolute log-probability difference. Axes use restricted ranges. Appendix~\ref{sec:swe-protocol} specifies the configurations and smoothing.}
\label{fig:swe-preliminary}
\label{fig:swe-additional}
\vspace{4pt}
\begin{minipage}{\linewidth}
\centering
\captionof{table}{\textbf{SWE-bench Verified evaluation (Average@8, \%).}
Learning rates match the training panels. Appendix~\ref{sec:swe-protocol} gives the evaluation protocol. Best scores are bold.}
\label{tab:swe-final-results}
{\rmfamily\fontsize{8}{9.5}\selectfont
\setlength{\tabcolsep}{2.0pt}
\renewcommand{\arraystretch}{1.12}
\begin{tabular*}{\linewidth}{@{\extracolsep{\fill}}l*{6}{c}@{}}
\toprule
LR & IS & TIS & SC & SC+TIS & RSC (w/o IS) & RSC (TIS) \\
\midrule
$10^{-6}$ & 57.08 & 52.33 & 58.48 & 59.06 & \textbf{61.14} & 54.77 \\
$2\times10^{-6}$ & 0.40 & 51.51 & 21.13 & 36.04 & 47.89 & \textbf{54.47} \\
\bottomrule
\end{tabular*}}

\end{minipage}
\end{figure}

\textbf{RSC leads both SWE configurations.} An RSC variant achieves the highest evaluation score in each configuration, A and B, where policy lag varies during training. Configuration B shows substantial gains: RSC (w/o IS) and RSC (TIS) raise evaluation scores over SC and SC+TIS by approximately $127\%$ and $51\%$, respectively. Figure~\ref{fig:swe-preliminary} shows the training curves, and Table~\ref{tab:swe-final-results} reports the SWE-bench Verified scores. SWE training is computationally expensive and slow, so we evaluate a subset of baselines.

\textbf{Larger KL gains coincide with greater policy mismatch in this comparison.} To examine the difference between configurations A and B, we compare their policy lag and probability mismatch. The configurations have similar measured policy lag, defined as the number of learner updates between the policy version used for sampling and the policy version used for training. SC+TIS in B has approximately $2\times$ the mean $|\log \pi_\theta-\log \mu|$ of SC+TIS in A over the same range of training updates. Thus, similar delays can correspond to different amounts of policy mismatch. The KL gains follow this contrast: with or without TIS, RSC's improvement over its corresponding SC baseline is substantially larger in configuration B than in A, where the gain is small without TIS and negative with TIS. These are observational comparisons of single runs at different learning rates. These results suggest that a larger learning rate can increase off-policyness and that RSC can sustain improvements at higher learning rates than SC.

\section{Conclusion and Future Work}
Regularized Score Centering removes the reward baseline contribution and adds reverse KL to penalize movement away from the behavior policy. With this control of policy movement, our analysis connects accurate surrogate fitting to reward improvement. Consistent with this motivation, the experiments show benefits when data come from stale policies and in agentic tasks where actions change subsequent states and rewards. RSC with TIS substantially improves sample efficiency on Crafter and Craftax, attaining higher training rewards for the same number of sampled responses. It also achieves top final performance across all six agentic tasks. Its gains over the strongest baselines include a $14.3\%$ improvement on AppWorld and $9.9\times$ the score on BabyAI. The software engineering results extend these gains to fully asynchronous training, where policy lag varies over time.

The theoretical guarantee also allows a token log-ratio radius that does not shrink
  with the response horizon and, for unit weights, accommodates Monte Carlo estimates of the
  center. Building on these guarantees, a useful direction for future work is to study how greater
  policy staleness affects RSC empirically, including how many
  optimizer updates are needed to maintain reward improvement
  as the learner moves further from the policies that
  generated the data.

\FloatBarrier
\section*{Acknowledgements}
This work was supported in part by NSF grants 2416897, 2505932, and 2601666 and ONR grant N000142512318. Computing resources were provided by the National Center for Supercomputing Applications (NCSA) through Delta (NSF OAC-2005572) and DeltaAI (NSF OAC-2320345), as well as Illinois Computes and the NAIRR Pilot (NAIRR250157).
\bibliographystyle{plainnat}
\bibliography{references}
\FloatBarrier
\appendix
\numberwithin{equation}{section}
\numberwithin{theorem}{section}
\allowdisplaybreaks[1]
\section{Experimental Settings}\label{sec:experimental-settings}\label{sec:hyperparameters}\label{sec:evaluation-protocols}
Figures~\ref{fig:sc8b-training} through \ref{fig:agentic-training} compare methods on math and agentic tasks with periodic behavior policy updates, Figure~\ref{fig:score-center-estimators} compares estimators of the conditional mean score, and Figure~\ref{fig:swe-preliminary} studies fully asynchronous training for software engineering. Section~\ref{sec:reasoning-experimental-setup} describes controlled sampling, and Section~\ref{sec:empirical-protocols} describes fully asynchronous data collection. Final scores use each run's last evaluation.

\textbf{Shared settings for controlled experiments.}
The controlled experiments in Figures~\ref{fig:sc8b-training} through \ref{fig:agentic-training} use eight responses per prompt and
one optimizer update per batch. The main SC comparisons use 128 independent
Monte Carlo samples to estimate the conditional mean score at the history preceding each generated token. SC sets the importance weight to one, SC+TIS caps it at two, and RSC adds reverse KL regularization
with $\beta=0.01$ and a separate cap of two on the penalty ratio. SC and SC+TIS subtract the group mean
reward, and RSC additionally applies $G/(G-1)$ with $G=8$ to match the theoretical normalization. The paired gradient diagnostics in Figure~\ref{fig:logit-diagnostic} apply this correction to every method. Appendix~\ref{sec:group-size-ablation} tests its effect. Each batch loss averages over the generated tokens. Rewards are not divided by their standard deviation. Entropy bonuses and KL regularization to a fixed reference model are disabled. IS uses an uncapped token ratio, TIS
caps it at five, and Sequence MIS accepts responses whose ratio of response probabilities lies in $[0.1,10]$. These runs and the Countdown estimator comparison in Figure~\ref{fig:score-center-estimators} sample at temperature one with
top-$k$ and top-$p$ filtering disabled. Section~\ref{sec:countdown-mc} describes SC sampling.
Appendix~\ref{sec:settings-swe} specifies reward normalization and the uncapped penalty for the software engineering experiments.

\textbf{Sampling the KL penalty.} The score center and KL penalty can both be expressed as conditional expectations over behavior actions. The experiments use two sampling procedures for the penalty. \textbf{(a)} For Figures~\ref{fig:sc8b-training} through \ref{fig:agentic-training}, controlled runs estimate the KL gradient in Equation~\eqref{eq:rsc-total-gradient} with the generated action, a sample from $\mu(\cdot\mid h)$. We multiply the learner's log-probability by $\rho^{\mathrm{KL}}\log\rho^{\mathrm{KL}}$, treating this coefficient as constant during differentiation. Independent auxiliary samples estimate the SC center. \textbf{(b)} For Figure~\ref{fig:swe-preliminary}, the software engineering experiments use the same 32 auxiliary samples from $\mu$ for the SC center and the uncapped reverse KL penalty, differentiating the sample mean of $\rho\log\rho$ through the learner probabilities with $\mu$ fixed. Both implementations use behavior policy token samples and require no additional rollouts.

\textbf{Hyperparameter selection.} RSC hyperparameters vary across settings to examine different design choices motivated by our theoretical analysis. We select these hyperparameters without tuning them for higher benchmark scores. The empirical improvements reported in this paper are achieved with these untuned configurations.

\subsection{Math}\label{sec:settings-math}\label{sec:center-estimator-protocol}

\textbf{Training dense models and mixtures of experts.}
\textbf{(a)} For Figure~\ref{fig:sc8b-training}, Qwen3-8B-Base trains on DAPO-Math-17K~\citep{yu2025dapo} with $R=64$. \textbf{(b)} For the MoE panels in Figure~\ref{fig:staleness-comparison}, we use the model and dataset studied by \citet{marek2026scorecentering}: Qwen3-30B-A3B-Base and an INTELLECT-2 math subset~\citep{primeintellect2025intellect2}. We select problems whose solve rates recorded in the dataset lie in $[0,0.375]$. On this subset, we vary staleness using $R\in\{64,256,512\}$. Both models use 32 prompts per
update, responses of at most 16,384 tokens, Adam at $10^{-6}$ and gradient norm
clipping at one. Training uses BF16, and rollouts use FP8 weights and
activations with BF16 key/value caches. Verified answers receive $+1$ and other
answers receive $-1$, giving training accuracy $50(1+\overline R)\%$. For each sampler refresh interval, the MoE curves end at the last update reached by every compared run. We smooth each training run with an exponential moving average (EMA) before computing the mean and sample standard deviation across runs, using $\alpha=0.05$ for dense models and $\alpha=0.1$ for MoE models. Each curve averages the same runs throughout.

\textbf{Math evaluation.}
To evaluate the dense models used in Figure~\ref{fig:sc8b-training}, we measure accuracy on heldout math tasks in BF16, with thinking enabled and at most 16,384 new tokens per response. MATH500~\citep{lightman2024verify},
OlympiadBench~\citep{he2024olympiadbench},
Minerva Math~\citep{lewkowycz2022minerva} and Gaokao2023 Math English use
one greedy response per question. AIME24/25 and AMC23/24 use 16 responses
at temperature $0.6$ and top $p=0.95$. The task set contains 1,974
questions and 4,149 responses. Accuracy is the fraction of correct responses across the task set. We report its mean and sample standard deviation across training runs.

\textbf{Comparing estimators of the conditional mean score.}
For the Math panels in Figure~\ref{fig:score-center-estimators}, we train Qwen3-32B with thinking mode disabled on DAPO-17K with
$R=64$ and 32 prompts per update, using
the Adam settings and response limit specified in this subsection, BF16
training, and separate BF16 and FP8 rollout configurations. We compare Top-128 score estimates with Monte Carlo estimates using 32, 64 or 128 samples, with KL and entropy penalties disabled.
Evaluation uses BF16. For each benchmark, Average@32 averages correctness over 32 sampled responses per question. We give equal weight to these benchmark scores on MATH500 and AIME24/25/26. Training curves use a trailing moving average over 50 updates.

\textbf{Token-level gradient measurements.}\phantomsection\label{sec:gradient-diagnostic-settings}
Figure~\ref{fig:logit-diagnostic} compares a Qwen3-8B-Base learner at update 1,500 with a behavior checkpoint at update 1,400, giving a gap of 100 updates independently of the training refresh schedule. The learner runs in BF16, and behavior sampling uses FP8 weights and activations with BF16 key/value caches. We sample eight responses for each of 32 DAPO prompts and hold sampled actions, advantages and 128 auxiliary Monte Carlo draws fixed across methods. Response advantages follow the normalization specified in Appendix~\ref{sec:experimental-settings}. RSC uses $\beta=0.1$, with both the TIS cap and the cap on the KL ratio set to two. Token-level $\ell_2$ norms include reward and KL contributions and are measured before optimizer updates and division by the batch token count. The output logit norm differentiates with respect to the token's vocabulary logits, and the parameter norm differentiates with respect to model parameters. Both rows use 25 probability bins of equal width on $[0,1]$. Percentiles use linear interpolation and include zero norms and values beyond the displayed axes.

\subsection{Agentic Tasks}\label{sec:settings-agentic}\label{sec:webshop-protocol}\label{sec:agentic-extension-protocol}\label{sec:native-agentic-protocol}

\textbf{Training.}
The agentic experiments in Figures~\ref{fig:kl-role-agentic}, \ref{fig:sc8b-training} and~\ref{fig:agentic-training} use Qwen3-4B-Base, Adam at
$10^{-6}$, and responses to 16 tasks per update. The
sampler refresh interval is $R=64$. For these agentic comparisons, \MethodISPPOClip{} clips $\pi_\theta/\pi_{\mathrm{prox}}$ to $[0.8,1.2]$ in the PPO objective and refreshes $\pi_{\mathrm{prox}}$ from the learner every 32 successful optimizer updates. Its outer importance weight $\pi_{\mathrm{prox}}/\mu$ is uncapped and held fixed during differentiation.
Training uses BF16, and rollouts use FP8 with BF16 key/value caches. The loss
covers generated action tokens, while observations remain in context,
and each episode supplies one scalar reward. All methods on AppWorld start from a model fine-tuned for 32 supervised updates on 539 action examples from 49 official training tasks. We construct these examples from the official training solutions, pairing each action with the preceding visible interaction history. We use this shared SFT initialization to obtain reasonable task performance before RL. The other tasks start from the base model.

\begin{table}[ht]
\centering\small
\setlength{\tabcolsep}{4pt}
\renewcommand{\arraystretch}{1.08}
\caption{Agentic episode limits. Token limits for responses and contexts include environment observations.}
\label{tab:settings-agentic-limits}
\begin{tabular*}{\linewidth}{@{\extracolsep{\fill}}llrrr@{}}
\toprule
Task & Action limit & Tokens/action & Response tokens & Context tokens \\
\midrule
WebShop & 15 & 64 & 8,192 & 12,288 \\
AppWorld & 30 & 400 & 12,288 & 16,384 \\
BabyAI & Varies by task & 16 & 24,576 & 32,768 \\
TextWorld & 50 & 64 & 12,288 & 16,384 \\
Crafter & 2,000 & 16 & 24,576 & 32,768 \\
Craftax & 100,000 & 16 & 24,576 & 32,768 \\
\bottomrule
\end{tabular*}
\end{table}

\textbf{Tasks and rewards.}\phantomsection\label{par:agentic-appworld-protocol}\label{par:agentic-babyai-protocol}\label{par:agentic-textworld-protocol}
For the tasks in Figure~\ref{fig:agentic-training}, WebShop uses the full product catalog and 10,587 training goals, with
the environment's partial credit reward at the end of each episode. AppWorld uses 90 official training tasks
and binary task success.
BabyAI uses GoToSeq, OpenDoorsOrderN2, PutNextLocal and UnlockPickup,
with partial environment observations rendered as text and seven primitive actions. The terminal reward is discounted by the number of steps taken. The training set contains 16,384 combinations of levels and seeds for environment generation.
TextWorld uses 1,000 generated Cooking games with three ingredients,
three objects to collect and six rooms. Reward is the final environment score
divided by the maximum score. Crafter and the full Craftax environment use textual observations and sum the environment's rewards for achievements and changes in health over each episode.
All tasks allow 4,096 tokens in the initial prompt. Token budgets may end an
episode before its action limit. BabyAI and Crafter additionally cap
the number of generated action attempts at $2\times$ the environment's step limit.

\textbf{Evaluation and reporting.} Table~\ref{tab:agentic-final-results} evaluates the policies compared in Figure~\ref{fig:agentic-training}.
Each policy generates one evaluation episode per heldout instance, using greedy decoding in BF16.
We pool the development and test episodes for WebShop, BabyAI, Crafter and Craftax, and the validation and test episodes for TextWorld. AppWorld uses its development split. BabyAI, Crafter and Craftax use separate seeds to generate training and evaluation environments. We report the mean and sample standard deviation of the per-model scores across training runs.
For Crafter, we first combine achievement counts across evaluation episodes, then compute its geometric achievement score. Craftax final scores average the total environment reward per episode.
All episodes enter the score, including failures and
truncations. For runs stopped because of low training scores, we evaluate the last
available checkpoint.

\subsection{Countdown}\label{sec:settings-countdown}\label{sec:countdown-setup}

\textbf{Training.}
For the Countdown experiments in Figures~\ref{fig:score-center-estimators} and~\ref{fig:staleness-comparison}, we use the Countdown task studied by \citet{marek2026scorecentering} and initialize the models from Qwen3-0.6B~\citep{qwen2025qwen3}. From a shuffled
dataset of 20,000 problems, we reserve 64 prompts for evaluation. Each update
uses 64 training prompts and SGD without momentum at $0.01$, with zero
weight decay, BF16 computation and an FP32 copy of the weights for optimizer updates. Thinking mode is disabled.
The limit of 512 tokens includes the prompt. The binary reward checks whether the response reaches the target using the supplied numbers. We scale the group-centered reward advantage by two.
The comparison of SC, SC+TIS and the two RSC variants uses $R\in\{16,32,65\}$,
with automatic early stopping for low training scores disabled.

\textbf{Evaluation.}
For the Countdown evaluation in Figure~\ref{fig:score-center-estimators} and the staleness comparison discussed alongside Figure~\ref{fig:staleness-comparison}, we sample eight responses per heldout prompt every 25 updates and before
the final update. The estimator comparison
uses Top-128 with a modeled tail, MC128 and MC512 with $R=65$.

\subsection{Fully Asynchronous Software Engineering}\label{sec:settings-swe}\label{sec:swe-protocol}\label{sec:swe-additional}

For Figure~\ref{fig:swe-preliminary}, Qwen3.5-35B-A3B-Base~\citep{qwen2026qwen35} trains on 2,394 SWE-Gym tasks~\citep{pan2025swegym} with BF16 training and generation. We exclude a small number of tasks because of environment setup failures or failure of the provided gold solution to pass validation: 44 of the 2,438 SWE-Gym tasks and two of the 500 SWE-bench Verified tasks. Each batch contains 16 tasks with eight trajectories each and
receives one Adam update. Up to 1,024 trajectories may be in progress concurrently, increasing policy staleness. Table~\ref{tab:settings-swe} defines the two
configurations and their learning rates.

\begin{table}[ht]
\centering\small
\caption{Asynchronous SWE configurations. Trajectory budgets include tool observations.}
\label{tab:settings-swe}
\begin{tabular*}{\linewidth}{@{\extracolsep{\fill}}lccc@{}}
\toprule
Configuration & Trajectory budget & Learning rate & Max model calls/episode \\
\midrule
A & 128K & $10^{-6}$ & 200 \\
B & 128K & $2\times10^{-6}$ & 200 \\
\bottomrule
\end{tabular*}
\end{table}

Both configurations in Figure~\ref{fig:swe-preliminary} allow 262,144 total context tokens and 32,768 generated
tokens per model call, with up to 131,072 tokens after the initial prompt and an agent time limit of 4,800 seconds.
We compute advantages by subtracting the group mean from each reward
and dividing by the sample standard deviation plus $10^{-6}$. SWE does not apply the $G/(G-1)$ correction. We average the loss over generated action tokens within each trajectory, then average over trajectories. RSC uses uncapped reverse KL with $\beta=0.001$, while
SC+TIS and RSC (TIS) cap the reward importance ratio at two. KL regularization to a fixed reference model and entropy bonuses are disabled.

Table~\ref{tab:swe-final-results} evaluates the Figure~\ref{fig:swe-preliminary} models on 498 SWE-bench Verified tasks~\citep{jimenez2024swebench,chowdhury2024sweverified}, with eight trajectories per task at temperature one. Average@8 is the mean success rate across these trajectories and tasks. Figure~\ref{fig:swe-preliminary} smooths training rewards and log-probability differences with EMA ($\alpha=0.05$).

\FloatBarrier

\section{Additional Analysis}\label{sec:additional-experiments}
\begin{wrapfigure}{r}{2.15in}
\vspace{-10pt}
\centering
\includegraphics[width=\linewidth]{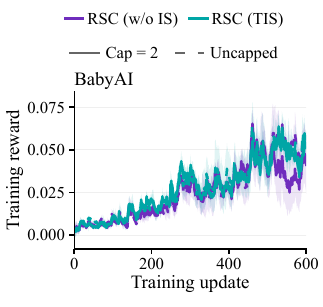}
\caption{\textbf{Clipping the KL penalty on BabyAI.} Mean $\pm$ sample standard deviation across three seeds.}
\label{fig:babyai-kl-cap}
\vspace{-6pt}
\end{wrapfigure}
The controlled experiments use a fixed cap on the KL penalty and a correction for the group reward baseline. We examine sensitivity to these choices before listing the surrogate formulas used in the gradient analysis.

\subsection{Sensitivity to Clipping the KL Penalty}\label{sec:agentic-additional}

\textbf{Clipping the KL penalty has a modest effect.}\label{sec:babyai-kl-cap}
Figure~\ref{fig:babyai-kl-cap} compares capped and uncapped KL on BabyAI. Both RSC variants learn with either choice, with similar training curves and small differences in mean reward. The cap used in the main experiments was fixed before this ablation and was not tuned against the uncapped results.

\WFclear
\needspace{3\baselineskip}
\subsection{Testing Sensitivity to Group-Size Correction}\label{sec:group-size-ablation}\label{sec:agentic-no-group}

\textbf{Group-size correction has no consistent advantage across tasks and variants.} Figure~\ref{fig:agentic-no-group} compares RSC with and without the correction defined in Appendix~\ref{sec:experimental-settings}, keeping the group size and KL coefficient fixed. Results do not support a uniform conclusion across tasks: the effect of group-size correction differs between AppWorld and BabyAI.

\begin{figure}[ht!]
\centering
\includegraphics[width=\linewidth]{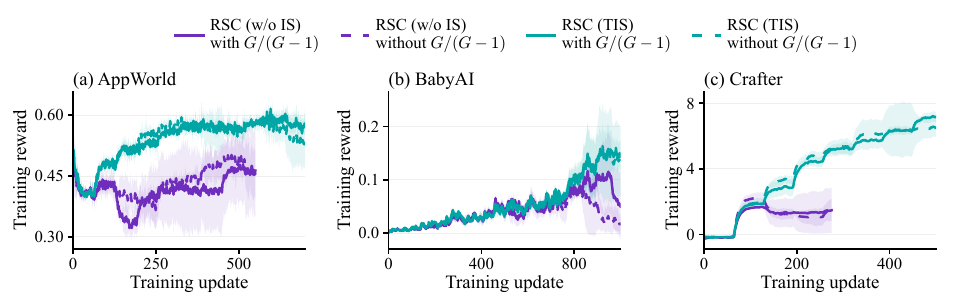}
\caption{\textbf{Sensitivity to group-size correction.} Solid and dashed curves compare each RSC variant with and without the group-size correction at fixed group size. Curves and bands show mean training reward $\pm$ sample standard deviation across three seeds.}
\label{fig:agentic-no-group}
\end{figure}

\FloatBarrier

\subsection{Surrogate Formulas}\label{sec:surrogate-formulas}
Table~\ref{tab:surrogate-methods} expands the surrogate integrands and logit derivatives in Figure~\ref{fig:regularized-potentials}.
\begin{table}[!ht]
\centering
\begingroup
\normalsize
\setlength{\tabcolsep}{3pt}
\setlength{\arrayrulewidth}{0.4pt}
\setlength{\extrarowheight}{2pt}
\renewcommand{\arraystretch}{1.35}
\definecolor{surrogateIS}{HTML}{009E73}
\definecolor{surrogateSC}{HTML}{111111}
\definecolor{surrogateTIS}{HTML}{0072B2}
\definecolor{surrogateRSC}{HTML}{6F2DBD}
\definecolor{surrogateRSCTIS}{HTML}{00A6A6}
\definecolor{surrogateKL}{HTML}{008000}
\begin{tabular}{|l|c|c|c|c|}
\hline
\rule{0pt}{22pt}\textbf{Method} & $\boldsymbol{a(\rho)}$ & $\boldsymbol{F_a(\rho)}$ & $\boldsymbol{w}$ & $\boldsymbol{\dfrac{\partial w/\partial z_a}{1-\pi_\theta(a\mid h)}}$ \\[2pt]
\hline
{\bfseries\color{surrogateIS}IS}
& ${\color{red}\rho}$
& $\rho-1$
& $A(\rho-1)$
& $A{\color{red}\boldsymbol{\rho}}$ \\[2pt]
\hline
{\bfseries\color{surrogateTIS}SC+TIS}
& $\min(\rho,c)$
& $\begin{gathered}
\rho-1\quad{\scriptstyle(\rho\le c)}\\
c-1+c\log\frac{\rho}{c}\quad{\scriptstyle(\rho>c)}
\end{gathered}$
& $A\begin{cases}
\rho-1,&\rho\le c,\\
c-1+c\log\frac{\rho}{c},&\rho>c
\end{cases}$
& $A\min(\rho,c)$ \\[2pt]
\hline
{\bfseries\color{surrogateSC}\renewcommand{\arraystretch}{1}\setlength{\extrarowheight}{0pt}\begin{tabular}{@{}l@{}}SC\\(w/o IS)\end{tabular}}
& $1$
& $\log\rho$
& $A\log\rho$
& $A$ \\[2pt]
\hline
{\bfseries\color{surrogateRSCTIS}\renewcommand{\arraystretch}{1}\setlength{\extrarowheight}{0pt}\begin{tabular}{@{}l@{}}RSC\\(TIS)\end{tabular}}
& $\min(\rho,c)$
& $\begin{gathered}
\rho-1\quad{\scriptstyle(\rho\le c)}\\
c-1+c\log\frac{\rho}{c}\quad{\scriptstyle(\rho>c)}
\end{gathered}$
& $\begin{gathered}
A\begin{cases}
\rho-1,&\rho\le c,\\
c-1+c\log\frac{\rho}{c},&\rho>c
\end{cases}\\[2pt]
{\color{surrogateKL}-\beta(\rho\log\rho-\rho+1)}
\end{gathered}$
& $A\min(\rho,c){\color{surrogateKL}-\beta\rho\log\rho}$ \\[2pt]
\hline
{\bfseries\color{surrogateRSC}\renewcommand{\arraystretch}{1}\setlength{\extrarowheight}{0pt}\begin{tabular}{@{}l@{}}RSC\\(w/o IS)\end{tabular}}
& $1$
& $\log\rho$
& $\begin{gathered}
A\log\rho\\[2pt]
{\color{surrogateKL}-\beta(\rho\log\rho-\rho+1)}
\end{gathered}$
& $A{\color{surrogateKL}-\beta\rho\log\rho}$ \\[2pt]
\hline
\end{tabular}
\endgroup
\caption{\textbf{Surrogate integrands and logit derivatives.}
Here $\rho=\pi_\theta(a\mid h)/\mu(a\mid h)$, $A=A^{\mu}(h,a)$ and $c\ge1$.
The last column factors out $1-\pi_\theta(a\mid h)$ from the logit derivative, with $\mu$, $A$ and the other logits fixed.
RSC uses $\beta>0$ and standard, uncapped reverse KL.}
\label{tab:surrogate-methods}
\end{table}

\FloatBarrier

\section{Convergence with and without KL regularization}\label{sec:appendix-theory}
We analyze policy updates with no regularization ($\beta=0$) and with
KL regularization ($\beta>0$), for unit or token importance sampling
(TIS) reward weights. Theorems~\ref{thm:sc-kl-progress} and~\ref{thm:chenlu-sc-kl-convergence}
state the main regularized convergence results after the setup below.
The analysis has three parts.
Section~\ref{sec:theory-setup} defines the objectives and policy updates
and explains the benefit of KL regularization.
Section~\ref{sec:theory-outer} proves how population fitting accuracy
implies reward progress.
Section~\ref{sec:theory-inner} bounds the population error of a policy
fitted to a fixed dataset and concludes with the combined reward and
sample guarantees.

\subsection{Setup and main convergence results}\label{sec:theory-setup}\label{sec:offpolicy-setting}
A prompt $x\sim\mathcal D$ generates a response $Y=(a_1,\ldots,a_H)$
on a finite deterministic token tree, with binary terminal reward
$R(x,Y)\in\{0,1\}$. Here $H$ is the horizon and $h_t=(x,a_{<t})$
is the history at token position $t$. Policies have common full support;
terminal histories are padded with deterministic actions.
Start from a full-support policy $\pi_0$. At outer round
$k=0,\ldots,K-1$, the behavior policy is fixed at $\mu=\pi_k$, and
$\pi$ denotes a candidate next policy. Here $K$ is the total number of
outer rounds. Define its token ratio and the
behavior continuation values by
\begin{align}
 \rho_t(h,a)&=\frac{\pi(a\mid h)}{\mu(a\mid h)},&
 Q_t^\mu(h,a)&=\E_\mu[R\mid h_t=h,a_t=a],\nonumber\\
 V_t^\mu(h)&=\E_{a\sim \mu(\cdot\mid h)}Q_t^\mu(h,a),&
 A_t^\mu(h,a)&=Q_t^\mu(h,a)-V_t^\mu(h).
 \label{eq:theory-basic-notation}
\end{align}
Write $d_t^\mu$ for the distribution of histories under $\mu$, including
the prompt distribution, and $\E_\mu$ for expectation over prompts and
responses sampled from $\mu$. The reward of a policy is
$J(\pi)=\E_{\pi}R$, with $J_x(\pi)=\E_\pi[R\mid x]$ for a fixed prompt. Write
$s_k(x)=J_x(\pi_k)$ and $\mu_t=\mu(\cdot\mid h_t)$. For any other policy
$\pi$, define $Q_t^\pi$, $V_t^\pi$, and $A_t^\pi$ by replacing $\mu$
in Equation~\eqref{eq:theory-basic-notation}; at a terminal history,
$V_{H+1}^\pi=R$. When $\pi=\pi_\theta$, its score is
$\psi_\theta(h,u)=\nabla_\theta\log\pi_\theta(u\mid h)$, as in
Equation~\eqref{eq:method-score-ratio}.

\paragraph{Population objective.}
Use the effective importance ratio of Section~\ref{sec:sc-background}:
$a(\rho)=\rho^{\mathrm{TIS}}$. For the one-sided TIS rule analyzed here,
$a(\rho)=\min(\rho,c)$ with cap $c\ge1$; unit-weight SC sets $a(\rho)=1$.
Integrating this weight gives a potential $F_a$ and the centered
reward objective:
\begin{equation}\label{eq:generalized-sc-potential}
 F_a(\rho)=\int_1^\rho\frac{a(u)}u\,du,\qquad \rho F_a'(\rho)=a(\rho).
\end{equation}\begin{equation}\label{eq:generalized-sc-objective}
 \mathcal C_a(\pi)=\E_{\mu}\sum_{t=1}^H
 A_t^{\mu}(h_t,a_t)F_a\!\left(\frac{\pi(a_t\mid h_t)}{{\mu}(a_t\mid h_t)}\right).
\end{equation}
The behavior policy $\mu$, its history distribution, and its advantages
stay fixed during fitting.
In particular, $F_a(\rho)=\log\rho$ for SC and
$F_a(\rho)=\rho-1$ for TIS whenever the cap is inactive.

Use the independent penalty ratio from Section~\ref{sec:rsc-method},
$b(\rho)=\rho^{\mathrm{KL}}=\min(\rho,c_b)$ with $c_b\in[1,\infty]$.
An infinite cap gives standard reverse KL. Section~\ref{sec:surrogate-details}
gives the explicit integrands and derivatives used in the main text.
The penalty potential is
\begin{equation}\label{eq:sc-penalty-potential}
 K_b(\rho)=\int_1^\rho\frac{b(v)\log b(v)}v\,dv.
\end{equation}
At round $k$, write $\mathcal C_k$ for the centered reward objective
with $\mu=\pi_k$. The weights $a$, $b$, and coefficient $\beta$ are fixed
within each argument, so we omit them from the subscripts of
$\mathcal C_k$ and the regularized objective $\mathcal W_k$. For a token log-ratio radius $r>0$, the full
feasible family is
\begin{equation}\label{eq:theory-neighborhood}
 \Gamma_k(r)=\left\{\pi:\left|\log\frac{\pi(u\mid h)}{\pi_k(u\mid h)}\right|
 \le r\ \text{for every reachable history $h$ and available token $u$}\right\}.
\end{equation}
Each member is a normalized token policy. Write $\Gamma_k=\Gamma_k(r)$
when the radius is fixed. With regularization coefficient $\beta\ge0$,
the population fitting problem is
\begin{equation}\label{eq:sc-kl-objective}
 \sup_{\pi\in\Gamma_k}\mathcal W_k(\pi),\qquad
 \mathcal W_k(\pi)=\mathcal C_k(\pi)
                  -\beta\E_{\mu}\sum_t K_b(\rho_t).
\end{equation}
Here $\rho_t=\pi(a_t\mid h_t)/\mu(a_t\mid h_t)$ in the expectation.
For $\beta=0$, this is just $\mathcal C_k$. When the penalty cap
is inactive, $K_b(\rho)=\rho\log\rho-\rho+1$, so the same objective is
\begin{equation}\label{eq:theory-uncapped-objective}
 \mathcal W_k(\pi)=\mathcal C_k(\pi)
 -\beta\sum_{t=1}^H\E_{h_t\sim d_t^\mu}
       \KL\bigl(\pi(\cdot\mid h_t)\|\mu(\cdot\mid h_t)\bigr).
\end{equation}
Equation~\eqref{eq:theory-uncapped-objective} is the main-text surrogate
in Equation~\eqref{eq:method-regularized-target}, with the round index
made explicit. Equation~\eqref{eq:sc-kl-objective} also covers an active
penalty cap through $K_b$. Thus the penalty compares current-token
distributions on behavior histories. The objective is defined for every
$\beta\ge0$. The unregularized analysis uses $r=1/(8H)$, while the
regularized theorems below specify a sufficient coefficient and use
$r=1/2$. At this constant radius, reward and penalty caps of two are
inactive throughout the feasible family.

\paragraph{Empirical objective and the update loop.}
The inner fit estimates this objective using response rewards and token
probabilities. For a behavior response, define
\begin{align}
 Z_\pi(Y)&=\sum_{t=1}^H\left[F_a(\rho_t(h_t,a_t))
 -\E_{u\sim \mu(\cdot\mid h_t)}F_a(\rho_t(h_t,u))\right],\nonumber\\
 B_\pi(Y)&=\sum_{t=1}^H K_b(\rho_t(h_t,a_t)).
 \label{eq:theory-sample-terms}
\end{align}
At round $k$, draw $m$ independent prompts and $G\ge2$ independent
responses $Y_{jg}$ from $\mu$ for each prompt $x_j$. Let
$R_{jg}=R(x_j,Y_{jg})$ and $\overline R_j=G^{-1}\sum_gR_{jg}$. Use
\begin{equation}\label{eq:theory-empirical-objective}
 \widehat{\mathcal W}_k(\pi)=\frac1m\sum_{j=1}^m
 \left[\frac1{G-1}\sum_{g=1}^G(R_{jg}-\overline R_j)Z_\pi(Y_{jg})
 -\frac\beta G\sum_{g=1}^G B_\pi(Y_{jg})\right].
\end{equation}
The reward term is the unbiased sample covariance of reward and centered
potential; the penalty is a response average. Hence its expectation is
$\mathcal W_k(\pi)$ for each fixed candidate. When $\beta=0$, denote
this empirical objective by $\widehat{\mathcal C}_k$.

Differentiating Equation~\eqref{eq:theory-empirical-objective} gives
the grouped gradient in Equation~\eqref{eq:sc-kl-group-update}.
It is the sum of the main-text contributions in
Equation~\eqref{eq:rsc-total-gradient}, with the corrected group reward
$A_i=G(R_i-\overline R)/(G-1)$, exact conditional centers, and the
generated action used to sample the penalty.
Proposition~\ref{prop:sc-kl-sgd} establishes both the exact empirical
gradient identity and its population expectation. Independent Monte
Carlo centers give the corresponding sampled-potential objective;
Section~\ref{sec:experimental-settings} describes the implementation
choices for sampling and reducing token contributions.

The inner solver holds $\mu$ and this dataset fixed and returns
$\pi_{k+1}\in\Gamma_k$ satisfying
\begin{equation}\label{eq:theory-fitting-condition}
 \widehat{\mathcal W}_k(\pi_{k+1})
 \ge\sup_{\pi\in\Gamma_k}\widehat{\mathcal W}_k(\pi)-e_k.
\end{equation}
The tolerance $e_k$ includes optimization error and representation error
relative to the full family. The outer loop then sets $\mu=\pi_{k+1}$ and
collects a fresh dataset. After $K$ rounds, the budget is $Km$ prompt
draws and $KmG$ responses.

\paragraph{Main convergence results.}
We compare the reward of one response from the fitted policy with
best-of-$N$ sampling, where $N$ is a positive integer counting
independent responses from the initial policy. The benchmark reward is
\begin{equation}\label{eq:theory-best-of-n}
 V_N(\pi_0)=\E_x\left[1-(1-J_x(\pi_0))^N\right].
\end{equation}
To state the sample bounds, measure the distance between policies by
\begin{equation}\label{eq:sc-cover-metric}
 d_\infty(\pi,\pi')=\sup_{t,h,u}|\log \pi(u\mid h)-\log \pi'(u\mid h)|.
\end{equation}
A radius-$v$ cover approximates every feasible policy within distance
$v$, with cover centers in the feasible family. The covering number
is the minimum number of such centers. Let $\mathcal N(v)$ be
a common finite nonincreasing upper bound on these covering numbers
across rounds. This bound holds conditional on the history before each
round; the cover may depend on that history but not on the fresh data.

The first result concerns SC+TIS+KL with exact conditional centers.
The second concerns unit-weight SC+KL and also allows Monte Carlo
centers. For a positive integer $M$, the number of auxiliary tokens
per visited history, replace the conditional expectation in
$Z_\pi$ from Equation~\eqref{eq:theory-sample-terms} by
\begin{equation}\label{eq:theory-mc-center}
 \widehat{\E}_{\mu}[F_a(\rho_t(h_t,u))\mid h_t]
 =\frac1M\sum_{i=1}^M F_a(\rho_t(h_t,U_{t,i})),\qquad
 U_{t,i}\sim \mu(\cdot\mid h_t).
\end{equation}
Conditional on each visited history, these auxiliary draws are
independent of the response's current and future actions. Use independent
randomness at different positions and for different responses, and hold
all responses and auxiliary draws fixed during fitting. Write $M=\infty$
for exact conditional expectations, with $1/M=0$.

Both theorems use the finite deterministic token tree, binary reward,
and common full support specified at the start of this subsection.
The target reward error is $\epsilon$, and $\delta$ is the allowed
failure probability over the sampled prompts, responses, and auxiliary
tokens when present. The fitted policy is evaluated by a single fresh
response, through $J(\pi_K)$.

\begin{theorem}[Horizon-independent outer rounds with regularization]
\label{thm:sc-kl-progress}
Use $a(\rho)=b(\rho)=\min(\rho,2)$ and $\beta=2$.
At each round, use the radius-$1/2$ family $\Gamma_k$ in
Equation~\eqref{eq:theory-neighborhood}.
For $N\ge1$, $0<\epsilon<1$, and $0<\delta<1$, choose
\[
 K=\left\lceil4\log\frac{6N}{\epsilon}\right\rceil,
 \qquad e_k\le\frac{\epsilon}{160\pi^2NK^2}.
\]
At every round, draw $m$ fresh independent prompts from $\mathcal D$
and $G\ge2$ independent
responses per prompt from $\mu=\pi_k$. Use exact conditional centers in
Equation~\eqref{eq:theory-empirical-objective} and suppose the returned
policy satisfies Equation~\eqref{eq:theory-fitting-condition}.
If
\begin{equation}\label{eq:sc-kl-samples}
 m\ge\left\lceil
 \frac{32K+1760\pi^2NK^2H/\lfloor G/2\rfloor}{\epsilon}
 \log\!\left[\frac{4K}{\delta}\,
 \mathcal N\!\left(\frac{\epsilon}{2160\pi^2NK^2H}\right)
 \right]\right\rceil,
\end{equation}
then $V_N(\pi_0)-J(\pi_K)\le\epsilon$ with probability at least
$1-\delta$.
\end{theorem}

\begin{theorem}[\MethodCurrentKL{} with finite samples and Monte Carlo centers]
\label{thm:chenlu-sc-kl-convergence}
Let $N\ge1$, $0<\epsilon<1$, $0<\delta<1$, $G\ge2$, and
$M\in\mathbb N\cup\{\infty\}$. Use $a(\rho)=1$, $b(\rho)=\min(\rho,2)$, $\beta=2$, and the
radius-$1/2$ family in Equation~\eqref{eq:theory-neighborhood}.
Use Equation~\eqref{eq:theory-empirical-objective} with the conditional
centers in Equation~\eqref{eq:theory-mc-center}. Set
\begin{align}
 K&=\left\lceil\frac{\log(6N/\epsilon)}{\log(67/60)}\right\rceil,
 &\kappa&=4\pi^2NK,\nonumber\\
 B_M&=24+8e(1+1/M),
 &\nu&=\frac{\epsilon}{288\kappa HK}.
 \label{eq:chenlu-sc-kl-constants}
\end{align}
At each round, draw $m$ fresh independent prompts from $\mathcal D$
and $G$ independent responses per prompt from $\mu=\pi_k$, with auxiliary
tokens sampled as specified above. Suppose
the fitting condition in Equation~\eqref{eq:theory-fitting-condition} holds with
$e_k\le\epsilon/(32\kappa K)$. If the number of prompt draws per
round satisfies
\begin{equation}\label{eq:chenlu-sc-kl-samples}
 m\ge\left\lceil\frac{\log[4K\mathcal N(\nu)/\delta]}{\epsilon}
       \left(32K+\frac{16\kappa B_MHK}{\lfloor G/2\rfloor}\right)\right\rceil,
\end{equation}
then, with probability at least $1-\delta$,
\begin{equation}\label{eq:chenlu-sc-kl-bon}
 V_N(\pi_0)-J(\pi_K)\le\epsilon,
 \qquad V_N(\pi_0)=\E_x\!\left[1-(1-J_x(\pi_0))^N\right].
\end{equation}
\end{theorem}
Theorems~\ref{thm:sc-kl-progress} and~\ref{thm:chenlu-sc-kl-convergence}
give sufficient outer-round counts independent of the horizon, under
their stated fitting conditions. Sections~\ref{sec:proof-tis-kl}
and~\ref{sec:proof-unit-kl} combine the outer and inner lemmas to prove
these results. The unregularized comparison appears in
Theorem~\ref{thm:sc-sample-complexity}.

The guarantees concern the accuracy achieved by fitting a fixed dataset;
they do not specify how many optimizer iterations are needed. Each outer
round uses fresh data from the current policy. Horizon-independent round
counts therefore still have horizon-dependent sampling costs.

\paragraph{Relation to the experimental coefficient.}
The coefficient fixed in these theorems is a sufficient analytical
choice. The controlled training experiments use $\beta=0.01$, and
asynchronous SWE uses $\beta=0.001$
(Section~\ref{sec:experimental-settings}). The objectives and gradient
identities apply at these smaller coefficients, but the stated
convergence and sample bounds use the theorem assumptions. Extending
those bounds to smaller positive coefficients requires tracking their
effect on curvature, target feasibility, and reward transfer.

\paragraph{Benefit of KL regularization.}
Without KL regularization, we control the change in visited histories
using the token log-ratio bound. A response ratio is a product of $H$
token ratios and can be as large as $e^{Hr}$. This bound is sharp, so
to keep sequence ratios uniformly bounded using the token constraint
alone, we must choose $r=O(1/H)$
(Lemma~\ref{lem:token-radius-sequence-ratio}). We use $r=1/(8H)$,
which also controls the local objective-to-reward approximation error.
The resulting sufficient round count is $O(H\log(N/\epsilon))$ for
reward accuracy $\epsilon$ against the best-of-$N$ benchmark in
Equation~\eqref{eq:theory-best-of-n}.
For $\beta>0$, KL adds a penalty for policy movement. The following
two mechanisms yield the improved bounds under the assumptions of
Theorems~\ref{thm:sc-kl-progress} and~\ref{thm:chenlu-sc-kl-convergence}.
\begin{enumerate}
\item \textbf{Inner fitting: objective accuracy controls distance to the target.}
For TIS with inactive caps, the population maximizer has token
probabilities proportional to $\mu(u\mid h)\exp(Q_t^\mu(h,u)/\beta)$.
The objective gap equals $\beta$ times the conditional KL from the
candidate to this target, averaged over behavior
histories (Lemma~\ref{lem:rsc-target-gap}). For unit-weight SC, curvature
also makes the gap control squared token-ratio error
(Lemma~\ref{lem:chenlu-sc-kl-curvature}). In both cases, the second moment
of the sampled objective difference is bounded by a constant times
$H$ times the population gap
(Lemmas~\ref{lem:rsc-response-moments} and~\ref{lem:chenlu-sc-kl-moments}).
Thus the noise in comparing a candidate with the target decreases
as the fit improves, which reduces the samples needed to certify an
accurate fit.

\item \textbf{Outer transfer: fitting accuracy controls reward at a constant radius.}
For the regularized TIS target, continuation-value differences telescope
along each response. Consequently, its target-to-behavior ratio is at
most $e^{1/\beta}$ for every prefix and full response at token
radius $1/\beta$ when the caps are inactive and the target is feasible
(Lemma~\ref{lem:rsc-prefix-coverage}). For fixed $\beta$, this bound is
independent of $H$, whereas the token constraint alone permits
$e^{H/\beta}$. This coverage bound converts the
objective gap into response-level reward control
(Lemma~\ref{lem:rsc-gap-transfer}). The ratio bound applies to the target;
an approximate fit is controlled through its objective gap.
Unit-weight SC uses a different transfer argument: under the stated
coefficient and radius conditions, its regularized objective directly
lower-bounds $J(\pi)-J(\mu)$
(Lemma~\ref{lem:chenlu-sc-kl-lower-bound}), so it does not require a
sequence-ratio bound for its optimizer.

For the settings in the main theorems, the TIS target increases success
odds by a constant factor, and a feasible unit-weight SC perturbation
gains a constant
fraction of reward variance
(Lemmas~\ref{lem:rsc-target-progress} and~\ref{lem:chenlu-sc-kl-progress}).
\end{enumerate}

Together, the inner fitting certificates and outer progress bounds give
$O(\log(N/\epsilon))$ sufficient outer rounds, compared with
$O(H\log(N/\epsilon))$ for the unregularized analysis. Up to logarithmic
and covering-number factors, the sufficient total prompt bounds change
from $(H^2+NH^3/G)/\epsilon$ to $(1+NH/G)/\epsilon$, where $G$ is the
number of responses per prompt; the latter statement uses a fixed
number of auxiliary center samples for unit-weight SC.
The precise budgets are in Equations~\eqref{eq:sc-total-samples},
\eqref{eq:sc-kl-total-prompts}, and~\eqref{eq:chenlu-sc-kl-total-samples}.

\subsection{Outer-loop reward improvement}\label{sec:theory-outer}
We first state the common lemma that accumulates one-round reward errors.
We then prove the progress bounds for unregularized SC and TIS,
unit-weight KL regularization, and TIS with KL regularization.
In each case, the fitting error is measured by the corresponding
population objective defined in Section~\ref{sec:theory-setup}.
\subsubsection{Angular transfer for reward improvement}\label{app:sc-angular-transfer}
This lemma accumulates one-round errors relative to a fixed increase
in success odds. Its two deficit bounds allow the later proofs to use
either a KL error or a direct reward shortfall.
\begin{lemma}[Relative angular transfer, adapted from \cite{bai2026proximal}]\label{lem:bernoulli-angular-transfer}
Let $s_k(x)=J_x(\pi_k)$ be success probabilities and fix a log-odds increment $\lambda>0$. Write
$\alpha_k=\arcsin\sqrt{s_k}$,
$T(\alpha)=\arctan(e^{\lambda/2}\tan\alpha)$, so that $\sin^2T(\alpha_k)$
has log-odds $\operatorname{logit}s_k+\lambda$, and define the one-sided
angular deficit
\[
 \xi_k(x)=[T(\alpha_k(x))-\alpha_{k+1}(x)]_+.
\]
Put $s(x)=1-(1-s_0(x))^N$. For $N\ge1$, $0<\epsilon<1$, and
$K\ge\lceil\lambda^{-1}\log(3N/\epsilon)\rceil$, at every prompt,
\begin{equation}\label{eq:pointwise-angular-transfer}
 s(x)-s_K(x)\le\epsilon+\min\left\{1,
 16N\left(\sum_{k<K}\xi_k(x)\right)^2\right\}
 \le\epsilon+\min\left\{1,16NK\sum_{k<K}\xi_k(x)^2\right\}.
\end{equation}
Each deficit admits a bound in terms of Bernoulli KL or probability shortfall:
\begin{enumerate}
\item For target success probabilities $s_k^*(x)$, if $\operatorname{logit}s_k^*(x)\ge\operatorname{logit}s_k(x)+\lambda$, then
 $\xi_k^2\le(\pi^2/8)D_{\mathrm{KL}}(\operatorname{Bern}(s_{k+1})
 \Vert\operatorname{Bern}(s_k^*))$, and the same holds with the
 Bernoulli arguments reversed.
\item $\xi_k^2\le(\pi^2/4)[\sin^2T(\alpha_k)-s_{k+1}]_+$, that is,
 $(\pi^2/4)\times$ the shortfall from the ideal step in probability.
\end{enumerate}
In particular, with
$E_k=\E_xD_{\mathrm{KL}}(\operatorname{Bern}(s_{k+1}(x))
\Vert\operatorname{Bern}(s_k^*(x)))$ and targets as in the first case,
$\E_x[s(x)-s_K(x)]\le\epsilon+2\pi^2NK\sum_{k<K}E_k$.
\end{lemma}
\begin{proof}
\textbf{Step 1: compare the actual and ideal sequences.}
The relative angular-transfer lemma of \cite{bai2026proximal}
uses comparator coverage $s_\star(x)\le C_\star s_0(x)$.
Here $s_\star(x)=1-(1-s_0(x))^N\le Ns_0(x)$, so $C_\star=N$.
We give the specialization explicitly. Fix a prompt and put $s=1-(1-s_0)^N$.
Extend $T$ continuously to the endpoints. This map is increasing and concave on $[0,\pi/2]$,
with $T(0)=0$ and $T(\alpha)\ge\alpha$. Indeed,
\[
 T'(\alpha)=\frac{e^{\lambda/2}}{1+(e^\lambda-1)\sin^2\alpha}
\]
is positive and nonincreasing, establishing monotonicity and concavity.
Define its ideal sequence $\alpha_k^*=T^k(\alpha_0)$.
If $s_0=0$, the comparator tail is zero and contributes no deficit.
Otherwise concavity implies, with
$a_k=[1-\alpha_k/\alpha_k^*]_+$,
\[
 a_{k+1}\le a_k+\frac{\xi_k}{\alpha_{k+1}^*},\qquad
 a_K\le\frac{\sum_{k<K}\xi_k}{\alpha_0}.
\]
The ideal success mass is
$P_K=\sin^2\alpha_K^*=e^{K\lambda}s_0/(1-s_0+e^{K\lambda}s_0)$.
Since $s\le Ns_0$ and $e^{K\lambda}\ge3N/\epsilon$,
$P_K\ge3s/(\epsilon+3s)$.
If $s<\epsilon$, the reward deficit is already at most $\epsilon$.
For $s\ge\epsilon$, $P_K\ge3/4$, so $\alpha_K^*\ge\pi/3$.
If $s\le1/2$ and $s_K<s$, then $\alpha_K<\pi/4$, hence
$a_K>1/4$ and $s-s_K\le16s a_K^2$.
If $s>1/2$ and $s_K\le P_K$, the Lipschitz bound for cosine gives
\[
 1-s_K\le2(1-P_K)+2(\alpha_K^*-\alpha_K)^2.
\]
The convex function $s-1+2\epsilon/(\epsilon+3s)$ is at most
$\epsilon$ on $s\in[\epsilon,1]$, by checking its endpoints.
Also $2(\alpha_K^*a_K)^2\le16s a_K^2$ when $s>1/2$.
If $s_K>P_K$, the ideal bound gives $s-s_K\le s-P_K\le s-1+\epsilon/(\epsilon+3s)\le\epsilon$, using the same endpoint bound.
Thus in every case
\[
 s-s_K\le\epsilon+16s a_K^2
 \le\epsilon+16N\left(\sum_{k<K}\xi_k\right)^2,
\]
using $s\le Ns_0\le N\alpha_0^2$.
Since also $s-s_K\le s\le1$, and
$\min\{1,\epsilon+y\}\le\epsilon+\min\{1,y\}$ for $y\ge0$, this
proves the first inequality in \eqref{eq:pointwise-angular-transfer}.
Cauchy-Schwarz proves the second.

\textbf{Step 2: bound each angular deficit.}
For the first deficit bound, the target angle
$\alpha_k^\dagger=\arcsin\sqrt{s_k^*}$ is at least $T(\alpha_k)$,
because $T$ adds exactly $\lambda$ to the log-odds. Thus
$\xi_k\le|\alpha_{k+1}-\alpha_k^\dagger|$.
Use the unnormalized squared Hellinger distance $\sum_u(\sqrt{P(u)}-\sqrt{Q(u)})^2$. For Bernoulli probabilities with angular difference $d\le\pi/2$, this distance is $4\sin^2(d/2)\ge8d^2/\pi^2$ and is
at most their KL divergence in either direction. This gives the
stated bound in both directions.
For the second, let $0\le\phi\le\theta\le\pi/2$. Then
$\theta+\phi\in[\theta-\phi,\pi-(\theta-\phi)]$, so
\[
 \sin^2\theta-\sin^2\phi=\sin(\theta+\phi)\sin(\theta-\phi)
 \ge\sin^2(\theta-\phi)\ge\frac4{\pi^2}(\theta-\phi)^2.
\]
Apply this with $\theta=T(\alpha_k)$ and $\phi=\alpha_{k+1}$ when
$\alpha_{k+1}<T(\alpha_k)$. Otherwise $\xi_k=0$.
The averaged statement follows from the first case and
\eqref{eq:pointwise-angular-transfer} after averaging prompts, since
$16NK\cdot\pi^2/8=2\pi^2NK$.
The argument applies to every realized sequence and hence also in
expectation for random fitted policies.
\end{proof}
\subsubsection{Unregularized reward progress}\label{app:generalized-sc}
\begin{assumption}[Optimization oracle accuracy]
\label{ass:generalized-sc-oracle}
At round $k$, fix behavior $\mu=\pi_k$ and use the radius-$r$ family
$\Gamma_k(r)$ from Equation~\eqref{eq:theory-neighborhood}.
The fitted policy is parameterized by the nonlinear model and must
belong to this neighborhood. Let $\mathcal C_{k,x}$ be
\eqref{eq:generalized-sc-objective} conditional on prompt $x$, and write
$\mathcal C_k=\E_x\mathcal C_{k,x}$. Given $\zeta_k\ge0$, the
optimization routine returns one policy $\pi_{k+1}\in\Gamma_k$ with
\begin{equation}\label{eq:generalized-sc-oracle}
 \mathcal C_k(\pi_{k+1})
 \ge\sup_{\pi\in\Gamma_k}\mathcal C_k(\pi)-\zeta_k.
\end{equation}
\end{assumption}

The next lemma isolates the history-distribution change caused by token
updates. It explains why a token-radius argument alone requires the radius
to decrease with the horizon.
\begin{lemma}[Token-radius control of prefix and response ratios]
\label{lem:token-radius-sequence-ratio}
Let $\pi,\mu$ have common full support and satisfy
$|\log(\pi(u\mid h)/\mu(u\mid h))|\le r$ at every history and token.
For every prompt $x$ and prefix $h_t=(x,a_{<t})$,
\begin{equation}\label{eq:token-radius-prefix-ratio}
 e^{-(t-1)r}\le\frac{d_t^\pi(h_t\mid x)}{d_t^\mu(h_t\mid x)}
 \le e^{(t-1)r},\qquad 1\le t\le H+1.
\end{equation}
In particular, $e^{-Hr}\le \pi(Y\mid x)/\mu(Y\mid x)\le e^{Hr}$.
The response bounds are attained by full-support binary token policies.
Thus a horizon-independent uniform ratio bound based only on the token
constraint requires $Hr=O(1)$. The choice $r=1/(8H)$ bounds all these
ratios between $e^{-1/8}$ and $e^{1/8}$.
\end{lemma}
\begin{proof}
At a fixed prompt, the prefix ratio is the product of the first $t-1$
conditional token ratios. Taking logarithms and summing their bounds
proves Equation~\eqref{eq:token-radius-prefix-ratio}, including the
full response when $t=H+1$.
For sharpness, let tokens be independent and binary at every position,
with $\mu(1)=1/(1+e^r)$ and $\pi(1)=e^r/(1+e^r)$.
The ratios for tokens $1$ and $0$ are $e^r$ and $e^{-r}$, respectively.
The all-one and all-zero responses therefore attain $e^{Hr}$ and
$e^{-Hr}$. Keeping a uniform upper bound independent of $H$ thus
requires $Hr$ to remain bounded.
\end{proof}

\begin{lemma}[Feasible token policy with positive objective gain]
\label{lem:common-progress-witness}
For $r>0$ and behavior $\pi_k$, define the token policy
\begin{equation}\label{eq:common-progress-witness}
 \widetilde\pi_k(b\mid h_t)=\pi_k(b\mid h_t)
 \frac{1+(e^r-1)Q_t^{\pi_k}(h_t,b)}
      {1+(e^r-1)V_t^{\pi_k}(h_t)}.
\end{equation}
It is normalized, has token log-ratios in $[-r,r]$, and, for SC or
SC+TIS with $c\ge e^r$, satisfies
\[
 \mathcal C_{k,x}(\widetilde\pi_k)
 \ge(1-e^{-r})s_k(x)(1-s_k(x)).
\]
For the symmetric neighborhood $\Gamma_k$ in
\eqref{eq:theory-neighborhood}, this policy belongs to $\Gamma_k$,
so the same lower bound holds for its conditional optimum.
\end{lemma}
\begin{proof}
Normalization follows from $\E_{\pi_k(\cdot\mid h_t)}Q_t^{\pi_k}=V_t^{\pi_k}$.
Both numerator and denominator of the ratio lie in $[1,e^r]$.
Put $u=e^r-1$. For SC, the conditional gain is
$\operatorname{Cov}(Q,\log(1+uQ))$. The derivative of
$z\mapsto\log(1+uz)$ on $[0,1]$ is at least $u/(1+u)$.
Using an independent copy of $Q$ in the covariance identity therefore
bounds this gain below by $(1-e^{-r})\operatorname{Var}(Q)$.
For inactive TIS, the gain is $u\operatorname{Var}(Q)/(1+uV)$,
which has the same lower bound because $V\le1$.
The reward-martingale variance identity sums these contributions to
$(1-e^{-r})s_k(x)(1-s_k(x))$.
\end{proof}

\begin{lemma}[Population optimization gives best-of-N improvement]
\label{lem:sc-population-progress}
Use either SC, $a(\rho)=1$, or SC+TIS, $a(\rho)=\min(\rho,c)$, and choose
\begin{equation}\label{eq:common-half-radius}
 0<r\le\frac{1}{8H},\qquad c\ge e^r\quad\text{for TIS}.
\end{equation}
Let $\Gamma_k$ be the neighborhood in \eqref{eq:theory-neighborhood}.
At every round, its prompt-averaged optimum satisfies
\begin{equation}\label{eq:sc-attainable-gain}
 \sup_{\pi\in\Gamma_k}\E_x\mathcal C_{k,x}(\pi)
 \ge(1-e^{-r})\E_x[s_k(x)(1-s_k(x))].
\end{equation}
For $N\ge1$ and $0<\epsilon<1$, use the averaged oracle
\eqref{eq:generalized-sc-oracle} at each round and choose
\begin{equation}\label{eq:sc-accumulated-error}
 K\ge\left\lceil\frac8r\log\frac{6N}{\epsilon}\right\rceil,
 \qquad
 \sum_{k=0}^{K-1}\zeta_k
 \le\frac{\epsilon}{16\pi^2NK}.
\end{equation}
Then
\begin{equation}\label{eq:generalized-sc-best-of-N}
 V_N(\pi_0)-J(\pi_K)\le\epsilon.
\end{equation}
The same conclusion holds in expectation if the oracle gaps and their
accumulated bound hold in expectation for random fitted policies.
\end{lemma}

\begin{proof}[Proof of Lemma~\ref{lem:sc-population-progress}]
\textbf{Step 1: attainable objective gain.}
The witness in Lemma~\ref{lem:common-progress-witness} attains
$(1-e^{-r})s_k(x)(1-s_k(x))$ at every prompt, proving
\eqref{eq:sc-attainable-gain} after averaging. For progress, we use the
full optimization-oracle condition in \eqref{eq:generalized-sc-oracle}.

\textbf{Step 2: local objective-to-reward comparison.}
Fix a round and a history, and write ${\mu}=\pi_k(\cdot\mid h)$,
$A=Q^{\mu}(h,\cdot)-V^{\mu}(h)$, and $a_h=\E_{\mu}|A|$.
For a feasible token policy $\pi$, let
\[
 c_h(\pi)=\E_{\mu}[A F_a(\pi/{\mu})],\qquad
 l_h(\pi)=\E_\pi A,\qquad
 \delta_h=\max_{v:\,|\log(v/{\mu})|\le r}c_h(v)-c_h(\pi).
\]
The maximum exists on the compact feasible token simplex, and
$\delta_h\ge0$. For SC or SC+TIS with cap at least one, the scalar
remainder bound gives
\[
 |l_h(\pi)-c_h(\pi)|\le B(r)r^2a_h,
 \qquad B(r)=(e^r-1-r)/r^2.
\]
Indeed, with $\rho=\pi/{\mu}$ and $g(\rho)=\rho-1-F_a(\rho)$, centering
$\E_{\mu}A=0$ gives $l_h(\pi)-c_h(\pi)=\E_{\mu}[A\,g(\rho)]$. For SC,
$g(\rho)=\rho-1-\log \rho\in[0,e^r-1-r]$ on $[e^{-r},e^r]$. For SC+TIS with
cap $c\ge1$, $g=0$ for $\rho\le c$, and
$0\le g(\rho)=\rho-c-c\log(\rho/c)\le \rho-1-\log \rho$ for $\rho>c$. Hence
$|\E_{\mu}[Ag]|\le(e^r-1-r)\max\{\E_{\mu}A_+,\E_{\mu}A_-\}=(e^r-1-r)a_h/2$.
To lower-bound the local optimum, put $b_h=\E_{\mu}\operatorname{sign}(A)$,
and choose
\[
 v(a)={\mu}(a)[1+\tfrac12(1-e^{-r})(\operatorname{sign}(A(a))-b_h)].
\]
This policy is normalized and its ratios lie in
$[e^{-r},2-e^{-r}]\subseteq[e^{-r},e^r]$.
Since $\E_{\mu} A=0$, it satisfies $l_h(v)=\tfrac12(1-e^{-r}) a_h$.
Applying the remainder bound to $v$ and to $\pi$ yields
\[
 l_h(\pi)\ge[\tfrac12(1-e^{-r})-2B(r)r^2]a_h-\delta_h
                  \ge\frac r4a_h-\delta_h.
\]
For the last inequality, use $r\le1/8$,
$\tfrac12(1-e^{-r})\ge r/2-r^2/4$, and $B(r)\le3/4$.

\textbf{Step 3: transfer along learner histories.}
The global objective is a sum of these local objectives weighted by
behavior-history probabilities. The full feasible neighborhood allows
each token policy to be chosen independently, so its optimization gap
is the corresponding sum of the nonnegative local gaps. Define
\[
 \Delta_k(x)=\sum_t\E_{h_t\sim d_t^{\mu}(\cdot\mid x)}\delta_{h_t}.
\]
The averaged oracle gives $\E_x\Delta_k(x)\le\zeta_k$.
The token-wise performance-difference identity, obtained by telescoping
$V_t^{\mu}$ along learner histories, is
\[
 s_{k+1}(x)-s_k(x)
 =\sum_t\E_{h_t\sim d_t^{\pi_{k+1}}(\cdot\mid x)}
 l_{h_t}(\pi_{k+1}(\cdot\mid h_t)).
\]
Lemma~\ref{lem:token-radius-sequence-ratio} bounds the prefix density
ratios between $4/5$ and $5/4$, since $e^{(H-1)r}\le e^{1/8}<5/4$. Therefore
\[
 s_{k+1}(x)-s_k(x)
 \ge\frac r5\sum_t\E_{d_t^{\mu}(\cdot\mid x)}a_{h_t}-\frac54\Delta_k(x)
 \ge\frac r5s_k(x)(1-s_k(x))-\frac54\Delta_k(x).
\]
Here $\E_{\mu}|A|\ge\E_{\mu} A^2$, and the reward-martingale total variance
is $\sum_t\E_{\mu} A_t^2=s_k(x)(1-s_k(x))$.
This step uses ordinary behavior histories, without conditioning on
the terminal reward.

\textbf{Step 4: accumulate the reward shortfalls.}
Since $r\le1/8$,
$\log(1+r/5)\ge(r/5)/(1+r/5)\ge r/8$.
The ideal log-odds increment $\lambda=r/8$ increases probability by at
most $(r/5)s_k(1-s_k)$. Its one-sided probability deficit is thus
at most $5\Delta_k(x)/4$. By the second deficit bound of
Lemma~\ref{lem:bernoulli-angular-transfer}, the squared angular
deficit is at most $5\pi^2\Delta_k(x)/16$. Applying
\eqref{eq:pointwise-angular-transfer} with accuracy $\epsilon/2$,
which the stated round count permits, and averaging prompts gives
\[
 V_N(\pi_0)-J(\pi_K)
 \le\epsilon/2+5\pi^2NK\sum_{k<K}\zeta_k
 \le\epsilon/2+8\pi^2NK\sum_{k<K}\zeta_k\le\epsilon.
\]
The argument holds for every fitted policy sequence. Taking expectations
proves the expected-oracle version.
\end{proof}

\subsubsection{One-round progress for unit-weight KL regularization}
\label{app:chenlu-pure-sc-kl}

For \MethodCurrentKL{}, set $a(\rho)=1$, $\beta=2$, and
$\Gamma_k=\Gamma_k(1/2)$ from Equation~\eqref{eq:theory-neighborhood}.
Its proof uses a direct reward lower bound and objective curvature,
rather than an explicit target formula. Theorem~\ref{thm:chenlu-sc-kl-convergence}
allows finite Monte Carlo centers; their fitting analysis is given in
Section~\ref{sec:unit-inner-fitting}.

At round $k$, hold $\mu=\pi_k$ fixed.
Write $\rho_t^\pi(h,u)=\pi(u\mid h)/{\mu}(u\mid h)$,
$k(v)=v\log v-v+1$, and $A_t^{\mu}=Q_t^{\mu}-V_t^{\mu}$. For each prompt define
\begin{align}
 C_{k,x}(\pi)&=\E_{\mu}\!\left[
       \sum_{t=1}^H A_t^{\mu}(h_t,a_t)\log \rho_t^\pi(h_t,a_t)
       \,\middle|\,x\right],\nonumber\\
 K_{k,x}(\pi)&=\sum_{t=1}^H\E_{h_t\sim d_t^{\mu}(\cdot\mid x)}
       \KL\!\left(\pi(\cdot\mid h_t)\|{\mu}(\cdot\mid h_t)\right),\nonumber\\
 W_{k,x}(\pi)&=C_{k,x}(\pi)-2K_{k,x}(\pi),
 \qquad \mathcal W_k(\pi)=\E_xW_{k,x}(\pi).
 \label{eq:chenlu-sc-kl-population}
\end{align}
Thus the KL coefficient is $\beta=2$. Equivalently, use
$\rho^{\mathrm{KL}}=b(\rho)=\min(\rho,2)$ in Equation~\eqref{eq:sc-kl-objective}:
its cap is inactive throughout $\Gamma_k$. The penalty is conditional
reverse KL averaged over behavior histories, not the KL between whole
response distributions.

\textbf{Population progress.}
Use $W_{k,x}$ and $\Gamma_k$ from
Section~\ref{app:chenlu-pure-sc-kl}, in particular
Equation~\eqref{eq:chenlu-sc-kl-population}. Fix a round and let $\pi_k^*$ maximize $W_{k,x}$ at every prompt over
$\Gamma_k$. Such an optimizer exists on each finite tree. The full
family separates across prompts and histories, so the conditional gap
\begin{equation}\label{eq:chenlu-sc-kl-gap}
 D_k(x,\pi)=W_{k,x}(\pi_k^*)-W_{k,x}(\pi)
\end{equation}
is nonnegative. We do not require a closed form for $\pi_k^*$.

The first lemma converts the regularized objective directly into reward gain.
\begin{lemma}[\MethodCurrentKL{} lower-bounds reward gain]
\label{lem:chenlu-sc-kl-lower-bound}
For terminal rewards in $[0,1]$, token ratios at least $e^{-r}$, and
$\beta\ge e^r$,
\[
 J_x(\pi)-J_x({\mu})\ge C_{k,x}(\pi)-\beta K_{k,x}(\pi).
\]
In particular, $J_x(\pi)-J_x({\mu})\ge W_{k,x}(\pi)$ on $\Gamma_k$.
\end{lemma}
\begin{proof}
Let $L(Y)=\pi(Y\mid x)/{\mu}(Y\mid x)$. Nonnegative reward and
$L-1\ge\log L$ imply
\begin{align*}
 J_x(\pi)-J_x({\mu})
 &\ge\E_{\mu}[R\log L\mid x]\\
 &=C_{k,x}(\pi)-\sum_t\E_{h_t\sim d_t^{\mu}(\cdot\mid x)}
           V_t^{\mu}(h_t)\KL({\mu}(\cdot\mid h_t)\|\pi(\cdot\mid h_t)).
\end{align*}
Here we condition each summand on its history and current action and
use $Q_t^{\mu}=A_t^{\mu}+V_t^{\mu}$. Since $0\le V_t^{\mu}\le1$, it remains to compare
the two conditional KL divergences. The functions
$f(u)=u-1-\log u$ and $k(u)=u\log u-u+1$ satisfy
$f(1)=f'(1)=k(1)=k'(1)=0$ and
$f''(u)\le e^r k''(u)$ for $u\ge e^{-r}$.
Taylor's integral remainder and normalization give
\[
 \KL({\mu}\|\pi)=\E_{\mu} f(\rho)
 \le e^r\E_{\mu} k(\rho)=e^r\KL(\pi\|{\mu}).
\]
Apply this at every history. For $r=1/2$, $\beta=2\ge e^{1/2}$.
\end{proof}

The next lemma constructs a feasible policy with positive objective gain.
Together with Lemma~\ref{lem:chenlu-sc-kl-lower-bound}, this makes
the objective gap a bound on lost reward progress.
\begin{lemma}[Constant reward progress from accurate fitting]
\label{lem:chenlu-sc-kl-progress}
Write $s_k(x)=J_x(\mu)$. Every $\pi\in\Gamma_k$ satisfies
\begin{equation}\label{eq:chenlu-sc-kl-one-round}
 J_x(\pi)-s_k(x)
 \ge\tfrac7{60} s_k(x)(1-s_k(x))-D_k(x,\pi).
\end{equation}
\end{lemma}
\begin{proof}
Consider $\widetilde\pi(u\mid h_t)={\mu}(u\mid h_t)(1+A_t^{\mu}(h_t,u)/4)$.
It is normalized because $\E_{{{\mu_t}}}A_t^{\mu}=0$, and feasible because its
ratios lie in $[3/4,5/4]\subset[e^{-1/2},e^{1/2}]$. For $|A|\le1$,
\[
 A\log(1+A/4)\ge A^2/5,
 \qquad k(1+A/4)\le A^2/24.
\]
The first inequality follows by integrating $1/(1+s)$ between $0$
and $A/4$. The second uses $k''(u)=1/u\le4/3$ on $[3/4,5/4]$.
On the deterministic response tree,
$A_t^{\mu}(h_t,a_t)=V_{t+1}^{\mu}(h_{t+1})-V_t^{\mu}(h_t)$ are martingale
differences ending at $R$. Hence
\[
 W_{k,x}(\widetilde\pi)
 \ge(1/5-1/12)\E_{\mu}\!\left[\sum_t(A_t^{\mu})^2\,\middle|\,x\right]
 =\tfrac7{60}\Var_{\mu}(R\mid x)=\tfrac7{60} s_k(x)(1-s_k(x)).
\]
The conditional objective of $\pi$ is at least that of
$\widetilde\pi$ minus $D_k(x,\pi)$. Apply
Lemma~\ref{lem:chenlu-sc-kl-lower-bound}.
\end{proof}

\subsubsection{Target progress and reward transfer for TIS with KL regularization}\label{sec:sc-kl}
We now specialize the objective in Equation~\eqref{eq:sc-kl-objective}
to cap-two reward and penalty weights,
$a(\rho)=b(\rho)=\min(\rho,2)$, as used by \MethodTISCurrentKL{}.
Theorem~\ref{thm:sc-kl-progress} gives its reward and sample bounds.
The unit-weight variant uses $a(\rho)=1$ and is treated in
Theorem~\ref{thm:chenlu-sc-kl-convergence}.

\textbf{The improving target.}
For the reward guarantee, choose $\beta=2$ and token log-ratio radius
$1/2$. Both caps are inactive throughout this family. The regularized
objective then balances the token advantage against conditional KL.
Its maximizing token policy is
\begin{equation}\label{eq:sc-kl-target}
 \pi_k^*(u\mid h_t)=
 \frac{\pi_k(u\mid h_t)\exp(Q_t^{\pi_k}(h_t,u)/\beta)}
 {\E_{v\sim\pi_k(\cdot\mid h_t)}\exp(Q_t^{\pi_k}(h_t,v)/\beta)}.
\end{equation}
The target gives more probability to tokens with higher reference
continuation value. It belongs to the feasible family because its
log-ratios are bounded by $1/\beta$. Fitting uses response rewards,
so the algorithm need not evaluate these continuation values.

The target improves success odds by at least a factor $1+1/\beta$
per round. To follow that improvement with an approximate fit, we
also need to control which histories the target visits. We show that
it can increase the probability of any prefix by at most $e^{1/\beta}$,
at every horizon. Jensen's inequality bounds each token normalizer,
and the reference continuation values telescope along the prefix.

The objective gap is characterized by conditional reverse KL
$\KL(\pi(\cdot\mid h_t)\|\pi_k^*(\cdot\mid h_t))$, averaged over
reference histories as specified in Equation~\eqref{eq:sc-kl-exact-gap}. The prefix bound lets
us convert this quantity to a divergence between full responses from
the two policies. Passing from responses to their binary rewards
then bounds the reward error. Lemmas~\ref{lem:rsc-target-gap}--\ref{lem:rsc-gap-transfer}
prove the target, coverage, and reward-transfer statements.

\textbf{The Q-function target and angular transfer.}
\label{app:sc-kl-progress}
The first lemma identifies the population optimizer and expresses its
objective gap as conditional KL. The second proves that this optimizer
improves reward, providing the ideal update for the outer analysis.
Write ${\mu}=\pi_k$, $\eta=1/\beta$, and initially allow any $\beta\ge2$.
Let $\Gamma_k(\eta)$ be the full family with token log-ratio radius
$r=\eta$. At $\beta=2$ it is the family $\Gamma_k$ of
Section~\ref{sec:sc-kl} with $r=1/2$. Let
$\mathcal W_{k,x}$ denote $\mathcal W_k$ conditional on the
prompt $x$, so that $\mathcal W_k=\E_x\mathcal W_{k,x}$.
\begin{lemma}[Feasible target and exact objective gap]\label{lem:rsc-target-gap}
For $\beta\ge2$, cap-two reward and penalty weights, and the full
radius-$1/\beta$ family, the target in \eqref{eq:sc-kl-target} is
feasible and maximizes the regularized objective at each prompt.
For every candidate $\pi$ in this family and every prompt,
\begin{equation}\label{eq:sc-kl-exact-gap}
 D_k(x):=\mathcal W_{k,x}(\pi_k^*)-
 \mathcal W_{k,x}(\pi)
 =\beta\sum_t\E_{h_t\sim d_t^{\mu}(\cdot\mid x)}
 \mathrm{KL}(\pi(\cdot\mid h_t)\Vert\pi_k^*(\cdot\mid h_t)).
\end{equation}
\end{lemma}
\begin{proof}
On $\Gamma_k(\eta)$ both caps are inactive. The local objective is
$\E_{\mu}[A(\rho-1)-\beta k(\rho)]$, with $k(\rho)=\rho\log \rho-\rho+1$.
Strict concavity and normalization give \eqref{eq:sc-kl-target}.
Since $Q\in[0,1]$, its ratios lie in $[e^{-\eta},e^\eta]$,
so the unconstrained optimum is feasible. For any fitted policy $\pi$
in this family, the Gibbs variational identity at each history gives
\[
 \E_{\pi_k^*}A-\beta\KL(\pi_k^*\|\mu)
 -\E_\pi A+\beta\KL(\pi\|\mu)
 =\beta\KL(\pi\|\pi_k^*).
\]
Averaging over behavior histories conditional on the prompt proves
Equation~\eqref{eq:sc-kl-exact-gap}.
The gap is nonnegative for each prompt.
\end{proof}

\begin{lemma}[Reward progress of the target]\label{lem:rsc-target-progress}
Write $s_k^*(x)=J_x(\pi_k^*)$ for the target success probability,
with $\pi_k^*$ defined in Equation~\eqref{eq:sc-kl-target} and
$s_k(x)=J_x(\pi_k)$ as in Section~\ref{sec:offpolicy-setting}.
For deterministic binary terminal rewards, the target satisfies
$\operatorname{logit}s_k^*(x)\ge\operatorname{logit}s_k(x)+\log(1+1/\beta)$.
\end{lemma}
\begin{proof}
Put $f(v)=(1+\eta)v/(1+\eta v)$. For any random variable
$Q\in[0,1]$ with mean $\overline Q$,
\[
 \E[e^{\eta Q}(f(Q)-f(\overline Q))]
 =\frac{1+\eta}{1+\eta \overline Q}
 \E\!\left[(Q-\overline Q)\frac{e^{\eta Q}}{1+\eta Q}\right]\ge0.
\]
The last expectation is nonnegative because
$e^{\eta v}/(1+\eta v)$ is nondecreasing, so its covariance with
$v$ is nonnegative. At terminal histories $f(R)=R$ for binary rewards.
Backward induction using the target's token probabilities therefore
proves $V^{\pi_k^*}(h)\ge f(V^{\mu}(h))$ at every history. In particular,
\[
 \operatorname{logit}s_k^*(x)\ge
 \operatorname{logit}s_k(x)+\log(1+1/\beta).
\]
Endpoint probabilities are interpreted by continuity.
\end{proof}

The target in Equation~\eqref{eq:sc-kl-target} has a prefix-probability
bound independent of the horizon. This lets us compare errors measured
under behavior histories with errors along target responses.
\begin{lemma}[Horizon-independent prefix and response coverage]\label{lem:rsc-prefix-coverage}
Let $\mu=\pi_k$ and let $\pi_k^*$ be the regularized TIS target in
Equation~\eqref{eq:sc-kl-target}, with $\beta\ge2$.
For every prompt and prefix, including terminal histories at $t=H+1$,
\begin{equation}\label{eq:sc-kl-target-coverage}
 \frac{d_t^{\pi_k^*}(h_t\mid x)}{d_t^{\mu}(h_t\mid x)}
 \le\exp\!\left(\frac{V_t^{\mu}(h_t)-V_1^{\mu}(x)}\beta\right)
 \le e^{1/\beta}.
\end{equation}
At a terminal history, $V_{H+1}^\mu=R$ and $V_1^\mu(x)=J_x(\mu)$, so
\begin{equation}\label{eq:sc-kl-target-response-coverage}
 \frac{\pi_k^*(Y\mid x)}{\mu(Y\mid x)}
 \le\exp\!\left(\frac{R(x,Y)-J_x(\mu)}\beta\right)
 \le e^{1/\beta}.
\end{equation}
At $\beta=2$, both prefix and response ratios are therefore at most
$e^{1/2}$, even though the target's token log-ratio radius is $1/2$.
\end{lemma}
\begin{proof}
Write $\eta=1/\beta$. Jensen's inequality gives
\[
 \log\frac{\pi_k^*(a\mid h)}{{\mu}(a\mid h)}
 =\eta Q^{\mu}(h,a)-\log\E_{u\sim {\mu}(\cdot\mid h)}e^{\eta Q^{\mu}(h,u)}
 \le\eta\bigl(Q^{\mu}(h,a)-V^{\mu}(h)\bigr).
\]
On the token tree, $Q_t^{\mu}(h_t,a_t)=V_{t+1}^{\mu}(h_{t+1})$.
Sum along the prefix. The value differences telescope, proving the
first inequality in \eqref{eq:sc-kl-target-coverage}. Values lie in $[0,1]$,
which proves its second inequality. At $t=H+1$, the prefix is the full
response and the terminal value is $R$, giving
Equation~\eqref{eq:sc-kl-target-response-coverage}.
\end{proof}

Combining this coverage bound with the exact gap in
Equation~\eqref{eq:sc-kl-exact-gap} controls the reward loss
accumulated across rounds.
\begin{lemma}[Objective gap controls accumulated reward error]
\label{lem:rsc-gap-transfer}
For $\beta\ge2$ and candidates in the radius-$1/\beta$ family,
\[
 \mathrm{KL}(\pi_k^*(Y\mid x)\Vert\pi(Y\mid x))
 \le e^{3/\beta}D_k(x)/\beta\le5D_k(x)/\beta.
\]
Consequently, the one-round reward shortfall satisfies
\[
 J_x(\pi_k^*)-J_x(\pi)\le\sqrt{\frac{e^{3/\beta}D_k(x)}{2\beta}}.
\]
If $K\ge\lceil2\beta\log(6N/\epsilon)\rceil$, the fitted sequence
satisfies
\begin{equation}\label{eq:sc-kl-capped-transfer}
 V_N(\pi_0)-J(\pi_K)
 \le\epsilon/2+\sum_{k<K}\E_x\mathcal L_k(x),\qquad
 \mathcal L_k(x)=\min\{1,\kappa D_k(x)\},\quad
 \kappa=\frac{10\pi^2NK}{\beta}.
\end{equation}
\end{lemma}
\begin{proof}
Both $\pi$ and $\pi_k^*$ have ratios to ${\mu}$ in
$[e^{-\eta},e^\eta]$. Comparing the second derivatives of the two
KL integrands on $[e^{-2\eta},e^{2\eta}]$ gives
$\mathrm{KL}(\pi_k^*\Vert\pi)\le
 e^{2\eta}\mathrm{KL}(\pi\Vert\pi_k^*)$ at every history.
The chain rule and \eqref{eq:sc-kl-target-coverage} imply
\[
 \mathrm{KL}(\pi_k^*(Y\mid x)\Vert\pi(Y\mid x))
 \le\frac{e^{3\eta}}{\beta}D_k(x).
\]
Pinsker's inequality bounds the reward difference by the square root of half the full-response KL, proving the one-round statement. For $\beta\ge2$, $e^{3/\beta}\le e^{3/2}<5$.
Take $\lambda=\log(1+1/\beta)$ in
Lemma~\ref{lem:bernoulli-angular-transfer}. The target satisfies
$\operatorname{logit}s_k^*\ge\operatorname{logit}s_k+\lambda$.
Data processing to the binary reward and the Bernoulli-KL deficit bound in
Lemma~\ref{lem:bernoulli-angular-transfer} bound the squared one-sided angular deficit $\xi_k(x)^2$ by
$5\pi^2D_k(x)/(8\beta)$.
With $s(x)=1-(1-s_0(x))^N$, the pointwise bound
\eqref{eq:pointwise-angular-transfer} at accuracy $\epsilon/2$ gives
\[
 s(x)-s_K(x)\le\epsilon/2+\min\left\{1,16NK\sum_{k<K}\xi_k(x)^2\right\}.
\]
Since $\min\{1,\sum_k y_k\}\le\sum_k\min\{1,y_k\}$ for $y_k\ge0$,
this implies
Equation~\eqref{eq:sc-kl-capped-transfer}.
The chosen $K$ suffices because
$\log(1+1/\beta)\ge1/(2\beta)$.
\end{proof}

\subsection{Inner fitting and statistical accuracy}\label{sec:theory-inner}
The inner solver uses a fixed round dataset. The following lemmas
connect its empirical accuracy to population accuracy by controlling
response noise within each prompt and variation between prompts.
The fitting certificates hold uniformly over candidates, so they also
apply to a policy chosen using that dataset.
\subsubsection{Empirical objectives and gradient identities}
\label{sec:generalized-sc}\label{app:algorithm-details}\label{app:sc-kl-solver}
\textbf{SC and SC+TIS optimize different objectives.}
Fix the behavior policy ${\mu}$. Write $\psi_\theta(h,a)$ for the learner's
log-policy gradient, and fix the reward weight $a(\rho)$ as defined
in Section~\ref{sec:theory-setup}. The weighted score is centered
under ${\mu}$ before it is multiplied by the response reward. The resulting
sampled update is
\begin{equation}\label{eq:generalized-sc-update}
 G_a(\theta,y)=R(y)\sum_{t=1}^H\left[
 a(\rho_{\theta,t}(h_t,a_t))\psi_{\theta,t}(h_t,a_t)
 -\E_{b\sim {\mu}(\cdot\mid h_t)}
   a(\rho_{\theta,t}(h_t,b))\psi_{\theta,t}(h_t,b)\right].
\end{equation}
Here $\rho_{\theta,t}(h,b)={\pi_\theta}(b\mid h)/{\mu}(b\mid h)$ and
$\psi_{\theta,t}=\nabla_\theta\log{\pi_\theta}$. The effective reward ratio is $\rho^{\mathrm{TIS}}=a(\rho)$: SC uses $\rho^{\mathrm{TIS}}=1$,
and SC+TIS uses $\rho^{\mathrm{TIS}}=\min(\rho,c)$ with cap $c\ge1$.
Each token in the conditional mean receives its own weight.

The potential $F_a$ and population objective $\mathcal C_a$ are defined
in Equations~\eqref{eq:generalized-sc-potential} and~\eqref{eq:generalized-sc-objective}.
In that objective, the behavior policy, its histories, and its
continuation advantages remain fixed. The expected centered update is
the gradient of this objective, including for shared nonlinear logits.
Proposition~\ref{prop:generalized-sc-sgd} proves the identity.
The objective can be estimated from rewards and token probabilities,
without evaluating continuation values. For a group of $G\ge2$ independent responses to the same prompt,
subtracting their mean reward uses the correction $G/(G-1)$ because
those responses also estimate their own baseline. We write
$\widehat{\mathcal C}_k$ for the resulting empirical objective at
round $k$. Equations~\eqref{eq:theory-empirical-objective} and~\eqref{eq:group-centered-update} give its exact form
and gradient.

\begin{proposition}[The population gradient of score centering]
\label{prop:generalized-sc-sgd}
Fix the behavior policy $\mu$, reward weight $a$, and a differentiable
full-support candidate $\pi_\theta$. With exact conditional centers,
the response update in Equation~\eqref{eq:generalized-sc-update} satisfies
\[
 \E_\mu G_a(\theta,Y)
 =\E_\mu\sum_t A_t^\mu a(\rho_{\theta,t})\psi_{\theta,t}
 =\nabla_\theta\mathcal C_a(\pi_\theta),
\]
whenever differentiation and expectation can be interchanged.
The expectation is for a fixed candidate, or one independent of the
sampled responses. On a fixed dataset, the grouped update instead gives
the gradient of the empirical objective, as stated in
Proposition~\ref{prop:sc-kl-sgd} with $\beta=0$.
\end{proposition}
\begin{proof}
Condition the sampled reward on $(h_t,a_t)$ in the first term of
Equation~\eqref{eq:generalized-sc-update}, and on $h_t$ in the centered
term. Their difference in conditional expectation is
$A_t^\mu a(\rho_t)\psi_{\theta,t}$. Since behavior and its values remain
fixed, $\rho F_a'(\rho)=a(\rho)$ and
$\nabla_\theta\rho=\rho\psi_{\theta,t}$ identify this expectation with
$\nabla_\theta\mathcal C_a$. The group-centering identity in
Appendix~\ref{app:group-centering} gives the same population expectation
for grouped responses.
\end{proof}

\paragraph{Sampling and group centering.} Exact conditional score centering and same-prompt reward
centering have separate roles: the first centers each token score,
while the second estimates a reward baseline from independent responses.
Conditional centering is exact in this analysis. An independently sampled,
conditionally unbiased Monte Carlo center adds variance without introducing
gradient bias. Systematic error in the center can bias the gradient.
For a fixed candidate, let $\widehat g$ average
Equation~\eqref{eq:generalized-sc-update} over $n$ independent responses.
If $\|\psi_{\theta,t}\|\le B$ and
$\E_{b\sim \mu(\cdot\mid h)}a(\rho_{\theta,t}(h,b))^2\le M_a$
uniformly, then
\begin{equation}\label{eq:generalized-sc-variance}
 \E\|\widehat g-\nabla\mathcal C_a\|^2
 \le\frac{4B^2H^2M_a}{n}.
\end{equation}
One may take $M_a=1$ for SC and $M_a=c$ for TIS, since
$\min(\rho,c)^2\le c\rho$ and $\E_{\mu}\rho=1$.
For the variance, put $X_t=a(\rho_t)\psi_t$. Conditional centering gives
$\E\|X_t-\E[X_t\mid h_t]\|^2\le B^2M_a$. Since $0\le R\le1$
and $\|\sum_t v_t\|^2\le H\sum_t\|v_t\|^2$, the second moment of a
response update is at most $B^2H^2M_a$. Independent averaging gives
\eqref{eq:generalized-sc-variance}, with its stated conservative factor four.

\textbf{Group centering.}
We can also subtract a reward baseline estimated from independent
responses to the same prompt. This group centering is in addition to
the conditional token-score centering in \eqref{eq:generalized-sc-update}.
For a prompt $x$, draw $G\ge2$ independent responses $y^{(g)}\sim {\mu}(\cdot\mid x)$.
For response $g$ and token position $t$, write
$h_t^{(g)}=(x,a_{<t}^{(g)})$ and define the centered weighted score
\[
 S_t^{(g)}=
 a\!\left(\rho_{\theta,t}(h_t^{(g)},a_t^{(g)})\right)
 \psi_{\theta,t}(h_t^{(g)},a_t^{(g)})
 -\E_{b\sim {\mu}(\cdot\mid h_t^{(g)})}\!\left[
 a\!\left(\rho_{\theta,t}(h_t^{(g)},b)\right)
 \psi_{\theta,t}(h_t^{(g)},b)\right].
\]
The expectation averages over the current token at the sampled history.
$g$ indexes responses and $t$ indexes tokens. The grouped update is
\begin{equation}\label{eq:group-centered-update}
 \widehat g_x^{\mathrm{grp}}
 =\frac1{G-1}\sum_{g=1}^G
   \bigl(R(y^{(g)})-\overline R\bigr)\sum_{t=1}^H S_t^{(g)},\qquad
 \overline R=\frac1G\sum_{g=1}^G R(y^{(g)}).
\end{equation}
The factor $1/(G-1)$ removes the bias from using the same group to
estimate the baseline and the update. Since $\E[S_t\mid x]=0$,
the grouped estimator has the same conditional expectation as the
original update. Averaging independent groups with prompts drawn from
$\mathcal D$ therefore preserves the objective in
\eqref{eq:generalized-sc-objective}. Both the group baseline and the
sampled returns are detached during differentiation.

Appendix~\ref{app:sc-variance} proves this identity and bounds the
within-prompt variance by $4B^2H^2M_a/(G-1)$ under the score and weight-moment assumptions of \eqref{eq:generalized-sc-variance}. Random prompt sampling also contributes between-prompt variance.

For $\beta=0$, Equation~\eqref{eq:theory-empirical-objective} gives
$\widehat{\mathcal C}_k$, whose gradient is the average of
Equation~\eqref{eq:group-centered-update} over the sampled prompts.
The dataset stays fixed during fitting; minibatches estimate the gradient
of this empirical objective.

\textbf{Regularized updates and grouped fitting.}
Write $\psi_{\theta,t}(h,u)=\nabla_\theta\log \pi_\theta(u\mid h)$ for the
score from Equation~\eqref{eq:method-score-ratio}. For a sampled response
with reward $R=R(Y)$, use the main-text weights
$\rho_t^{\mathrm{TIS}}=a(\rho_t)$ and
$\rho_t^{\mathrm{KL}}=b(\rho_t)$. The response update is
\begin{equation}\label{eq:sc-kl-update}
 \begin{aligned}
 G_k(\theta,Y)=\sum_t\Big[&R\bigl(
 \rho_t^{\mathrm{TIS}}\psi_{\theta,t}(h_t,a_t)
 -\E_{u\sim \mu(\cdot\mid h_t)}[
 \rho_t^{\mathrm{TIS}}(h_t,u)\psi_{\theta,t}(h_t,u)]\bigr)\\
 &-\beta\rho_t^{\mathrm{KL}}\log\rho_t^{\mathrm{KL}}
 \psi_{\theta,t}(h_t,a_t)\Big].
 \end{aligned}
\end{equation}
The sampled reward $R(Y)$ is held fixed inside the token expectation.
The reward term is score-centered. The penalty coefficient satisfies
$-1/e\le b(\rho)\log b(\rho)\le c_b\log c_b$.

\textbf{Group centering.}
For a prompt $x$, draw $G\ge2$ independent responses
$y^{(g)}\sim {\mu}(\cdot\mid x)$ and define the corrected centered rewards
\[
 \overline R=\frac1G\sum_{g=1}^G R(y^{(g)}),\qquad
 \widetilde R^{(g)}=\frac{G}{G-1}\bigl(R(y^{(g)})-\overline R\bigr).
\]
For response $g$, let $h_t^{(g)}$ be its history and abbreviate
$\psi_t^{(g)}=\psi_{\theta,t}(h_t^{(g)},a_t^{(g)})$ and
$\rho_t^{(g)}=\rho_t(h_t^{(g)},a_t^{(g)})$.
The grouped update is
\begin{equation}\label{eq:sc-kl-group-update}
 \begin{aligned}
 \widehat g_{k,x}^{\mathrm{grp}}=\frac1G\sum_{g,t}\Big[&
 \widetilde R^{(g)}\bigl(
 a(\rho_t^{(g)})\psi_t^{(g)}
 -\E_{u\sim \mu(\cdot\mid h_t^{(g)})}[
 a(\rho_t(h_t^{(g)},u))\psi_{\theta,t}(h_t^{(g)},u)]\bigr)\\
 &-\beta b(\rho_t^{(g)})\log b(\rho_t^{(g)})\psi_t^{(g)}\Big].
 \end{aligned}
\end{equation}
\begin{proposition}[The empirical objective and its stochastic gradient]
\label{prop:sc-kl-sgd}
Let $\pi_\theta$ be differentiable with full support, and hold behavior
$\mu=\pi_k$ fixed. With exact conditional centers, for the empirical
objective in Equation~\eqref{eq:theory-empirical-objective} and grouped
update in Equation~\eqref{eq:sc-kl-group-update},
\[
 \nabla_\theta\widehat{\mathcal W}_k(\pi_\theta)
 =\frac1m\sum_{j=1}^m\widehat g_{k,x_j}^{\mathrm{grp}}.
\]
For each fixed $\theta$, independent prompts and conditionally independent
responses give, whenever differentiation and expectation can be interchanged,
\[
 \E\nabla_\theta\widehat{\mathcal W}_k(\pi_\theta)
 =\E_{\mu}G_k(\theta,Y)=\nabla_\theta\mathcal W_k(\pi_\theta).
\]
The empirical identity holds on the fixed dataset, including when that
dataset is reused for fitting. The expectation identity is for a fixed
candidate, or for a candidate independent of the fresh data.
Replacing exact centers by conditionally unbiased estimates, sampled independently of the response actions and rewards given the visited history, preserves the population expectation identity. The empirical identity above uses exact centers.
\end{proposition}

The centered reward $\widetilde R^{(g)}$ is held fixed inside the
token expectation. As in \eqref{eq:group-centered-update}, the factor
$G/(G-1)$ corrects the bias from estimating the baseline with the same
group. The penalty is averaged over responses with coefficient $\beta$. Averaging independent prompt groups gives an unbiased gradient estimator
of $\mathcal W_k$. Equation~\eqref{eq:sc-kl-update} also gives an
unbiased single-response gradient. The grouped objective and convergence
theorems use $G\ge2$. When implementing the update
through weighted log-policy gradients, detach the reward baseline and
both score coefficients. The following sample objective gives a direct
proof of these identities.

For the proofs, write $\psi_{\theta,t}(h,u)=\nabla_\theta\log{\pi_\theta}(u\mid h)$.
Using $F_a$ from Equation~\eqref{eq:generalized-sc-potential} and
$K_b$ from Equation~\eqref{eq:sc-penalty-potential}, define
\[
 z_{a,t}(h,u)=F_a(\rho_t(h,u))-
\E_{v\sim {\mu}(\cdot\mid h)}F_a(\rho_t(h,v)).
\]
An unbiased sample objective is
\begin{equation}\label{eq:sc-kl-sample-objective}
 \mathcal H_k(\pi,Y)=R(Y)\sum_t z_{a,t}(h_t,a_t)
                         -\beta\sum_t K_b(\rho_t(h_t,a_t)).
\end{equation}
\begin{proof}[Proof of Proposition~\ref{prop:sc-kl-sgd}]
Because behavior is fixed, $\nabla_\theta \rho_t=\rho_t\psi_{\theta,t}$.
The identities $\rho F_a'(\rho)=a(\rho)$ and $\rho K_b'(\rho)=b(\rho)\log b(\rho)$ give
\[
 \nabla_\theta z_{a,t}=a(\rho_t)\psi_{\theta,t}-\E_{{\mu_t}}a(\rho_t)\psi_{\theta,t},
 \qquad \nabla_\theta K_b(\rho_t)=b(\rho_t)\log b(\rho_t)\psi_{\theta,t}.
\]
Differentiating \eqref{eq:sc-kl-sample-objective} proves the exact
sample-gradient identity
$\nabla_\theta \mathcal H_k({\pi_\theta},Y)=G_k(\theta,Y)$.
Conditional advantage centering makes the
reward potential unbiased for $\mathcal C_k$, and the penalty is
averaged under the same fixed behavior distribution. This proves the
population identity. Same-prompt group centering preserves the reward
expectation by Appendix~\ref{app:group-centering}. Response averaging
preserves the penalty expectation and its gradient. Specifically, for
$G\ge2$, use the unbiased same-prompt sample covariance of $R$ with
$\sum_t z_{a,t}$ for the reward term and the response average for the
penalty. Averaging these quantities over $m$ prompts is precisely
$\widehat{\mathcal W}_k$ in
Equation~\eqref{eq:theory-empirical-objective}. Differentiating this
objective gives the average of Equation~\eqref{eq:sc-kl-group-update}
over prompts.
For independent conditionally unbiased center estimates, conditioning on the response and then averaging the auxiliary draws recovers the exact-center update. The tower property therefore preserves the population gradient identity.
\end{proof}

\textbf{Properties of the penalty.}
Monotonicity and $b(1)=1$ imply $b(\rho)\log b(\rho)\le0$ for $\rho<1$
and $b(\rho)\log b(\rho)\ge0$ for $\rho>1$. Integrating from $1$ therefore
gives $K_b(\rho)\ge0$. The bound on $b\log b$ follows from the minimum
of $x\log x$ at $x=e^{-1}$ and its monotonicity for $x\ge1$.
If $b$ is differentiable at $1$, then $K_b'(1)=0$ and
$K_b''(1)=b'(1)$. Thus
$K_b(\rho)=\tfrac12 b'(1)(\rho-1)^2+o((\rho-1)^2)$. Choosing $b'(1)=1$
matches KL's local quadratic form.
For $b(\rho)=\min(\rho,c_b)$, integrate separately below and above the cap
to obtain
\[
 K_b(\rho)=\begin{cases}
 \rho\log \rho-\rho+1,&\rho\le c_b,\\
 c_b\log c_b-c_b+1+c_b\log c_b\log(\rho/c_b),&\rho>c_b.
 \end{cases}
\]
Above the cap, its score coefficient is $c_b\log c_b$.
The proof of Lemma~\ref{lem:rsc-response-moments} uses the interval on which $b(\rho)=\rho$.

\subsubsection{Variance and sample complexity for SC variants}\label{app:sc-variance}
We begin with variance bounds for the SC and TIS score estimators.
These also describe the reward component of RSC; the regularized
objective bounds below include the penalty. For a
fixed conditional centering policy $\tau_t(\cdot\mid h)$ and a token
weight $\chi_t(h,a)$, consider one behavior response's contribution
\[
 X(\theta)=\sum_t \chi_t(h_t,a_t)U_t
 \left[\nabla_\theta\log{\pi_\theta}(a_t\mid h_t)
 -\sum_a \tau_t(a\mid h_t)\nabla_\theta\log{\pi_\theta}(a\mid h_t)\right].
\]
Suppose the logits satisfy $\|\nabla_\theta f_\theta(h,a)\|\le B$,
$|U_t|\le M$, and the token weights have bounded conditional second moments.
For $n$ independent responses and their average $\widehat g$,
\begin{equation}\label{eq:sc-variance}
 \E\|\widehat g-\E\widehat g\|^2
 \le\frac{4B^2M^2H^2}{n}
 \sup_{t,h}\E_{a\sim {\mu}(\cdot\mid h)}\chi_t(h,a)^2.
\end{equation}
Indeed the centered token score is
$\nabla f_\theta(h,a)-\sum_u \tau_t(u\mid h)\nabla f_\theta(h,u)$,
whose norm is at most $2B$. Apply
$\|\sum_t X_t\|^2\le H\sum_t\|X_t\|^2$, then average over independent responses. No independence between tokens of one response is assumed.
For SC, $\chi_t=1$, $U_t=R$, and $M=1$.
For TIS, $\chi_t=\min(\rho_t,c)$ and $\E_\mu\chi_t^2\le c\E_\mu\rho_t=c$, giving the corresponding second-moment constant.
Appendix~\ref{app:sc-kl-samples} treats the full \MethodTISCurrentKL{}
objective, including its sample penalty.

For the centered weighted TIS score, use the moment calculation
following Proposition~\ref{prop:generalized-sc-sgd}, with $M_a=c$ and
$\|\nabla_\theta\log\pi_\theta\|\le2B$. This gives a bound of $16B^2H^2c/n$ when $B$ bounds
the logit gradient.

\textbf{Same-prompt group centering.}
\label{app:group-centering}
Fix the prompt and current iterate, and draw $G\ge2$ independent
responses. Write $\overline U_t=G^{-1}\sum_g U_t^{(g)}$ for their
mean token return and $S_t^{(g)}$ for response $g$'s centered token
score, as in Equation~\eqref{eq:group-centered-update}.
For that equation $U_t^{(g)}=R(y^{(g)})$. Suppose each token score satisfies $\E[S_t\mid x]=0$.
For the estimator in \eqref{eq:group-centered-update}, independence
between distinct responses gives
\[
 \E[\overline U_t S_t^{(g)}\mid x]
 =\frac1G\E[U_tS_t\mid x].
\]
Summing over responses and dividing by $G-1$ proves
\[
 \E[\widehat g_x^{\mathrm{grp}}\mid x]
 =\E\left[\sum_t U_tS_t\,\middle|\,x\right].
\]
Thus group centering preserves the generalized SC reward gradient.
For the regularized objective in Proposition~\ref{prop:sc-kl-sgd},
group-center this reward term and average the $b(\rho)\log b(\rho)$ penalty gradient
directly. This preserves the gradient of the stated empirical objective.

For a variance bound, assume $|U_t|\le M$ and
$\E[\|S_t\|^2\mid x]\le\sigma_S^2$ uniformly. The grouped estimator is
the average over all unordered response pairs of the symmetric kernel
\[
 \mathcal{H}(y,y')=\frac12\sum_t (U_t(y)-U_t(y'))(S_t(y)-S_t(y')).
\]
The two responses are independent, and their scores have mean zero.
Consequently,
$\E[\|\mathcal{H}\|^2\mid x]\le2H^2M^2\sigma_S^2$.
To bound the variance of the pair average, average over random pairings
of the $G$ responses, leaving one response unused if $G$ is odd.
Each pairing contains $\lfloor G/2\rfloor$ independent pairs. Its
conditional expectation given the responses is the full pair average.
Conditional Jensen's inequality therefore gives
\begin{equation}\label{eq:group-centered-variance}
 \E\bigl[\|\widehat g_x^{\mathrm{grp}}
       -\E[\widehat g_x^{\mathrm{grp}}\mid x]\|^2\mid x\bigr]
 \le\frac{4H^2M^2\sigma_S^2}{G-1}.
\end{equation}
For generalized SC, $M=1$ and
$\sigma_S^2=B^2M_a$ when $B$ bounds the token score norm.
The separate penalty term is not part of this centered-score bound.
Appendix~\ref{app:sc-kl-samples} bounds the full regularized objective.
Averaging $m$ independent prompt groups divides the variance by $m$.
The unconditional variance also includes
$\operatorname{Var}_{x}(\E[\widehat g_x^{\mathrm{grp}}\mid x])/m$,
where vector variance denotes expected squared distance from the mean.
This separates response noise from variation between prompts.
The group baseline removes a constant return exactly within a group,
but its estimation noise can increase variance for other return
and score distributions.

\textbf{Statistical fitting from the radius margin.}
\label{app:sc-radius-margin}
Theorem~\ref{thm:sc-sample-complexity} combines the radius margin
with separate concentration bounds for prompts and responses.
Throughout the radius-margin argument, ${\mu}=\pi_k$, $0<r\le1/(8H)$, and the TIS cap satisfies $c\ge e^r$.
For a fixed prompt define
\[
 a_x=\sum_t\E_{h_t\sim d_t^{\mu}(\cdot\mid x)}
                \E_{b\sim {\mu}(\cdot\mid h_t)}|A_t^{\mu}(h_t,b)|,
 \qquad M_x(s)=\sup_{\pi\in\Gamma_k(s)}\mathcal C_{k,x}(\pi).
\]
Write $W=1$ for SC and $W=e^r$ for TIS. Under $c\ge e^r$, the
TIS potential is $\rho-1$ throughout the feasible interval.

\begin{lemma}[Gain margin between two radii]\label{lem:sc-radius-margin}
For SC or inactive-cap TIS at $0<r\le1/(8H)$,
$M_x(r/2)\ge7ra_x/32$, $M_x(r)\ge\tfrac32M_x(r/2)$, and
$M_x(r)-M_x(r/2)\ge ra_x/16$ at every prompt.
\end{lemma}
\begin{proof}
At one history the sign-based policy from the proof of
Lemma~\ref{lem:sc-population-progress}, at radius $s=r/2$, has
linear advantage $(1-e^{-s})\E_{\mu}|A|/2$. The scalar remainder is at
most $B(s)s^2\E_{\mu}|A|$, where $B(s)\le3/4$. Hence
\[
 M_x(r/2)\ge(r/4-r^2/4)a_x\ge\frac{7r}{32}a_x.
\]
The full class permits independent token choices, so the local
bounds can be summed. For SC, raise any feasible radius-$r/2$ token
ratio $\rho$ to the power $3/2$ and normalize. Since
$1\le\E_{\mu}\rho^{3/2}\le e^{r/4}$, the new log ratios lie in $[-r,3r/4]$.
Conditional advantage centering cancels the normalization term and
multiplies the objective by $3/2$. For TIS use the normalized ratio
$1+\tfrac32(\rho-1)$. It lies in $[e^{-r},e^r]$ for $r\le1/8$, and
also multiplies the objective by $3/2$. Consequently,
\begin{equation}\label{eq:sc-derived-radius-margin}
 M_x(r)\ge\frac32M_x(r/2),\qquad
 M_x(r)-M_x(r/2)\ge\frac r{16}a_x.
\end{equation}

\end{proof}

\begin{lemma}[Conditional moments of objective differences]\label{lem:sc-pair-moment}
For two radius-$r$ candidates, let $Z$ be the difference of their
centered potentials. The pair kernel
$\tfrac12(R-R')(Z-Z')$ has conditional second moment at most
$36HW^2r^2a_x$. The complete grouped average obeys the same bound.
\end{lemma}
\begin{proof}
For any two radius-$r$ policies, let $Z$ be the difference of their
centered potentials and write $Z_t=\sum_{\tau\le t}\xi_\tau$.
The increments have conditional mean zero, absolute value at most
$4Wr$, and conditional second moment at most $4W^2r^2$.
Let $s_k(x)=\E_{\mu}[R\mid x]$. Telescoping $Z_t^2$, then conditioning on
history and action, gives
\[
 \E_{\mu}[RZ^2\mid x]
 =\sum_t\E_{\mu}[Q_t^{\mu}(2Z_{t-1}\xi_t+\xi_t^2)\mid x]
 \le16HW^2r^2s_k(x)+32HW^2r^2a_x.
\]
For the cross term, conditional centering replaces $Q_t^{\mu}$ by
$A_t^{\mu}$, and $|Z_{t-1}|\le4Wr(t-1)$. Replacing $R$ by $1-R$ gives
the analogous bound with $s_k(x)$ replaced by $1-s_k(x)$. Also
$\E_{\mu}Z^2\le4HW^2r^2$.
For two conditionally independent responses define
$D(Y,Y')=\tfrac12(R(Y)-R(Y'))(Z(Y)-Z(Y'))$.
Since $\E_{\mu}Z=0$ and $R$ is binary,
\[
 \E[D^2\mid x]
 =\frac12\bigl[(1-s_k(x))\E(RZ^2\mid x)
                  +s_k(x)\E((1-R)Z^2\mid x)\bigr]
       +(\E[RZ\mid x])^2
 \le36HW^2r^2a_x.
\]
Here covariance Cauchy-Schwarz gives
$(\E RZ)^2\le4HW^2r^2s_k(x)(1-s_k(x))$, and the reward-martingale identity
implies $s_k(x)(1-s_k(x))\le a_x$. The same second-moment bound holds for
the complete group average by Jensen's inequality.

\end{proof}

\begin{lemma}[A fitting margin controls response noise]\label{lem:sc-margin-noise}
For each candidate $\pi$, the full feasible family contains a comparison
policy $\nu_\pi$ whose conditional objective gain $d_\pi$ dominates the
positive shortfall $\Delta_\pi$ from the radius-$r/2$ optimum. Their pair
objective difference has conditional second moment at most $96d_\pi$
and absolute value less than one.
\end{lemma}
\begin{proof}
Choose a radius-$r$ policy $\pi_r^*$ maximizing the objective at each
prompt. For a candidate $\pi\in\Gamma_k(r)$, let $\nu_\pi$ use
$\pi_r^*$ on prompts where $\mathcal C_{k,x}(\pi)<M_x(r/2)$,
and use $\pi$ elsewhere. The full token family contains $\nu_\pi$.
Write
\begin{equation}\label{eq:sc-margin-shortfall}
 \Delta_\pi(x)=[M_x(r/2)-\mathcal C_{k,x}(\pi)]_+,\qquad
 d_\pi(x)=\mathcal C_{k,x}(\nu_\pi)-\mathcal C_{k,x}(\pi).
\end{equation}
Let $\mathcal H_\pi(Y,Y')$ be the pair objective difference between
$\nu_\pi$ and $\pi$. On a prompt with a shortfall, the radius margin
gives $d_\pi(x)\ge ra_x/16$. Elsewhere the pair difference is zero.
Lemma~\ref{lem:sc-pair-moment} therefore gives, at every prompt,
\[
 d_\pi(x)\ge\Delta_\pi(x)\ge0,\qquad
 \E[\mathcal H_\pi^2\mid x]\le576W^2Hr\,d_\pi(x)\le96d_\pi(x).
\]
Here $Hr\le1/8$ and $W^2\le e^{1/4}<4/3$.
Also $|\mathcal H_\pi|\le4HWr<1$, so the centered pair difference has
absolute value less than two.

\end{proof}

\begin{lemma}[Promptwise fitting shortfall controls reward]\label{lem:sc-shortfall-transfer}
For $r\le1/(8H)$, a feasible candidate satisfies
$J_x(\pi)-s_k(x)\ge(r/32)s_k(x)(1-s_k(x))-\Delta_\pi(x)$.
Over $K\ge(64/r)\log(6N/\epsilon)$ rounds,
\begin{equation}\label{eq:sc-capped-transfer}
 V_N(\pi_0)-J(\pi_K)
 \le\epsilon/2+\sum_{k<K}\E_x\mathcal L_k(x),\qquad
 \mathcal L_k(x)=\min\{1,\kappa\Delta_{\pi_{k+1}}(x)\},\quad
 \kappa=4\pi^2NK.
\end{equation}
\end{lemma}
\begin{proof}
At a history put $l_h=\E_{\pi(\cdot\mid h)}A_t^{\mu}$ and let $c_h$
be its SC or TIS objective contribution. Then
$|l_h-c_h|\le B(r)r^2\E_{\mu}|A_t^{\mu}|$ and
$|l_h|\le(e^r-1)\E_{\mu}|A_t^{\mu}|$.
The learner-to-behavior history ratio differs from one by at most
$e^{(H-1)r}-1$. The performance-difference identity gives
\[
 J_x(\pi)-s_k(x)\ge\mathcal C_{k,x}(\pi)
 -\left[B(r)r^2+(e^{(H-1)r}-1)(e^r-1)\right]a_x.
\]
For $0\le u\le1/8$, $e^u-1\le8u/7$, and $B(r)\le3/4$.
The bracket is at most $(4/3)Hr^2\le r/6$. Combining this with
$M_x(r/2)\ge7ra_x/32$ yields
\[
 J_x(\pi)-s_k(x)\ge\frac r{32}s_k(x)(1-s_k(x))-\Delta_\pi(x).
\]
The ideal log-odds increment $r/64$ improves success probability by
at most the positive term. By the second deficit bound of
Lemma~\ref{lem:bernoulli-angular-transfer}, its squared one-sided
angular shortfall is therefore at most $\pi^2\Delta_\pi(x)/4$.
With $K\ge(64/r)\log(6N/\epsilon)$, the pointwise bound
\eqref{eq:pointwise-angular-transfer} at accuracy $\epsilon/2$ gives
Equation~\eqref{eq:sc-capped-transfer}.
Indeed, the second form of \eqref{eq:pointwise-angular-transfer}
bounds the accumulated angular error by
$4\pi^2NK\sum_k\Delta_{\pi_{k+1}}(x)$, and
$\min\{1,\sum_k y_k\}\le\sum_k\min\{1,y_k\}$ for $y_k\ge0$.

\end{proof}

The next lemma turns an empirical fitting tolerance into the capped
prompt loss used by Lemma~\ref{lem:sc-shortfall-transfer}.
\begin{lemma}[Uniform fitting certificate for unregularized SC and TIS]
\label{lem:sc-uniform-fitting}
Use the radius-$r$ family and weights of Lemma~\ref{lem:sc-margin-noise}.
Let $L=\lfloor G/2\rfloor$, $\kappa=4\pi^2NK$, and
$\ell=\log(4K\mathcal N(u)^2/\delta)$ for a radius-$u$ cover in
Equation~\eqref{eq:sc-cover-metric}. Define
$\mathcal L_\pi(x)=\min\{1,\kappa\Delta_\pi(x)\}$, where
$\Delta_\pi$ is defined in Equation~\eqref{eq:sc-margin-shortfall}.
Conditional on the past, with probability at least $1-\delta/K$,
every empirical fit $\widehat\pi$ with full-family error at most $e_k$ satisfies
\begin{equation}\label{eq:sc-prompt-margin-certificate}
 \E_x\mathcal L_{\widehat\pi}(x)
 \le4\kappa e_k+\frac{392\kappa\ell}{mL}
       +\frac{4\ell}{m}+120\kappa Hu.
\end{equation}
\end{lemma}
\begin{proof}
Fix one round and condition on its reference. Put
$L=\lfloor G/2\rfloor$. For a candidate and its comparison policy,
write
\[
 \overline d_\pi=\frac1m\sum_jd_\pi(x_j),\qquad
 \widehat d_\pi=\widehat{\mathcal C}_k(\nu_\pi)
                  -\widehat{\mathcal C}_k(\pi).
\]
Conditional on the sampled prompts, randomly pair the responses in
each group. The actual group estimator is the conditional expectation
of the paired estimator. Exponential Jensen and one-sided Bernstein
for the $mL$ independent pairs yield
\[
 \overline d_\pi-\widehat d_\pi
 \le\sqrt{\frac{192\overline d_\pi\ell}{mL}}
         +\frac{4\ell}{3mL},\qquad
 \overline d_\pi\le2\widehat d_\pi+\frac{196\ell}{mL}.
\]
For the capped prompt loss $\mathcal L_\pi(x)=\min\{1,\kappa\Delta_\pi(x)\}$,
write $P_\pi=\E_x\mathcal L_\pi$ and
$\widehat P_\pi=m^{-1}\sum_j\mathcal L_\pi(x_j)$.
Since $\operatorname{Var}_x\mathcal L_\pi\le P_\pi$, prompt-level Bernstein
gives $P_\pi\le2\widehat P_\pi+4\ell/m$.

A radius-$u$ cover of $\Gamma_k(r)$ gives at most
$\mathcal N(u)^2$ cells for the pairs $(\nu_\pi,\pi)$.
Choose one actual pair in each nonempty cell. Each component lies
within distance $2u$ of its representative. The group objective and
conditional objective are $2HW$-Lipschitz. The shortfall has the same
Lipschitz bound. With
\[
 \ell=\log\frac{4K\mathcal N(u)^2}{\delta},
\]
a union bound and extension from the pair representatives give
\[
 \overline d_\pi\le2\widehat d_\pi+
          \frac{196\ell}{mL}+24HWu,\qquad
 P_\pi\le2\widehat P_\pi+\frac{4\ell}{m}+12\kappa HWu.
\]
Because $\Delta_\pi\le d_\pi$, we have
$\widehat P_\pi\le\kappa\overline d_\pi$.
The empirical oracle and $\nu_{\widehat\pi}\in\Gamma_k(r)$ give
$\widehat d_{\widehat\pi}\le e_k$, where $e_k$ is its empirical fitting tolerance. Since $W<2$,
this proves Equation~\eqref{eq:sc-prompt-margin-certificate}.
\end{proof}

\textbf{Covering-number control of empirical fitting.}
\label{app:sc-finite-oracle}
The following lemma treats a restricted model class, as an alternative to
fitting against the full family in Theorem~\ref{thm:sc-sample-complexity}.
\begin{lemma}[Additive fitting over a model class]\label{lem:sc-additive-fitting}
Use the radius-$r$ family, with $W_a=1$ for SC and
$W_a=\min\{c,e^r\}$ for TIS. Let $\mathcal P_k\subseteq\Gamma_k$ be
the policies represented by the model. In this lemma,
$\mathcal N_k(u)$ is the size of a radius-$u$ cover of $\mathcal P_k$
in $d_\infty$ with centers in $\mathcal P_k$. It is at most the size
of a radius-$u/2$ cover of $\Gamma_k$. Define the
representation gap and the empirical optimization gap of the fitted
$\widehat\pi\in\mathcal P_k$ by
\[
 b_k=\sup_{\pi\in\Gamma_k}\mathcal C_k(\pi)
     -\sup_{\pi\in\mathcal P_k}\mathcal C_k(\pi),
 \qquad
 e_k=\sup_{\pi\in\mathcal P_k}\widehat{\mathcal C}_k(\pi)
     -\widehat{\mathcal C}_k(\widehat\pi).
\]
Then, with probability at least $1-\delta_k$, the population gap
$\sup_{\Gamma_k}\mathcal C_k-\mathcal C_k(\widehat\pi)$ is at
most $b_k+e_k$ plus $2\times$ the uniform error in \eqref{eq:sc-additive-uniform-error}.
\end{lemma}

\begin{proof}
Write $W=W_a$. One prompt group's contribution to the empirical
objective is the complete average of the pair kernel
\[
 \mathcal H_\pi(y,y')=\frac12(R(y)-R(y'))(Z_\pi(y)-Z_\pi(y')).
\]
Conditional centering gives $\E Z_\pi=0$, so its expectation is
$\mathcal C_k(\pi)$. The potential is $W$-Lipschitz in the token
log ratio. Consequently the group contribution has absolute value
at most $B_{\mathrm{pair}}=2HWr$, and is $2HW$-Lipschitz in $d_\infty$.
Under behavior, the centered token potentials form a martingale
difference sequence. The conditional range bound and orthogonality give
$\E_{\mu}[Z_\pi^2\mid x]\le HW^2r^2$. Since rewards are binary,
\[
 \operatorname{Var}_x(\E[\mathcal H_\pi\mid x])\le\frac{HW^2r^2}{4},
 \qquad
 \E_x\operatorname{Var}(\mathcal H_\pi\mid x)\le\frac{HW^2r^2}{2}.
\]
The first inequality follows from the reward covariance identity and
$\operatorname{Var}(R\mid x)\le1/4$. The second uses
$|R-R'|\le1$ and independence of $Z_\pi,Z_\pi'$ at a fixed prompt.
Random pairings of the $G$ responses therefore give group variance
at most
\[
 \frac{HW^2r^2}{4}\left(1+\frac2{\lfloor G/2\rfloor}\right).
\]
Put $\ell=\log(4\mathcal N_k(u)/\delta_k)$. Bernstein's inequality,
a union bound over the cover, and the Lipschitz extension give a uniform error bound. The extension term is $4HWu$ because both
$\widehat{\mathcal C}_k$ and $\mathcal C_k$ are
$2HW$-Lipschitz:
\begin{equation}\label{eq:sc-additive-uniform-error}
 \sup_{\pi\in\mathcal P_k}
 |\widehat{\mathcal C}_k(\pi)-\mathcal C_k(\pi)|
 \le 4HWu+
 Wr\sqrt{\frac{H}{2m}\left(1+\frac2{\lfloor G/2\rfloor}\right)\ell}
 +\frac{8HWr\ell}{3m}.
\end{equation}
The full population gap is at most $b_k+e_k$ plus $2\times$ the uniform
error in \eqref{eq:sc-additive-uniform-error}. Uniformity permits the
fitted policy to depend on all observations. Conditioning on earlier
rounds and taking $\delta_k=\delta/K$ gives a simultaneous guarantee.
\end{proof}

\textbf{A sufficient prompt count.}
For oracle tolerance $\zeta_k$, suppose $b_k+e_k\le\zeta_k/2$.
Taking $u=\zeta_k/(48HW_a)$, a sufficient count is
\[
 m\ge\left\lceil\left[
 \frac{72HW_a^2r^2}{\zeta_k^2}
       \left(1+\frac2{\lfloor G/2\rfloor}\right)
 +\frac{32HW_ar}{\zeta_k}\right]
 \log\frac{4\mathcal N_k(\zeta_k/(48HW_a))}{\delta_k}\right\rceil.
\]
Each of the three error terms is then at most $\zeta_k/12$, so $2\times$
the uniform error is at most $\zeta_k/2$.

\subsubsection{Fitting TIS with KL regularization}
\textbf{Separate concentration for prompts and responses.}
\label{app:sc-kl-samples}
Fix a round and use $\beta\ge2$, $\eta=1/\beta$, and $G\ge2$.
Write $\ell>0$ for the tail parameter in the concentration bounds below.
The fitted objective is Equation~\eqref{eq:theory-empirical-objective}
with cap-two reward and penalty weights. Its unbiasedness and gradient
identity are given by Proposition~\ref{prop:sc-kl-sgd}.
Throughout this argument, $D_k(x)$ is the TIS objective gap in
Equation~\eqref{eq:sc-kl-exact-gap}, and $\mathcal L_k$ and $\kappa$ are
the capped loss and coefficient in Equation~\eqref{eq:sc-kl-capped-transfer}.
We first bound the response noise by the objective gap; accurate
candidates then require less response sampling.
For a candidate $\pi$, write $\rho_t=\pi(a_t\mid h_t)/\mu(a_t\mid h_t)$
and $\rho_t^*=\pi_k^*(a_t\mid h_t)/\mu(a_t\mid h_t)$. Let $D=\E_xD_k(x)$ be the exact gap in
\eqref{eq:sc-kl-exact-gap}, and put
$V=\sum_t\E_{d_t^{\mu}}\E_{{\mu_t}}(\rho_t^*-\rho_t)^2$.
For one response define
$Z=\sum_t(\rho_t^*-\rho_t)$ and
$B=\sum_t[k(\rho_t^*)-k(\rho_t)]$.
The sample objective excess is $\mathcal H=RZ-\beta B$.
For two independent responses to the same prompt, define
\begin{equation}\label{eq:rsc-response-pair-kernel}
 \mathcal H_2(Y,Y')=\tfrac12(R-R')(Z-Z')-\tfrac\beta2(B+B').
\end{equation}
\begin{lemma}[Curvature and conditional response moments]\label{lem:rsc-response-moments}
The objective gap satisfies
\begin{equation}\label{eq:sc-kl-curvature}
 D\ge\tfrac12\beta e^{-\eta}V.
\end{equation}
For the symmetric pair kernel in \eqref{eq:rsc-response-pair-kernel}, conditional on a prompt,
$\E \mathcal H_2=D_k(x)$, $\E \mathcal H_2^2\le16HD_k(x)/\beta$, and
$|\mathcal H_2|\le9H/\beta$.
\end{lemma}
\begin{proof}
The scalar curvature $\beta k''(\rho)=\beta/\rho$ and feasibility of
$\pi_k^*$ imply
Equation~\eqref{eq:sc-kl-curvature}.

Normalization makes the increments of $Z$ martingale differences,
so $\E Z^2=V$. Since $|k'(\rho)|\le\eta$,
$\E B^2\le H\eta^2 V$. Therefore
\[
 \E\mathcal H^2\le2(1+H)V
 \le\frac{4e^\eta(1+H)}\beta D
 \le\frac{16H}\beta D.
\]
All these bounds also hold conditional on a prompt, with $D_k(x)$
and the corresponding conditional squared-ratio error.
For $G\ge2$, the group excess is the average of the symmetric pair
kernel
\eqref{eq:rsc-response-pair-kernel}.
Its conditional expectation is $D_k(x)$. The squared reward term
has expectation at most $\E[Z^2\mid x]/2$. Jensen's inequality bounds the squared
penalty term by $\beta^2\E[B^2\mid x]$.
Writing $V_x$ for $V$ conditional on the prompt, we obtain
\[
 \E[\mathcal H_2^2\mid x]\le(1+2H)V_x
 \le\frac{2e^\eta(1+2H)}\beta D_k(x)
 \le\frac{16H}\beta D_k(x),
\]
Jensen's inequality gives the same bound for the complete pair average $U_G$.

The interval bounds give $|Z|\le2e^\eta H/\beta$ and
$0\le k(\rho)\le e^\eta\eta^2$. In particular, the pair kernel
satisfies $|\mathcal H_2|\le9H/\beta$. Its conditional mean is $D_k(x)$ and
its conditional second moment is at most $16HD_k(x)/\beta$.
We use these moment bounds in Lemma~\ref{lem:rsc-response-concentration}.
\end{proof}

The moment bound gives a concentration inequality conditional on the
sampled prompts.
\begin{lemma}[Response concentration within sampled prompts]
\label{lem:rsc-response-concentration}
For each fixed candidate, conditional on the prompts, with probability
at least $1-e^{-\ell}$,
$\overline D\le2\widehat D+44H\ell/(\beta m\lfloor G/2\rfloor)$.
Here $\overline D=m^{-1}\sum_jD_k(x_j)$, with $D_k$ from
Equation~\eqref{eq:sc-kl-exact-gap}, and
$\widehat D=\widehat{\mathcal W}_k(\pi_k^*)-
\widehat{\mathcal W}_k(\pi)$ is the empirical gap for the
objective in Equation~\eqref{eq:theory-empirical-objective}.
\end{lemma}
\begin{proof}
Write $L=\lfloor G/2\rfloor$. Conditional on the $m$ prompts,
randomly pair the responses in each group, leaving one unused when
$G$ is odd. The complete pair average is the conditional expectation
of this paired estimator. Jensen's inequality for its exponential
moment therefore transfers the Bernstein bound for $mL$ independent
pairs to the actual grouped estimator.

Let $\overline D=m^{-1}\sum_jD_k(x_j)$ and let $\widehat D$ be
the empirical objective gap to $\pi_k^*$. The bounds on the pair
kernel give, with tail parameter $\ell$,
\[
 \overline D-\widehat D
 \le\sqrt{\frac{32H\overline D\ell}{\beta mL}}
       +\frac{6H\ell}{\beta mL}
 \le\overline D/2+\frac{22H\ell}{\beta mL}.
\]
Thus $\overline D\le2\widehat D+44H\ell/(\beta mL)$.
This step controls response noise conditional on the prompts,
so its effective sample size is $mL$.
\end{proof}

A separate concentration bound handles variation between prompts.
The cap in Equation~\eqref{eq:sc-kl-capped-transfer} bounds each
prompt loss by one.
\begin{lemma}[Concentration of capped prompt losses]\label{lem:rsc-prompt-concentration}
For a fixed candidate and $\mathcal L_k(x)=\min\{1,\kappa D_k(x)\}$,
with probability at least $1-e^{-\ell}$,
$\E_x\mathcal L_k\le2m^{-1}\sum_j\mathcal L_k(x_j)+4\ell/m$.
\end{lemma}
\begin{proof}
Use $\mathcal L_k(x,\pi)=\min\{1,\kappa D_k(x,\pi)\}$ from
\eqref{eq:sc-kl-capped-transfer}. For a fixed candidate, write
$P=\E_x\mathcal L_k$ and $\widehat P=m^{-1}\sum_j\mathcal L_k(x_j)$.
Since $0\le\mathcal L_k\le1$, its variance is at most $P$.
One-sided Bernstein gives
\[
 P-\widehat P\le\sqrt{2P\ell/m}+\ell/(3m),
 \qquad P\le2\widehat P+4\ell/m.
\]
The cap bounds the cost of each prompt by its possible reward loss.
Its range is therefore independent of $H$ and $\beta$.

\end{proof}

A covering argument makes both concentration bounds hold for the
policy selected using the same dataset.
\begin{lemma}[Uniform fitting certificate for TIS with KL]\label{lem:rsc-uniform-fitting}
Let $L=\lfloor G/2\rfloor$. For a radius-$v$ cover and
$\ell=\log(4K\mathcal N(v)/\delta)$, outside an event of probability
at most $\delta/K$, every fitted policy satisfying the empirical
accuracy condition in Equation~\eqref{eq:theory-fitting-condition} obeys
\begin{equation}\label{eq:sc-kl-two-level-certificate}
 \E_x\mathcal L_k(x,\pi_{k+1})
 \le4\kappa e_k+\frac{88\kappa H\ell}{\beta mL}
       +\frac{4\ell}{m}+54\kappa Hv.
\end{equation}
\end{lemma}
\begin{proof}
For the log-policy metric in Equation~\eqref{eq:sc-cover-metric}, the sampled reward term is
$2He^\eta$-Lipschitz, and the penalty is
$\beta H\eta e^\eta$-Lipschitz. Their sum is at most $6H$.
The same bound holds for the conditional gap $D_k(x,\pi)$ and for
the grouped empirical gap $\widehat D$: its reward weights satisfy
$\sum_g|R_g-\overline R|/(G-1)\le G/(2(G-1))\le1$, and its penalty is
a response average. The capped loss is $6\kappa H$-Lipschitz.
Apply both concentration inequalities to a radius-$v$ cover, with
\[
 \ell=\log\frac{4K\mathcal N(v)}\delta.
\]
Conditional on the past, a union bound and extension from the cover
give, simultaneously for every candidate,
\[
 \overline D\le2\widehat D+\frac{44H\ell}{\beta mL}+18Hv,
 \qquad
 P\le2\widehat P+\frac{4\ell}{m}+18\kappa Hv.
\]
Since $\widehat P\le\kappa\overline D$, the fitted policy satisfies
Equation~\eqref{eq:sc-kl-two-level-certificate}.
Here $\widehat D\le e_k$ follows directly from the empirical oracle,
since the target belongs to the full feasible family.
\end{proof}

\subsubsection{Fitting unit-weight SC with KL regularization}\label{sec:unit-inner-fitting}
The preceding TIS fitting analysis uses exact conditional centers. For unit-weight SC, we also allow sampled centers, so their randomness must be included in the empirical objective. Use the independent auxiliary tokens and fixed-dataset sampling scheme
of Equation~\eqref{eq:theory-mc-center}. For the unit reward weight,
$F_a(\rho)=\log\rho$. The following sample terms make that specialization
explicit for the moment bounds below.

For an augmented response $(Y,U)$, put
\begin{align}
 Z_\pi(Y,U)&=\sum_{t=1}^H\left[
       \log \rho_t^\pi(h_t,a_t)
       -\frac1M\sum_{i=1}^M\log \rho_t^\pi(h_t,U_{t,i})
       \right],\nonumber\\
 B_\pi(Y)&=\sum_{t=1}^H k\!\left(\rho_t^\pi(h_t,a_t)\right).
 \label{eq:chenlu-sc-kl-sample-terms}
\end{align}
The empirical objective is
\begin{equation}\label{eq:chenlu-sc-kl-empirical}
 \widehat{\mathcal W}_k(\pi)=\frac1m\sum_{j=1}^m
 \left[
 \frac1{G-1}\sum_{g=1}^G(R_{jg}-\overline R_j)Z_\pi(Y_{jg},U_{jg})
 -\frac2G\sum_{g=1}^G B_\pi(Y_{jg})
 \right].
\end{equation}
Here $\overline R_j=G^{-1}\sum_{g=1}^G R_{jg}$.
The reward term is the same-prompt sample covariance, including its
$G/(G-1)$ correction. Since $\E[Z_\pi\mid x]=0$ and
$\E[RZ_\pi\mid x]=C_{k,x}(\pi)$, this objective is unbiased for
$\mathcal W_k(\pi)$ for each fixed candidate. Its reward gradient uses
the unweighted conditional score center, estimated by $M$ independent
draws. The penalty potential is $k(\rho)$, with gradient
$\rho\log \rho\,\nabla_\theta\log\pi_\theta$. No random advantage-standard-deviation
normalization or variable-length token weighting is included.

The fitting condition is Equation~\eqref{eq:theory-fitting-condition},
with the radius-$1/2$ family and the covering numbers defined before
Theorem~\ref{thm:chenlu-sc-kl-convergence}.

\textbf{Curvature and conditional response moments.}
Use the unit-weight objective gap in Equation~\eqref{eq:chenlu-sc-kl-gap}.
Write $\rho_t^* =\pi_k^*/{\mu}$ and define
\[
 \mathcal V_k(x,\pi)=\E_{\mu}\!\left[
       \sum_t\bigl(\rho_t^*(h_t,a_t)-\rho_t^\pi(h_t,a_t)\bigr)^2
       \,\middle|\,x\right].
\]
\begin{lemma}[Quadratic control by the objective gap]
\label{lem:chenlu-sc-kl-curvature}
For the family $\Gamma_k$ in Section~\ref{app:chenlu-pure-sc-kl}
and the gap in Equation~\eqref{eq:chenlu-sc-kl-gap}, every $\pi\in\Gamma_k$ satisfies
$D_k(x,\pi)\ge\mathcal V_k(x,\pi)/4$.
\end{lemma}
\begin{proof}
The negative local objective has scalar integrand
$\phi_A(u)=-A\log u+2k(u)$. For $|A|\le1$ and
$u\in[e^{-1/2},e^{1/2}]$,
\[
 \phi_A''(u)=\frac{A+2u}{u^2}
 \ge\frac{2u-1}{u^2}\ge\frac12.
\]
The last inequality holds on $[2-\sqrt2,2+\sqrt2]$, which contains
the feasible ratio interval. At each history, feasible ratio vectors
form a convex compact set with linear normalization. Strong convexity
and the first-order optimality inequality, including at boundary
optima, yield the stated gap bound after averaging histories.
\end{proof}

For Lemma~\ref{lem:chenlu-sc-kl-moments}, fix $k,x,\pi$ and abbreviate
$D=D_k(x,\pi)$, $\mathcal V=\mathcal V_k(x,\pi)$,
$Z=Z_{\pi_k^*}-Z_\pi$, and $B=B_{\pi_k^*}-B_\pi$,
using the sample terms in Equation~\eqref{eq:chenlu-sc-kl-sample-terms}.
For two conditionally independent augmented responses define
\begin{equation}\label{eq:chenlu-sc-kl-pair}
 \mathcal H((Y,U),(Y',U'))
   =\tfrac12(R-R')(Z-Z')-(B+B').
\end{equation}
This kernel has conditional mean $D$. Its complete pair average is
the grouped empirical excess objective in
Equation~\eqref{eq:chenlu-sc-kl-empirical}.

\begin{lemma}[Conditional moments with finite Monte Carlo centers]
\label{lem:chenlu-sc-kl-moments}
For every $\pi\in\Gamma_k$, with $\Gamma_k$ as in Section~\ref{app:chenlu-pure-sc-kl}, the kernel in Equation~\eqref{eq:chenlu-sc-kl-pair} satisfies
\[
 \E[\mathcal H^2\mid x]\le C_MHD,\qquad |\mathcal H|\le3H,
 \qquad C_M=8+4e(1+1/M).
\]
\end{lemma}
\begin{proof}
Put $\Delta z_t=\log \rho_t^*-\log \rho_t^\pi$.
The increments
\[
 \Delta z_t(h_t,a_t)
       -\frac1M\sum_{i=1}^M\Delta z_t(h_t,U_{t,i})
\]
are martingale differences on the augmented response filtration.
Their conditional variance is
$(1+1/M)\Var_{{{\mu_t}}}(\Delta z_t)$.
The logarithm is $e^{1/2}$-Lipschitz on the feasible ratio interval,
so orthogonality of these increments gives
\[
 \E[Z\mid x]=0,\qquad
 \E[Z^2\mid x]\le(1+1/M)e\,\mathcal V.
\]
Since $|k'(u)|=|\log u|\le1/2$, Cauchy-Schwarz gives
$\E[B^2\mid x]\le(H/4)\mathcal V$. Using
$(a-b)^2\le2a^2+2b^2$, $|R-R'|\le1$, and independence of the
two augmented responses,
\begin{align*}
 \E[\mathcal H^2\mid x]
 &\le\E[Z^2\mid x]+8\E[B^2\mid x]\\
 &\le[(1+1/M)e+2H]\mathcal V
 \le C_MHD,
\end{align*}
where the last step uses Lemma~\ref{lem:chenlu-sc-kl-curvature} and
$H\ge1$. Also $|\Delta z_t|\le1$ gives $|Z|\le2H$.
On the feasible interval $0\le k(u)<1/4$, so $|B|\le H/4$.
Equation~\eqref{eq:chenlu-sc-kl-pair} then gives
$|\mathcal H|\le(5/2)H\le3H$.
\end{proof}

The moment bound vanishes with the objective gap. Finite independent
Monte Carlo centers retain this property because the same auxiliary
draws are used for all candidate policies in the empirical objective.
When the candidate equals the comparator, their sampled centers agree
on every dataset, so this source of comparison noise also vanishes.

\textbf{From empirical fitting to population fitting.}
Condition on the past before round $k$. For sampled prompts $x_j$, let
\[
 \overline D(\pi)=\frac1m\sum_jD_k(x_j,\pi),\qquad
 \widehat D(\pi)=\widehat{\mathcal W}_k(\pi_k^*)
                         -\widehat{\mathcal W}_k(\pi).
\]
For $\kappa>0$, define
\[
 \mathcal L_k(x,\pi)=\min\{1,\kappa D_k(x,\pi)\},\qquad
 P(\pi)=\E_x\mathcal L_k(x,\pi),\qquad
 \widehat P(\pi)=\frac1m\sum_j\mathcal L_k(x_j,\pi).
\]

\begin{lemma}[Uniform fitting certificate for unit-weight SC with KL]
\label{lem:chenlu-sc-kl-fitting}
Let $L=\lfloor G/2\rfloor$, $B_M=2(C_M+4)$, and
$\ell=\log(4K\mathcal N(\nu)/\delta)$.
Except on an event of conditional probability at most $\delta/K$,
every candidate satisfying Equation~\eqref{eq:theory-fitting-condition} obeys
\begin{equation}\label{eq:chenlu-sc-kl-fitting-bound}
 P(\pi_{k+1})\le
 4\kappa e_k+\frac{2\kappa B_MH\ell}{mL}
 +\frac{4\ell}{m}+36\kappa H\nu.
\end{equation}
\end{lemma}
\begin{proof}
\textbf{Step 1: response concentration.}
First fix a candidate and condition on the sampled prompts. Randomly
pair the $G$ responses in each group, leaving one out when $G$ is odd.
For a fixed pairing there are $mL$ conditionally independent kernels.
The complete pair average is the conditional expectation over these
pairings, so Jensen's inequality transfers the exponential-moment bound
to the actual grouped estimator. By
Lemma~\ref{lem:chenlu-sc-kl-moments}, the summed conditional variance is
at most $mL C_MH\overline D$, and each centered kernel has absolute
bound $6H$. Bernstein's inequality therefore implies, with probability
at least $1-e^{-\ell}$,
\begin{align*}
 \overline D-\widehat D
 &\le\sqrt{\frac{2C_MH\overline D\,\ell}{mL}}
             +\frac{4H\ell}{mL}\\
 &\le\tfrac12\overline D+\frac{(C_M+4)H\ell}{mL},
\end{align*}
and hence
\begin{equation}\label{eq:chenlu-sc-kl-response-concentration}
 \overline D\le2\widehat D+\frac{B_MH\ell}{mL}.
\end{equation}
\textbf{Step 2: prompt concentration.}
For the independent prompt draws, $0\le\mathcal L_k\le1$ and
$\Var_x\mathcal L_k\le P$. The same Bernstein inversion gives
\begin{equation}\label{eq:chenlu-sc-kl-prompt-concentration}
 P\le2\widehat P+\frac{4\ell}{m}
\end{equation}
with probability at least $1-e^{-\ell}$.

\textbf{Step 3: extend uniformly over fitted policies.}
To make Equations~\eqref{eq:chenlu-sc-kl-response-concentration} and~\eqref{eq:chenlu-sc-kl-prompt-concentration} uniform, use the continuity of the empirical objective. The centered sampled log-potential
$Z_\pi$ is $2H$-Lipschitz in $d_\infty$. For $R_g\in[0,1]$,
\[
 \frac1{G-1}\sum_g|R_g-\overline R|
 \le\frac{G}{2(G-1)}\le1.
\]
The derivative of $2k(u)$ with respect to $\log u$ is $2u\log u$,
whose absolute value is at most $e^{1/2}$ on the feasible interval.
Thus the grouped empirical objective and its population expectation
are at most $(2+e^{1/2})H<4H$-Lipschitz. The conditional gap
$D_k(x,\pi)$ has the same bound, and $\mathcal L_k$ is
$4\kappa H$-Lipschitz.

Apply Equations~\eqref{eq:chenlu-sc-kl-response-concentration}
and~\eqref{eq:chenlu-sc-kl-prompt-concentration} to a feasible
radius-$\nu$ cover. The union bound has failure probability at most
$2\mathcal N(\nu)e^{-\ell}\le\delta/K$.
Extending from each cover center to its candidate gives, simultaneously,
\[
 \overline D\le2\widehat D+\frac{B_MH\ell}{mL}+12 H\nu,
 \qquad
 P\le2\widehat P+\frac{4\ell}{m}+12\kappa H\nu.
\]
Finally $\widehat P\le\kappa\overline D$ and
$\widehat D(\pi_{k+1})\le e_k$, because $\pi_k^*$ is a feasible
empirical comparator. Substitution proves
Equation~\eqref{eq:chenlu-sc-kl-fitting-bound}.
\end{proof}

We now combine the outer-loop progress lemmas with the uniform fitting
certificates. Fresh data at each round allow the certificates to be
applied conditional on all previous rounds. The resulting sufficient
round counts are $O(H\log(N/\epsilon))$ for $\beta=0$ and
$O(\log(N/\epsilon))$ for the regularized settings in the main
theorems.
\subsubsection{Combined bounds for unregularized SC and TIS}\label{sec:theory-joint}
The uniform fitting bound controls each round's population loss. Substituting this bound into the outer progress inequality gives the following reward and sample guarantee.
\begin{theorem}[Best-of-N improvement and sample budget]
\label{thm:sc-sample-complexity}
Use SC, or SC+TIS with $c\ge e^r$, and set
\[
 r=\frac1{8H},\qquad
 K=\left\lceil512H\log\frac{6N}{\epsilon}\right\rceil,
 \quad N\ge1,\quad 0<\epsilon<1.
\]
At each round set ${\mu}=\pi_k$ and let
\[
 \Gamma_k(r)=\left\{\pi:
 \left|\log\frac{\pi(b\mid h_t)}{{\mu}(b\mid h_t)}\right|\le r
 \text{ for every reachable }h_t,b,t\right\}.
\]
These are full families of normalized token policies. Let
$\mathcal N(u)$ be a common nonincreasing upper bound on the
radius-$u$ covering numbers of $\Gamma_k(r)$ in $d_\infty$,
with centers in $\Gamma_k(r)$. Draw $m$ fresh prompts per round and $G\ge2$
independent behavior responses per prompt. Suppose the fitted policy
$\pi_{k+1}\in\Gamma_k(r)$ satisfies
\[
 \widehat{\mathcal C}_k(\pi_{k+1})
 \ge\sup_{\pi\in\Gamma_k(r)}\widehat{\mathcal C}_k(\pi)
                  -\frac{\epsilon}{128\pi^2NK^2}.
\]
For $0<\delta<1$, if
\begin{equation}\label{eq:sc-cover-samples}
 m\ge\left\lceil
 \frac{32K+12544\pi^2NK^2/\lfloor G/2\rfloor}{\epsilon}
 \log\!\left[\frac{4K}{\delta}\,
 \mathcal N\!\left(\frac{\epsilon}{3840\pi^2NK^2H}\right)^2
 \right]\right\rceil,
\end{equation}
then after $K$ rounds, using $Km$ prompt draws and $KmG$ responses,
$V_N(\pi_0)-J(\pi_K)\le\epsilon$ with probability at least $1-\delta$.
\end{theorem}

\begin{proof}[Proof of Theorem~\ref{thm:sc-sample-complexity}]
Apply Lemma~\ref{lem:sc-uniform-fitting} with
$e_k=\epsilon/(128\pi^2NK^2)$ and $\kappa=4\pi^2NK$.
Choose $u=\epsilon/(3840\pi^2NK^2H)$.
Each of the four terms is at most $\epsilon/(8K)$ under
\eqref{eq:sc-cover-samples}. Thus each round contributes at most
$\epsilon/(2K)$ to \eqref{eq:sc-capped-transfer}.

The concentration events hold conditionally on earlier rounds and
uniformly over fitted policies. Fresh samples and a union bound give
simultaneous success with probability at least $1-\delta$.
Equation~\eqref{eq:sc-capped-transfer} then proves the reward guarantee.
Finally, $K\le513H\log(6N/\epsilon)$ and
$\lfloor G/2\rfloor\ge G/3$ give \eqref{eq:sc-total-samples}.
\end{proof}
\textbf{Unregularized budget substitution.}
Choosing $m$ as the sufficient ceiling in Equation~\eqref{eq:sc-cover-samples}
gives the total prompt budget
\begin{equation}\label{eq:sc-total-samples}
 \begin{aligned}
 Km=O\!\Biggl(&\frac{H^2\log^2(6N/\epsilon)
            +(NH^3/G)\log^3(6N/\epsilon)}{\epsilon}\\
 &\times\log\!\left[\frac{2052H\log(6N/\epsilon)}{\delta}\,
 \mathcal N\!\left(
 \frac{\epsilon}{3840\cdot513^2\pi^2NH^3\log^2(6N/\epsilon)}
 \right)^2\right]\Biggr).
 \end{aligned}
\end{equation}

\subsubsection{Proof of Theorem~\ref{thm:sc-kl-progress}}\label{sec:proof-tis-kl}
For regularized TIS, the same combination uses the exact target gap and its reward-transfer bound. The proof below uses the fitting certificate to bound the error in each round.

\begin{proof}[Proof of Theorem~\ref{thm:sc-kl-progress}]
Choose
\[
 v=\frac{\epsilon}{2160\pi^2NK^2H},\qquad
 e_k\le\frac{\epsilon}{160\pi^2NK^2}.
\]
With $\kappa=10\pi^2NK/\beta$ and $\beta=2$,
the first and fourth terms in
\eqref{eq:sc-kl-two-level-certificate} are each at most
$\epsilon/(8K)$. The sample count \eqref{eq:sc-kl-samples}
bounds each of the other two terms by $\epsilon/(8K)$.
Thus $\E_x\mathcal L_k\le\epsilon/(2K)$ at every successful round.

\textbf{Combining the rounds.}
The concentration statements are uniform over fitted policies and
hold conditionally on all earlier rounds. Fresh data at each round
and a union bound give simultaneous success with probability at
least $1-\delta$. Summing
\eqref{eq:sc-kl-capped-transfer} proves
Theorem~\ref{thm:sc-kl-progress}.
Finally, choose $m$ as the sufficient ceiling in
Equation~\eqref{eq:sc-kl-samples}. Then $K\le5\log(6N/\epsilon)$ and
$\lfloor G/2\rfloor\ge G/3$ yield
\eqref{eq:sc-kl-total-prompts}. The first budget term estimates
capped prompt losses. The second estimates conditional objective
gaps using the responses within each group.

\end{proof}

Choosing the sufficient prompt count in
Equation~\eqref{eq:sc-kl-samples} gives the total prompt budget
\begin{equation}\label{eq:sc-kl-total-prompts}
\begin{aligned}
 Km=O\!\Biggl(&\frac{\log^2(6N/\epsilon)
       +(NH/G)\log^3(6N/\epsilon)}{\epsilon}\\
 &\times\log\!\left[\frac{20\log(6N/\epsilon)}{\delta}\,
 \mathcal N\!\left(
 \frac{\epsilon}{54000\pi^2NH\log^2(6N/\epsilon)}
 \right)\right]\Biggr).
\end{aligned}
\end{equation}

\subsubsection{Proof of Theorem~\ref{thm:chenlu-sc-kl-convergence}}\label{sec:proof-unit-kl}

\textbf{Reward transfer and the outer-round guarantee.}
\begin{proof}[Proof of Theorem~\ref{thm:chenlu-sc-kl-convergence}]
Write $s_k(x)=J_x(\pi_k)$ and
$f(v)=67v/(60+7v)$.
Lemma~\ref{lem:chenlu-sc-kl-progress} and
$v+\tfrac7{60}v(1-v)\ge f(v)$ imply
\[
 [f(s_k(x))-s_{k+1}(x)]_+\le D_k(x,\pi_{k+1}).
\]
Let $\alpha_k=\arcsin\sqrt{s_k}$ and
$T(\theta)=\arctan(e^{\lambda/2}\tan\theta)$, with
$\lambda=\log(67/60)$. Then $\sin^2T(\alpha_k)=f(s_k)$.
The probability-shortfall part of
Lemma~\ref{lem:bernoulli-angular-transfer} gives
\[
 \xi_k(x)^2:=[T(\alpha_k(x))-\alpha_{k+1}(x)]_+^2
 \le\frac{\pi^2}{4}D_k(x,\pi_{k+1}).
\]
Use the same lemma at accuracy $\epsilon/2$. The choice of $K$ in
Equation~\eqref{eq:chenlu-sc-kl-constants} and
$\min\{1,\sum_k y_k\}\le\sum_k\min\{1,y_k\}$ for $y_k\ge0$ yield
\begin{equation}\label{eq:chenlu-sc-kl-transfer}
 V_N(\pi_0)-J(\pi_K)
 \le\frac\epsilon2+
 \sum_{k<K}\E_x\min\{1,4\pi^2NKD_k(x,\pi_{k+1})\}.
\end{equation}
For the stated $e_k$, $\nu$, and $m$, each of the four terms in
Equation~\eqref{eq:chenlu-sc-kl-fitting-bound} is at most
$\epsilon/(8K)$. Thus each expectation in
Equation~\eqref{eq:chenlu-sc-kl-transfer} is at most
$\epsilon/(2K)$, except on an event of conditional probability
$\delta/K$. The data are fresh at every round, so the uniform
certificates apply conditionally on all previous rounds. A union bound
over $K$ rounds and Equation~\eqref{eq:chenlu-sc-kl-transfer} prove
the claim.
\end{proof}

\textbf{Total sample budget.}
For fixed $M$, choosing $m$ as the ceiling in
Equation~\eqref{eq:chenlu-sc-kl-samples} gives the sufficient total counts
\begin{equation}\label{eq:chenlu-sc-kl-total-samples}
 Km=O\!\left[
 \frac{K^2+NHK^3/\lfloor G/2\rfloor}{\epsilon}
 \log\!\left(\frac{4K\mathcal N(\nu)}{\delta}\right)\right],
 \qquad\text{responses}=GKm.
\end{equation}
Since $K=O(\log(N/\epsilon))$, the sufficient prompt and response
factors are $(1+NH/G)/\epsilon$ and $(G+NH)/\epsilon$, respectively,
up to logarithms and the covering-number term. Finite Monte Carlo
centers enter through $B_M$ in Equation~\eqref{eq:chenlu-sc-kl-constants}.
For finite $M$, they require at most $MHGKm$ additional token draws.

\subsection{Surrogate integrands and derivatives used in the main text}
\label{sec:surrogate-details}
The convergence arguments above use population objectives and fitting gaps. To relate these objectives to the token updates plotted in the main text, the following calculations give the integrands and derivatives in Table~\ref{tab:surrogate-methods}
and Figure~\ref{fig:regularized-potentials}. At a fixed history, write
$\rho=\pi_\theta(a\mid h)/\mu(a\mid h)$ and $A=A^\mu(h,a)$, and hold $\mu$, $A$, and
the other learner logits fixed. Here $A$ is the expected advantage under $\mu$ defined in
Equation~\eqref{eq:theory-basic-notation}; $A_i$ in the sampled update
of Equation~\eqref{eq:rsc-total-gradient} is a group-centered reward.

For the clipped reward weight used in Section~\ref{sec:sc-background},
\begin{equation}\label{eq:sc-unit-weight-clipping}
 a(u)=\operatorname{clip}(u,c_{\mathrm{low}},c_{\mathrm{high}}),
 \qquad 0\le c_{\mathrm{low}}\le1\le c_{\mathrm{high}},
\end{equation}
integrating $a(u)/u$ gives
\begin{equation}\label{eq:continuous-weight-primitive}
 F_a(u)=
 \begin{cases}
 c_{\mathrm{low}}-1+c_{\mathrm{low}}\log(u/c_{\mathrm{low}}),
     &u<c_{\mathrm{low}},\\
 u-1,&c_{\mathrm{low}}\le u\le c_{\mathrm{high}},\\
 c_{\mathrm{high}}-1+c_{\mathrm{high}}\log(u/c_{\mathrm{high}}),
     &u>c_{\mathrm{high}}.
 \end{cases}
\end{equation}
The first branch is absent when $c_{\mathrm{low}}=0$. Thus the intervals
$[1,1]$, $[0,c]$, and $[0,\infty)$ give unit-weight SC, one-sided TIS,
and full IS, respectively. In particular, SC uses $F_a(u)=\log u$,
whereas full IS uses $F_a(u)=u-1$.

With standard reverse KL, define $k(\rho)=\rho\log\rho-\rho+1$.
The single-action contribution to the main-text objective is
\[
 w=A F_a(\rho)-\beta k(\rho).
\]
Averaging over actions under $\mu$ and over behavior histories gives
Equation~\eqref{eq:method-regularized-target}, since
$\E_\mu[\rho\mid h]=1$ and $\E_\mu[k(\rho)\mid h]=\KL(\pi_\theta\|\mu)$.
The identity $\rho F_a'(\rho)=a(\rho)$ gives
\[
 \frac{\partial w}{\partial\rho}
 =\frac{A a(\rho)}\rho-\beta\log\rho.
\]
For the selected action's softmax logit $z_a$,
$\partial\rho/\partial z_a=\rho[1-\pi_\theta(a\mid h)]$. Therefore
\begin{equation}\label{eq:main-text-logit-derivative}
 \frac{\partial w}{\partial z_a}
 =\bigl[A a(\rho)-\beta\rho\log\rho\bigr][1-\pi_\theta(a\mid h)].
\end{equation}
Setting $\beta=0$ gives the unregularized rows of
Table~\ref{tab:surrogate-methods}; choosing $a=1$ or $a(\rho)=\min(\rho,c)$
gives the two RSC rows. At the TIS cap, the derivatives from both sides
agree. Equation~\eqref{eq:main-text-logit-derivative} describes the
selected action's contribution. The complete conditional objective
averages all action contributions under $\mu$.

Figure~\ref{fig:regularized-potentials} uses $A=1$, TIS cap $c=2$,
$\beta=0.05$ for RSC, and uncapped reverse KL. Its derivative coefficients
are those in Equation~\eqref{eq:main-text-logit-derivative} after removing
the common factor $1-\pi_\theta(a\mid h)$. The curves use positive ratios from
$10^{-5}$ to $10$, with denser sampling near zero and explicit points at
one and two. At fixed $\mu(a\mid h)>0$, the ratio is bounded by
$1/\mu(a\mid h)$ and the displayed logit derivative vanishes as
$\pi_\theta(a\mid h)\to1$. Across histories, large ratios can arise from small
behavior probabilities, as discussed in Section~\ref{sec:sc-analysis}.
For an active penalty cap, the corresponding coefficient is instead
$-\beta b(\rho)\log b(\rho)$, derived in
Proposition~\ref{prop:sc-kl-sgd}.

\paragraph{Population gradient identities.}
With behavior $\mu$, its advantages, and clipping bounds fixed, $\nabla_\theta\rho=\rho\psi_\theta$ and $\rho F_a'(\rho)=a(\rho)$ give
\begin{equation}\label{eq:method-reward-surrogate-gradient}
 \nabla_\theta\E_\mu\!\left[\sum_{t=1}^H A_t^\mu F_a(\rho_t)\right]
 =\E_\mu\!\left[\sum_{t=1}^H A_t^\mu\rho_t^{\mathrm{TIS}}\psi_\theta(h_t,a_t)\right].
\end{equation}
For standard reverse KL at a fixed history, the same differentiation gives
\begin{equation}\label{eq:method-kl-gradient}
 \begin{aligned}
 \nabla_\theta\KL({\pi_\theta}(\cdot\mid h)\|{\mu}(\cdot\mid h))
 &=\E_{\mu}[\rho(\log \rho+1)\psi_\theta\mid h]\\
 &=\E_{\mu}[\rho\log \rho\,\psi_\theta\mid h].
 \end{aligned}
\end{equation}
Normalization gives $\E_\mu[\rho\psi_\theta\mid h]=0$, which removes the $+1$ term. Combining the reward and penalty identities gives Equation~\eqref{eq:method-regularized-surrogate-gradient}.

\end{document}